\pdfoutput=1

\documentclass[11pt]{article}

\usepackage{acl}

\usepackage{pdflscape}

\usepackage{times}
\usepackage{latexsym}
\usepackage{acronym}
\usepackage{subcaption}
\usepackage{booktabs}
\usepackage{longtable}
\usepackage{hyperref}
\usepackage{geometry}
\usepackage{amsmath}
\usepackage{cleveref}
\usepackage{enumitem}
\usepackage[T1]{fontenc}

\usepackage[utf8]{inputenc}

\usepackage{microtype}

\usepackage{inconsolata}

\usepackage{graphicx}

\usepackage{soul}
\usepackage{booktabs}
\usepackage{tabularx}
\usepackage{longtable}
\usepackage{amssymb}
\DeclareMathSymbol{\shortminus}{\mathbin}{AMSa}{"39}
\usepackage{amsmath}

\usepackage[acronym]{glossaries}
\glsdisablehyper

\newacronym{LM}{LM}{Language Model}
\newacronym{LLM}{LLM}{Large Language Model}
\newacronym{QA}{QA}{Question Answering}
\newacronym{BPE}{BPE}{Byte-Pair Encoding}
\newacronym{BBPE}{BBPE}{Byte-Level BPE}
\newacronym{SP}{SP}{SentencePiece}
\newacronym{HF}{HF}{Hugging Face}
\newacronym{NLP}{NLP}{Natural Language Processing}
\newacronym{xMMLU}{EU20-MMLU}{}
\newacronym{xHella}{EU20-HellaSwag}{}
\newacronym{xARC}{EU20-ARC}{}
\newacronym{xTQA}{EU20-TruthfulQA}{}{}
\newacronym{xGSM8K}{EU20-GSM8K}{}

\title{Towards Multilingual LLM Evaluation for European Languages}

\author{
   Klaudia Thellmann\textsuperscript{1}$^\dagger$
   Michael Fromm\textsuperscript{2,3}$^\dagger$
   Bernhard Stadler\textsuperscript{1}$^\dagger$  
   Jasper Schulze Buschhoff\textsuperscript{2}$^\dagger$\\
    Alex Jude\textsuperscript{2} 
    Fabio Barth\textsuperscript{4} 
    Johannes Leveling\textsuperscript{2} 
    Nicolas Flores-Herr\textsuperscript{2} 
    Joachim Köhler\textsuperscript{2} 
    René Jäkel\textsuperscript{1} 
    Mehdi Ali\textsuperscript{2,3}     \\
    \\
  \textsuperscript{1}TU Dresden, Germany, \textsuperscript{2} Fraunhofer IAIS, Germany \\ \textsuperscript{3}Lamarr Institute, Germany,  \textsuperscript{4} DFKI, Germany\\
  \\
  \\
  \\  \\
   \\   \\
   \Thanks{\textdagger Equal contribution.}
}

\begin{document}

\maketitle

\begin{abstract}
The rise of Large Language Models (LLMs) has revolutionized natural language processing across numerous languages and tasks.
However, evaluating LLM performance in a consistent and meaningful way across multiple European languages remains challenging, especially due to the scarcity of language-parallel multilingual benchmarks.
We introduce a multilingual evaluation approach tailored for European languages.
We employ translated versions of five widely-used benchmarks to assess the capabilities of 40 LLMs across 21 European languages.
Our contributions include examining the effectiveness of translated benchmarks, assessing the impact of different translation services, and offering a multilingual evaluation framework for LLMs that includes newly created datasets: \emph{\acrshort{xMMLU}, \acrshort{xHella}, \acrshort{xARC}, \acrshort{xTQA}, and \acrshort{xGSM8K}}.
The benchmarks and results are made publicly available to encourage further research in multilingual LLM evaluation.
\end{abstract}

\section{Introduction}
The development of Large Language Models (LLMs) has led to transformative advancements to natural language processing, enabling complex tasks such as question-answering, summarization, and machine translation.
However, despite these advances, the evaluation of LLMs across languages\textemdash especially those beyond English and Chinese \textemdash remains a significant challenge.
Initiatives like CLEF\footnote{\url{https://clef2024.imag.fr/}}, FIRE\footnote{\url{http://fire.irsi.res.in/}}, and WMT\footnote{\url{https://www2.statmt.org/wmt24/}} have focused on multilingual evaluations, yet there is a lack of dedicated benchmarks that allow for consistent and comparable evaluation of LLM performance across different European languages.

Creating custom benchmarks for each language is a costly and time-consuming, which limits the scalability of evaluations and can result in a fragmented understanding of model performance across different languages.
Without comprehensive multilingual evaluations, comparisons between languages are often constrained, leading to an incomplete understanding of how models perform in languages beyond high-resource ones like English, German, or French.

To address these challenges, we investigate whether automatically translated benchmarks can reliably assess model performance across languages and whether this performance correlates with human evaluations, similar to benchmarks conducted in English.

In this paper, we focus on evaluating multilingual LLMs using translated versions of existing benchmarks.
Our goal is to determine the effectiveness of these translated benchmarks and assess whether they can substitute manually generated ones, allowing for a more scalable and uniform evaluation across multiple languages.

We make the following contributions:

\begin{itemize}
\item We evaluate the performance of 40 state-of-the-art LLMs for five multilingual datasets covering English and 20 additional languages in \Cref{sec:evaluation}.
\item We assess whether translated benchmarks correlate with human evaluations to asses their viability in \Cref{sec:human_pref}.
\item We compared translation services in \Cref{sec:okapi}, in particular DeeplPro and ChatGPT, to identify correlations in LLM performance and understand how translation quality impacts the models' benchmark results.
\item We release EU20-MMLU, EU20-HellaSwag, EU20-ARC, EU20-TruthfulQA, and EU20-GSM8K in \emph{20 European languages} as evaluation benchmarks for the NLP community.\footnote{\url{https://hf.co/collections/openGPT-X/eu20-benchmarks-67093b13db8ff192dc39312d}}.
\end{itemize}

\section{Related Work}\label{sec:rel_work}
Multilingual benchmarks have predominantly been generated using two widely adopted methodologies: either through human annotation to create datasets from scratch \cite{kocmi-etal-2023-findings,goyal-etal-2022-flores,conneau-etal-2018-xnli}, or by translating existing benchmarks using large language models (LLMs) \cite{lai-etal-2023-okapi,tiedemann-2012-parallel}.

While datasets developed by human annotators offer high-quality translations and task-specific accuracy, they require significant time and financial investment \cite{yang-etal-2019-paws}.
Furthermore, as models improve, these benchmarks can quickly become outdated, limiting their usefulness \cite{ott2022mapping,51569}. 

In contrast, the machine translation of existing benchmarks is more efficient and cost-effective, but its quality is often inconsistent, especially for medium- and low-resource languages, where translation inaccuracies can impact model evaluation \cite{meng2022generating,nllbteam2022language}.

These translation issues can lead to imbalanced evaluations across languages, particularly in languages with limited pre-training and evaluation resources \cite{goyal-etal-2022-flores,conneau-etal-2018-xnli}. 
Thus, selecting the right languages and ensuring translation quality is critical for creating a robust multilingual benchmark.

In addition to the choice of languages, it is crucial to consider the range of tasks that need to be covered for a benchmark.
Creating a multilingual benchmark for a specific task provides valuable insights into model performance for that exact task but says little about a model's general language capability \cite{conneau-etal-2018-xnli}.
This also includes benchmarks that, for example, cover the question-answering (QA) task for a wide range of domains, as only the QA task is tested \cite{lewis-etal-2020-mlqa}.

Another important factor is the range of tasks covered in a benchmark. 
Task-specific benchmarks provide insights into a model’s performance for specific tasks but do not fully capture a model’s overall language capabilities \cite{conneau-etal-2018-xnli,lewis-etal-2020-mlqa}. 
To address this, benchmarks that span multiple tasks, such as the OKAPI dataset \cite{lai-etal-2023-okapi}, offer a broader evaluation of model capabilities. 
OKAPI, for example, translates four diverse tasks (ARC, HellaSwag, MMLU and TruthfulQA) into 31 languages, including high-, medium-, and low-resource languages.

Building on this approach, we have expanded the linguistic reach through additional translations with special consideration on European languages and extend the variety of tasks. By including GSM8K~\cite{cobbe2021gsm8k} we offer evaluations on multilingual grade school math word problems.
Furthermore, we show that our data pre-processing has improved quality in certain language areas (See Section \ref{sec:experiment}).

\section{Experimental Setup}\label{sec:experiment}
Firstly, we describe in  \Cref{sec:translation} the translation process of the benchmarks, secondly \Cref{sec:evaluation_framework} details our evaluation framework and process.

\subsection{Translation Process}\label{sec:translation}
We chose DeepL as translation service due to its balance between translation accuracy and scalability.
We translated five well-known data\-sets, ARC~\cite{DBLP:journals/corr/abs-1803-05457}, HellaSwag ~\cite{DBLP:conf/acl/ZellersHBFC19}, TruthfulQA~\cite{lin-etal-2022-truthfulqa}, GSM8K~\cite{cobbe2021gsm8k}, and MMLU~\cite{hendrycks2020measuring} from English into 20 European languages.
These datasets encompass a mix of multiple-choice and open-ended generation tasks, each presenting unique translation challenges.
The translations preserved the original structure of each task to ensure  consistency across languages.
By leveraging DeepL’s XML tag handling, we avoided issues with prompt formatting and ensured that key contextual elements, such as 'Question' and 'Response' prompts, were appropriately localized (cf.~\Cref{fig:hellaswag_base}).
To enhance the visibility and accessibility of our benchmarks, we launched a multilingual leaderboard\footnote{\url{https://hf.co/spaces/openGPT-X/european-llm-leaderboard}} on Hugging Face, inspired by the popular Open LLM leaderboard\footnote{\url{https://hf.co/spaces/open-llm-leaderboard-old/open_llm_leaderboard}}.

\subsection{Evaluation Process} \label{sec:evaluation_framework}
We conducted the model evaluations on the HPC Cluster Alpha Centauri (TUD/ZIH)\footnote{\url{https://compendium.hpc.tu-dresden.de}}, the JUWELS Booster (JSC)\footnote{\url{https://apps.fz-juelich.de/jsc/hps/juwels/index.html}}, and, for larger models, on the JURECA Cluster\footnote{\url{https://westai.de/services/hardware/}}, using systems powered by NVIDIA A100 and H100 Tensor Core GPUs.
For model evaluation, we used adapted versions of the LM Evaluation Harness Framework from Eleuther.AI in versions v0.4.1 and v0.4.3, evaluating in a multi-GPU setting using hf/accelerate with both data and model parallelism.
For larger models, we deployed a local vLLM-based inference server (v0.5.3)\footnote{\url{https://github.com/vllm-project/vllm/releases/tag/v0.5.5}} using the LM Evaluation Harness local-completions implementation\footnote{commit 42dc24 LM Evaluation Harness} to perform pipeline-parallel evaluations across multiple GPUs on a single node.
We expended about 45,000 GPU hours conducting the model evaluations.

\section{Evaluation Results}
\label{sec:evaluation}

This section provides an overview of the performance of LLMs across various sizes, tasks and languages.
\Cref{sec:21_evaluation} focuses on the evaluation of all 21 European languages considered in this paper and investigates which models provide a competitive performance in a multilingual setting.
\Cref{sec:training_corpora} explores the relationship between the proportion of languages in the CommonCrawl dataset and the performance of the best current open-weights model.
Lastly, \Cref{sec:language_family} examines how model performance varies across different language families.
Since many models disclose little to no information about the composition of their training corpora, we focus our detailed analysis on the LLMs that exhibit the most consistent performance across the 21 languages.

\subsection{European LLM Evaluation}
\label{sec:21_evaluation}

This section provides an overview of the performance of several LLMs on various tasks and languages, grouped by their number of parameters in small-sized ($\leq$9B parameters), medium-sized ($>$9B \& $<$27B parameters) and large ($>$27B parameters) models.
The models are evaluated on our newly introduced tasks EU21-ARC, EU21-GSM8K, EU21-HellaSwag, EU21-MMLU, and EU21-TruthfulQA.
A comprehensive overview of all models are provided in the Appendix, including Tables (cf.~\Cref{tab:acc_std_21lang_1}, \ref{tab:acc_std_21lang_2}) and task specific heatmaps (cf.~\Cref{fig:ds-taskacc-EU21-ARC}, ~\ref{fig:ds-taskacc-EU21-HeSw},  ~\ref{fig:ds-taskacc-EU21-MMLU}, ~\ref{fig:ds-taskacc-EU21-TQA}, ~\ref{fig:ds-taskacc-EU21-GSM8k}).

\begin{figure}[ht]
    \centering
    \includegraphics[width=\linewidth]{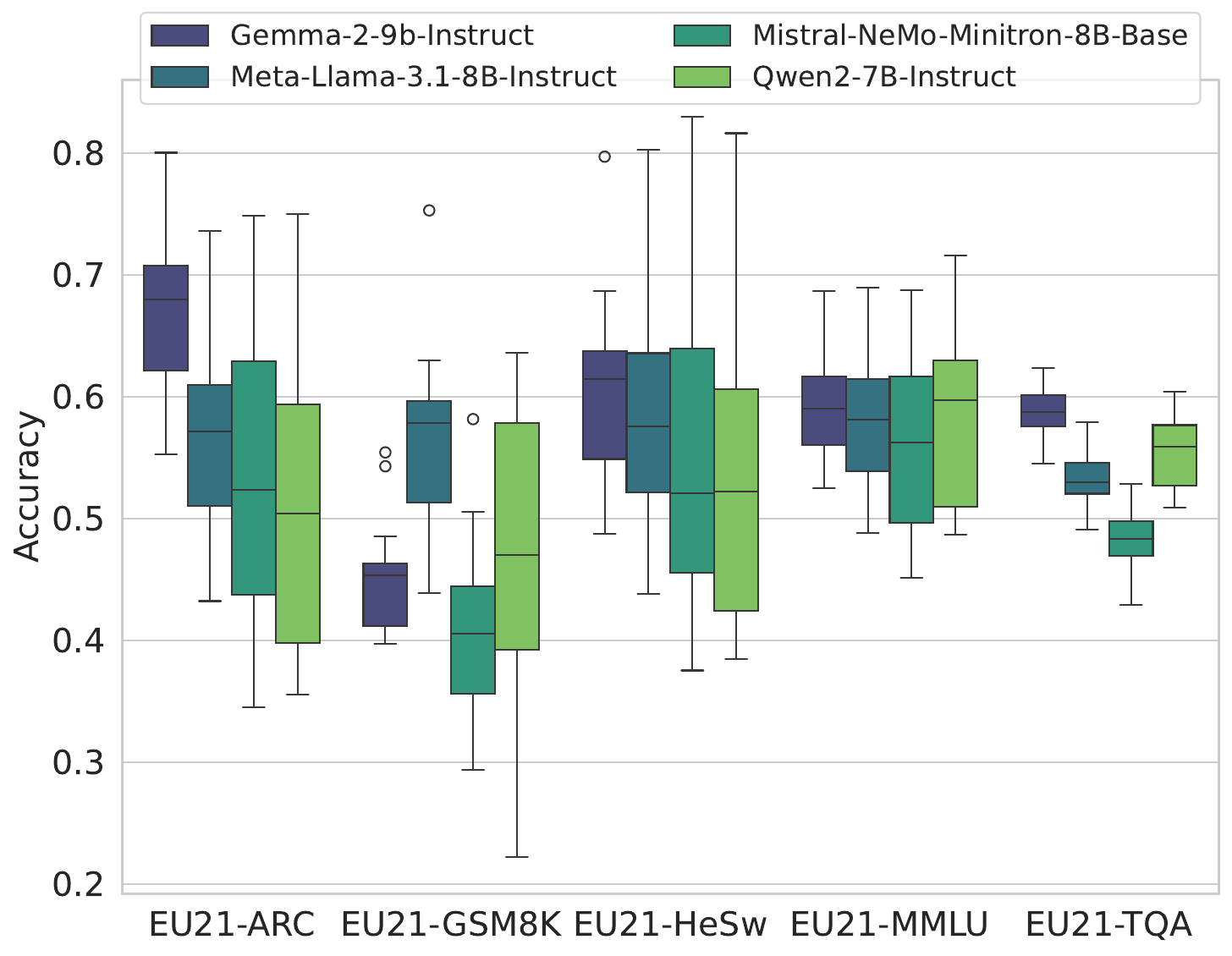}
    \caption{Accuracy results for four small sized LLMs on the EU21 tasks.}
    \label{fig:boxplot_small}
\end{figure}

\paragraph{Small-Sized Models ($\leq$9B Parameters)}
Among the small-sized models, the focus is on \emph{Gemma-2-9b-Instruct ~\cite{gemmateam2024gemma2improvingopen}, Meta-Llama-3.1-8B-Instruct ~\cite{dubey2024llama3herdmodels}, Mistral-NeMo-Minitron-8B-Base~\cite{sreenivas2024llmpruningdistillationpractice}, and Qwen2-7B-Instruct~\cite{yang2024qwen2technicalreport}}. Models with a lower average performance, like Gemma-7b~\cite{gemmateam2024gemmaopenmodelsbased}, are listed in Appendix Tables \ref{tab:acc_std_21lang_1}, \ref{tab:acc_std_21lang_2}.

Gemma-2-9b-Instruct performs consistent across languages (cf.~\Cref{fig:boxplot_small}), as indicated by its narrow interquartile range, but lags behind Meta-Llama-3.1-8B-Instruct on the EU21-GSM8K benchmark, signaling lower math abilities.
In EU21-MMLU, the model shows a robust behaviour, but the median is slightly lower than Qwen2-7B-Instruct.
The results on the latter two benchmarks indicate that the capacity of small models might not allow for a reliable performance on all languages \emph{and} a good performance on specialized knowledge.

Meta-Llama-3.1-8B-Instruct excels in math-related tasks, achieving the highest median performance in EU21-GSM8K, but its performance lags behind Gemma-2-9b-Instruct across all 21 languages.

Mistral-NeMo-Minitron-8B-Base shows competitive performance in widely spoken languages but struggles in less-resourced languages and complex tasks. 
Its performance on EU21-GSM8K and EU21-MMLU is notably lower than the other models.

Qwen2-7B-Instruct demonstrates the highest median accuracy in EU21-MMLU, but performs worse in EU21-HellaSwag and EU21-ARC.
In EU21-GSM8k the model performs similar to Mistral-NeMo-Minitron-8B-Base and Meta-Llama-3.1-8B-Instruct.

\paragraph{Medium-Sized Models ($>$9B \& $<$27B Parameters)}
\begin{figure}[ht]
    \centering
    \includegraphics[width=\linewidth]{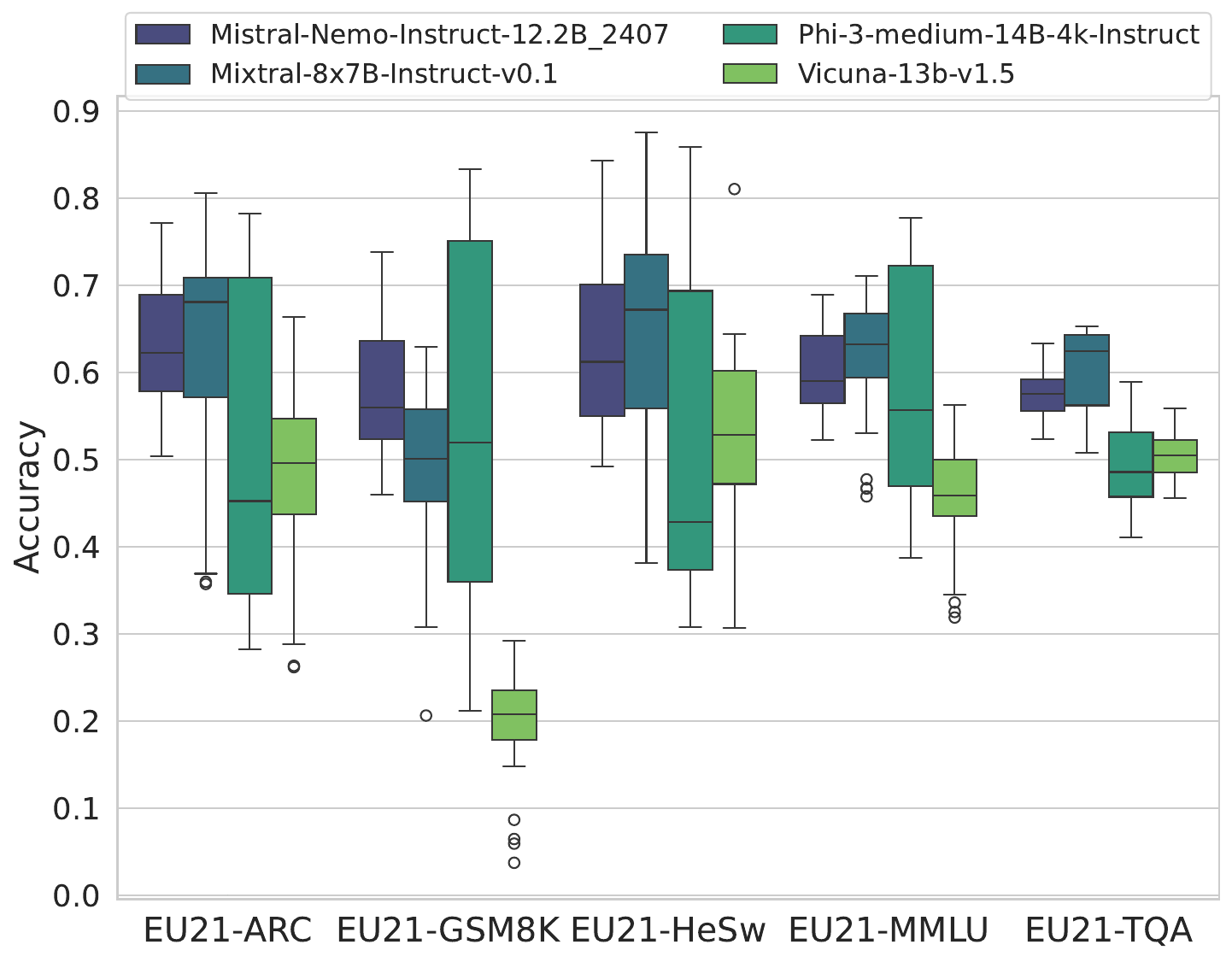}
    \caption{Accuracy results for four medium sized LLMs on the EU21 tasks.}
    \label{fig:boxplot_medium}
\end{figure}
This analysis (cf.~\Cref{fig:boxplot_medium}) focuses on medium-sized LLMs such as \emph{Mistral-Nemo-Instruct-12.2B\footnote{\url{https://hf.co/mistralai/Mistral-Nemo-Instruct-2407}}, Mixtral-8x7B-Instruct-v0.1~\cite{jiang2024mixtralexperts}, Phi-3-medium-14B-4k-Instruct~\cite{abdin2024phi3technicalreporthighly} and Vicuna-13b-v1.5~\cite{peng2023instructiontuninggpt4}}.

Mixtral-8x7B-Instruct-v0.1 shows the highest median performance on four out of the five tasks.
Only on EU21-GSM8K its performance is lacking behind Mistral-Nemo-Instruct-12.2B. %
However, its performance tends to be more variable across languages than Mistral-Nemo-Instruct-12.2B\_240, indicated by the low minima in 4 out of 5 tasks.

Mistral-Nemo-Instruct-12.2B\_2407 delivers solid results across all tasks with a particularly strong median performance in the EU21-GSM8K task. As evidenced by its narrow interquartile ranges, the model signals low variability across all 21 languages compared to the other models.

Phi-3-medium-14B-4k-Instruct presents competitive results in English by having the highest GSM8K and MMLU values, but lags vastly behind the previous two models in median peformance across all languages and tasks, indicating a focus on a few selected languages in the training corpora.
Lastly, Vicuna-13-b-v1.5 based on the older LLama-2-13B is not competitive on any task, due to its limited two trillion English-centric training tokens.

Compared to the small-sized category, Mixtral-8x7B-Instruct-v0.1 provides a substantial gain in EU21-HellaSwag by achieving an median performance of 0.64, an absolute increase of 4\% compared to Gemma-2-9b-Instruct.
On EU21-GSM8K Mistral-Nemo-Instruct-12.2B achieves 57\% accuracy a small increase of 1\% compared to Meta-Llama-3.1-8B-Instruct.
On EU21-MMLU Qwen2-7B achieved a accuracy of 59.4\%, slightly behind medium sized models such as Mistral-Nemo-Base-12.2B with 60.2\% and Mixtral-8x7B-Instruct-v0.1 with an performance of 60.8\%.

Overall the medium category provides only small task dependend gains with specific models compared to the smaller Gemma-2-9b-Instruct model, indicating its strength in multilinguality.

\paragraph{Large-Sized Models ($\geq$27B Parameters)}
\begin{figure}[ht]
    \centering
    \includegraphics[width=\linewidth]{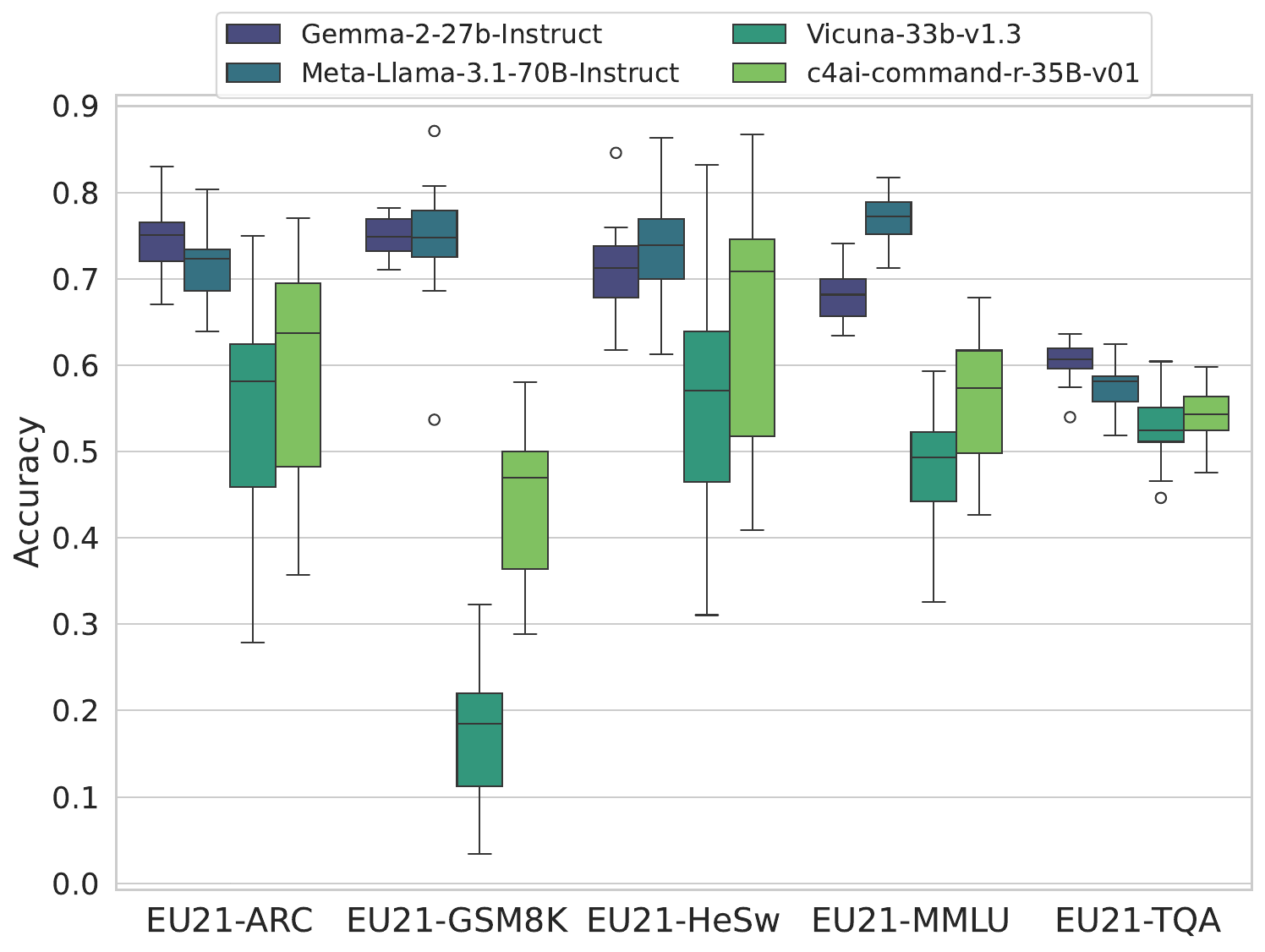}
    \caption{Accuracy results for four large sized LLMs on the EU21 tasks.}
    \label{fig:boxplot_large}
\end{figure}
This analysis (cf.~\Cref{fig:boxplot_large}) focuses on large-sized LLMs such as \emph{Gemma-2-27b-Instruct \cite{gemmateam2024gemma2improvingopen}, Vicuna-33b-v1.3~\cite{peng2023instructiontuninggpt4}, Meta-Llama-3.1-70B-Instruct \cite{dubey2024llama3herdmodels} and c4ai-command-r-35B-v01\footnote{\url{https://hf.co/CohereForAI/c4ai-command-r-v01}}}.

Gemma-2-27b-Instruct and Meta-Llama-3.1-70B-Instruct reliable outperforms the other models across most tasks, with median scores often exceeding an accuracy of 0.7.
Their performance is not only high but also stable across different languages, as evidenced by its narrow interquartile ranges, signaling low variability.
We assume that due to the higher capacity (compared to Gemma-2-9b), Gemma-2-27b-Instruct now also scores higher on specialized knowledge tasks such as EU21-MMLU and EU21-GSM8K.

Comparable, to the smaller Vicuna model also the 33 billion parameter version is not competitive on any task, due to its limited two trillion English-centric training tokens.
c4ai-command-r-35B-v01 performs better in English compared to the Vicuna model.

Especially Gemma-2-27b-Instruct convinces with a remarkable performance.
It provides a vast improvements over the small and medium sized LLMs and is even competitive with the more than twice as large Meta-LLama-3.1 70B model.

\subsection{Impact of available training corpora}\label{sec:training_corpora}
To better analyze the performance of LLMs across different languages, we categorized our target languages into high-resource (HRL) and medium-resource (MRL) groups, based on their proportion in the CommonCrawl dataset, which serves as the primary pre-training source for many LLMs.
High-resource languages (HRL) have a CommonCrawl representation of 1\% or more, whereas medium-resource languages (MRL) have between 0.01\% and 1\%.
The language statistics accompanied with evaluation results of the LLama-3.1-70B-Instruct \cite{dubey2024llama3herdmodels} model is depicted in \Cref{tab:language-statistics}.

There is a downwards trend with outliers in Swedish and Danish, both Germanic languages that might profit from English or other Germanic training corpora (cf.~\Cref{tab:language-statistics}).
The performance of EU21-TQA, EU21-MMLU, and EU21-GSM8K shows a slight decline in correlation with the number of speakers in millions and the percentage share of CommonCrawl data across different languages, while EU21-ARC and EU21-Hellaswag exhibit a more pronounced performance drop.

Both EU21-ARC and EU21-Hellaswag are benchmarks that focus on reasoning, commonsense knowledge, and context understanding. These tasks might require more language specific knowledge, which may not be as well represented in languages with fewer speakers or less CommonCrawl data. In contrast, tasks like MMLU and GSM8K, while challenging, may rely more on factual recall, basic math, or language understanding. We hypothesize these are capabilities a model can easier transfer across languages.

\begin{table*}
\centering
\begin{tabular}{lllcccccccc}
\hline
      &                                                                             &                                                                                          &      &      & EU21- & EU21- & EU21- & EU21- & EU21- \\
Lang. & Speak. M\footnote{\url{https://www.ethnologue.com/insights/ethnologue200/}} & CC(\%)\footnote{\url{https://commoncrawl.github.io/cc-crawl-statistics/plots/languages}} & Cat. & Avg. &  ARC  & GSM8K & HeSw  & MMLU  & TQA\\
\hline
English & 1456 & 45.05 & HRL & .79 & .80 & .87 & .86 & .82 & .60 \\
Spanish  & 559 & 4.50 & HRL & .73 &.76 & .73 & .79 & .80 & .56\\
French & 310 & 4.22 & HRL & .72 & .75& .75& .78& .78 & .56\\
Port. & 264 & 2.02 & HRL & .73 &.73&.78&.78&.78&.57\\
German& 133 & 5.47 & HRL & .71&.73&.74&.75&.78&.57 \\
Italian & 68 & 2.64 & HRL & .73 &.76&.76&.77&.79&.58\\
Polish & 40.6 & 1.68 & HRL & .70 &.72&.69&.74&.78&.59  \\
Dutch & 24.5 & 1.83 & HRL & .73 &.73&.80&.76&.78&.60\\
Czech& 12.3 & 1.09 & HRL & .71 &.72&.75&.72&.78&.58 \\
\hline
Romanian & 24.5 & 0.52 & MRL & .72 &.73&.73&.74&.79&.59 \\
Greek & 13.1 & 0.55 & MRL & .70 &.69 &.78 &.71&.74&.58\\
Swedish & 13.1 & 0.63 & MRL & .75 &.74&.81&.78&.78&.62\\
Hungarian & 12.6 & 0.56 & MRL & .70 &.70&.75&.72&.77&.59 \\
Bulgarian & 7.7 & 0.29 & MRL & .66 &.76&.76&.70&.67&.62 \\
Slovak & 7.3 & 0.37 & MRL & .69 &.75&.74&.70&.68&.60 \\
Danish & 5.6 & 0.45 & MRL& .73& .72&.79&.75&.79&.59\\
Finnish & 5.6 & 0.35 & MRL & .68 &.67&.77&.71&.75&.52\\
Lithuanian & 3.8 & 0.16 & MRL & .65 &.72&.72&.66&.64&.60 \\
Slovenian & 2.9 & 0.15 & MRL& .67&.68&.72&.66&.74&.55 \\
Latvian & 1.5 & 0.12 & MRL& .66 & .64&.71&.61&.73&.58 \\
Estonian & 1.1 & 0.14 &MRL & .66 &.64&.76&.64&.74&.52\\
\hline
\end{tabular}
\caption{Language statistics table sorted by number of speakers (approximate total speakers worldwide), and the accuracy score for the LLama-3.1 70B-Instruct model in each task and language.}
\label{tab:language-statistics}
\end{table*}

\subsection{Influence of Language Family}\label{sec:language_family}
In this subsection we investigate how the LLM performance depend on the language family.
We categorize the larger language groups based on their corresponding language family like the following:

\begin{itemize}
    \item Germanic: Danish, German, English, Dutch, Swedish (in total 53.38\% of CC)
    \item Romance: Spanish, French , Italian, Portuguese, Romanian (in total 13.9\% of CC)
    \item Slavic: Bulgarian, Czech, Polish, Slovak, Slovenian (in total 3.58\% of CC)
    \item Not categorized: Greek (hellenic), Hungarian (ugric), Estonian (finnic), Finnish (finnic), Lithuanian (baltic),  Latvian (baltic)
\end{itemize}

Examining the average accuracies (cf.~Appendix~\Cref{tab:acc_std_lang_group_germanic_1}, \ref{tab:acc_std_lang_group_germanic_2}, \ref{tab:acc_std_lang_group_romance_1}, \ref{tab:acc_std_lang_group_romance_2} \ref{tab:acc_std_lang_group_slavic_1}, \ref{tab:acc_std_lang_group_slavic_2}) across the three language families, we observe that models generally achieve higher performance on Romance and Germanic languages compared to Slavic languages.
For instance, the~\textit{Gemma-2-27b-Instruct}~\cite{gemmateam2024gemma2improvingopen} model attains average accuracies of 0.727 on Germanic languages (cf.~\Cref{tab:acc_std_lang_group_germanic_1}), 0.716 on Romance languages (cf.~\Cref{tab:acc_std_lang_group_romance_1}), and 0.694 on Slavic languages (cf.~\Cref{tab:acc_std_lang_group_slavic_1}).

Similarly, the \textit{Meta-Llama-3.1-70B-Instruct} \cite{dubey2024llama3herdmodels} model achieves average accuracies of 0.745 (Germanic), 0.726 (Romance), and 0.680 (Slavic), reinforcing the trend of lower performance on Slavic languages.

In the following we compare the model performance on similar sized (in CC \%) languages such as Romanian (Romance, 0.52\%), Swedish (Germanic, 0.63\%) and Polish (Slavic, 1.09\%).
\Cref{tab:language-statistics} indicates that even though, Swedish is less represented than Polish, it achieves higher performance across all tasks.
Also Romanian as the smallest represented language achieves higher performance across all five tasks, indicating that the language family is of influence in the performance analysis.

\section{Correlation with Human Preferences}\label{sec:human_pref}
We examine the correlation between the introduced EU21 benchmarks and the Elo scores from the LMSYS Chatbot Arena\footnote{\url{https://chat.lmsys.org/?leaderboard}} to better understand how human preferences are reflected.

In the LMSYS Chatbot Arena~\cite{Chiang:2024}, users can vote on a pair of LLM responses in anonymized and randomized "battles".
The resulting preference data are used to compute Elo rankings published on the LMSYS chatbot arena leaderboard.
There are multiple independently computed rankings in different  categories, some of which correspond to the languages of the conversations.
We utilize an input dataset with timestamp \texttt{20240814} containing about 1.8 million human quality judgements provided for reproducing the rankings (cf.~\Cref{appendix:lmsys:battles_overview}).
With these data, we compute the Elo scores using FastChat\footnote{commit \texttt{e5dc446f}}.

The LLMs for the subsequent correlation analysis were mostly chosen using the following criteria:
\begin{itemize}
    \item Text-only, Causal Transformer-based LM
    \item Checkpoints are publicly available on HF Hub
    \item Model Size below 100B parameters
\end{itemize}

To quantify how well the accuracies of a given model on our translated datasets correlate with the Elo scores, Then we compute the Pearson and Spearman Correlation Coefficients between the accuracy and the Elo scores of \(n = 17\) models using \texttt{scipy.stats} with default settings and confidence level 95\%.
We compute these correlations for Czech, French, German, Italian, Polish, Portuguese, Spanish, as for these languages, at least 50\% of the Elo ranks of the models we consider can be distinguished with about 95\% confidence, and for English as a reference point.
The remaining mostly smaller medium-resource languages have a high uncertainty in their Elo scores as fewer battles (cf.~\Cref{tab:battle_overview})  are available.

\begin{table}[htbp]
\centering
\begin{tabular}{lll}
\toprule
Lang. & Pearson       & Spearman        \\
\midrule
CS & .570**     & .430*       \\
EN & .716***     & .618**       \\
FR & .770***     & .745***      \\
DE & .775***    & .740***      \\
IT & .694***     & .674***   \\
PL & .895***    & .858***     \\
PT & .811***    & .806***      \\
ES & .751***    & .784***    \\
\bottomrule
\end{tabular}
\caption{Correlation Coefficients for Task Avg.~between Few-Shot Accuracy and LMSYS Elo Score. Significance levels: *\emph{p}~<~0.05, **\emph{p}~<~0.01, ***\emph{p}~<~0.001.}
\label{tab:lmsys_pearson_alltasks_few_shot_main}
\end{table}

The \acrshort{xARC} dataset exhibits strong positive correlations between model accuracy and Elo scores across all languages (cf.~\Cref{tab:lmsys_pearson_arc_few_shot}).
Pearson correlation coefficients are highest in Polish ($r = .868$), followed by German ($r = .815$) and Portuguese ($r = .779$), the correlation is lowest for Czech ($r = .600$).
Spearman correlations align with this trend, indicating a robust relationship between Elo scores and performance on reasoning tasks in multilingual contexts.

Significant positive correlations are also observed for the \acrshort{xGSM8K} dataset (cf.~\Cref{tab:lmsys_pearson_gsm8k_few_shot}), especially in Portuguese (Pearson $r = .786$) and Polish ($r = .785$, $p = .0001$).
These results suggest that higher Elo scores are associated with better performance on mathematical problem-solving tasks in these languages.

For \acrshort{xHella} (cf.~\Cref{tab:lmsys_pearson_hellaswag_few_shot}), which assesses commonsense reasoning, strong correlations are evident in the non-English languages.
Pearson coefficients are $r = .851$ for Polish, $r = .729$ for Portuguese, and $r = .722$ for French.
In English, the correlations are lower (Pearson $r = .430$, $p = .0428$) but still statistically significant, indicating that Elo scores may be better predictors of performance on commonsense reasoning tasks in other languages.
Another cause could be contaminated training datasets.

MMLU (cf.~\Cref{tab:lmsys_pearson_mmlu_few_shot}) shows the highest correlations among all tasks, particularly in Polish (Pearson $r = .885$) and French ($r = .814$).
The high Spearman coefficients (French: $\rho = .804$, Polish: $\rho = .863$) indicate that Elo scores are excellent predictors of a model's broad knowledge and reasoning capabilities in these languages.

The \acrshort{xTQA} (cf.~\Cref{tab:lmsys_pearson_truthfulqa_few_shot}) dataset exhibits lower correlations compared to other tasks, with some not reaching statistical significance, particularly in French (Pearson $r = .360$, $p = .07793$) and German ($r = .343$, $p = .08872$).
In Spanish, the correlations are moderate (Pearson $r = .447$, $p = .03615$).
This suggests that Elo scores may be less effective at predicting performance on tasks requiring truthful responses.

We also investigated the average across all five tasks (cf.~\Cref{tab:lmsys_pearson_alltasks_few_shot_main}).
The high correlations across all languages highlight the strong relationship between Elo scores and the average benchmark accuracies.
The highest Pearson and Spearman correlation is in Polish ($r = .895$, $\rho = .858$).
Portuguese shows similar values ($r = .811$, $\rho = .806$), while English exhibits lower correlations (Pearson $r = .716$, $p = .00062$; Spearman $\rho = .618$, $p = .00412$).
This suggests that the correlation between LMSYS Elo scores and our average benchmark results are similar to English in the most languages.
For Czech we reached the lowest correlation ($r = .570$, $\rho = .430$) but even there we still have statistical correlations.

\section{Influence of the translation service}\label{sec:okapi}

While OKAPI~\cite{lai-etal-2023-okapi} and our evaluation datasets share a similar structure, they differ in key aspects such as translation methodology and sample alignment.
The most notable difference between the two datasets is the translation approach:
Our dataset was translated using \textit{DeepL}, whereas OKAPI employed \textit{ChatGPT} for this task.
The translations for the OKAPI tasks cover 31 languages, including both EU and non-EU languages.
For our evaluation datasets, we focused on translations into 20 European languages.
The common and new set of EU languages between the OKAPI and our benchmarks are shown in \Cref{tab:okapi_opengptx}.

\begin{table}[ht]
\centering
\begin{tabular}{l|l}
\hline
\textbf{Common Languages}  & \textbf{New Languages} \\ \hline
Danish          & Czech      \\
German          & Greek      \\
Spanish         & Estonian   \\
French          & Finnish    \\
Hungarian       & Lithuanian \\
Italian         & Latvian    \\
Dutch           & Bulgarian  \\
Romanian        & Polish     \\
Slovak          & Slovenian  \\
Swedish         &            \\
\end{tabular}
\caption{Comparison of languages between OKAPI and our Datasets}
\label{tab:okapi_opengptx}
\end{table}

In this section, we present a correlation analysis with a focus on the \acrshort{xARC}, \acrshort{xHella}, and \acrshort{xMMLU} tasks, alongside a COMET score \cite{rei-etal-2023-scaling} evaluation to assess the translation quality across both datasets.

The idea behind the correlation analysis is to record\footnotemark whether the model responded correctly or incorrectly for each OKAPI sample and sample from our dataset and to examine if the predictions are correlated, allowing direct performance comparison on sample-level on the same data.
\footnotetext{The evaluation was performed using the LM-Evaluation-Harness framework, with sample-level logging enabled.}
For a sample-level comparison, we ensured that our examples and OKAPI samples are matched and can be linked to their original translations.
While \acrshort{xARC}, \acrshort{xHella}, and \acrshort{xMMLU} samples have IDs matching the original English subsets, enabling mapping, this does not apply to \acrshort{xTQA}.

For the correlation analysis, we calculated the correlation matrices for different models, tasks, and languages, listed in \Cref{appendix:okapi:correlations}.

The highest correlations, as shown in \Cref{tab:model_performance_sorted}, are with Aya-23-8B (0.57), Meta-Llama-3-8B-Instruct (0.56), and gemma-7b (0.56), while bloom-7b1 (0.40) and bloomz-7b1 (0.44) show lower correlations.
This indicates that Aya and Meta-Llama perform more consistently, while bloom-7b1 and bloomz-7b1 exhibit greater variability.

As seen in \Cref{tab:task_correlation}, RC has the highest correlation at 0.61, followed by HellaSwag and MMLU (both 0.54).
ARC’s structure and reliance on factual knowledge likely make model performance more stable and predictable.

\Cref{tab:language_correlation} shows that correlations range from 0.44 for Hungarian to 0.59 for Spanish.
Hungarian’s lower score may reflect fewer training resources, while Spanish likely benefits from better data availability, leading to more consistent performance.

Translation quality likely plays a role in these results, as ambiguous or incorrect translations, poorly adapted prompts, or task mismatches can negatively affect model performance.
Thus, the observed correlations may also reflect the degree to which translation quality impacts each language's evaluation results.
Examples illustrating translation quality issues in the ten common languages can be found in \Cref{tab:EU21_okapi_hellaswag} for the HellaSwag task.

As part of the further analysis, we determined the contingency table (\Cref{tab:crosstab}) to compare the answers of the models on our evaluation datasets and the OKAPI datasets.
Both evaluations show similar alignment, with agreements on correct answers (38.1\%) and incorrect answers (39.3\%) being fairly balanced.
However, OKAPI seems to perform slightly better in cases of disagreement, as OKAPI is correct and our model is incorrect in 11.8\% of cases, compared to 10.8\% where our model is correct and OKAPI is incorrect.
This suggests that, while the performance of both models is comparable, OKAPI might have a slight edge in terms of accuracy in cases where they disagree.

\begin{table}[ht]
\centering
\begin{tabular}{l|c c}
\textbf{}               & \textbf{OKAPI Corr.} & \textbf{OKAPI Inc.} \\\hline
\textbf{Ours Corr.}     & 0.3808               & 0.1183              \\
\textbf{Ours Inc.}      & 0.1077               & 0.3929              \\
\end{tabular}
\caption{Contingency table comparing the answers of the models on the evaluations performed on our translations and the OKAPI translations.}
\label{tab:crosstab}
\end{table}

Further we used the COMET score \cite{rei-etal-2023-scaling} to analyze the translation quality of both datasets.
Since there is no target translation for the different benchmarks, we evaluated the translation quality of the our and OKAPI datasets using the COMET-KIWI metric \cite{rei-etal-2023-scaling}.
This metric uses a pre-trained language model to predict translation quality.
A score of 1 indicates a high-quality translation and a score close to 0 indicates a bad translation.
Both our and OKAPI datasets have a relatively high average score of 0.8208 and 0.8190, respectively, across all tasks and translations.
The exact scores can be found in the Appendix in Table \Cref{tab:comet-kiwi_score}.
It is important to highlight that because there is no target translation, no statistical significance can be calculated between the scores of OKAPI and ours.
Therefore, we cannot conclude that our translations are automatically higher quality based on the COMET KIWI score.
Nevertheless, it can be shown that our datasets have a consistently high translation quality across all languages (cf. Table \Cref{tab:comet-kiwi_score}).

We additionally compared all our LLMs on the ARC, (\Cref{fig:okapi-arc}), HellaSwag (\Cref{fig:okapi-hellaswag}) and MMLU benchmarks (\Cref{fig:okapi-mmlu}).

Overall we can see that all evaluated LLMs achieve higher accuracies on our tasks compared to OKAPI.
This analysis highlights the extent to which our and OKAPI results align across models, tasks, and languages, providing insights into the consistency and reliability of model evaluations across datasets.

\section{Conclusion}\label{sec:conclusion}

This work provides a comprehensive framework for evaluating Large Language Models (LLMs) across multiple European languages by utilizing translated versions of widely-used benchmarks.
Our experiments demonstrate that translated benchmarks can serve as a reliable proxy for human preference in various languages.
We also highlight the influence of translation services on the quality of the benchmarks.
By releasing our benchmarks, we aim to foster further research and development in multilingual LLM evaluation, driving improvements in cross-lingual NLP applications.
Future work will focus on refining cultural localization in translations and exploring deeper correlations between translation quality and model performance across diverse language families.

\clearpage

\section{Limitations}
It depends much on the task whether translations are a good way to create corpora in parallel languages. The ambiguity resolution benchmark WinoGrande \cite{winogrande} for instance is posed as a multiple choice task, where a object or subject in a sentence is masked and each choice constitutes a replacement of the mask.
The model is then tasked with selecting the sentence that is semantically correct.
\Cref{fig:winogrande_base} and \ref{fig:winogrande_spa} in the Appendix demonstrate how in languages with gendered nouns or gendered adjectives in predicative use, the benchmark becomes easier due to strong grammatical signals indicating the correct choice of subject or object.
Depending on the benchmark, some sequences were sporadically not translated, so some answer options are either in the original language (English) or are left blank. We detected the issue by analyzing the translated sequences with unexpectedly low COMET KIWI scores (<0.5) during the quality analysis of the translations. The maximum error rate for any language (assuming no errors for scores~$\geq 0.5$) is 2.3\% for HellaSwag (IT), 1.8\% for ARC (ET), and 2.1\% for MMLU (PT) across all languages.

Variations in accuracy were noted when testing language models with different versions of the LM Eval Harness framework (v0.4.1, v0.4.3, and 42dc24/main).
The differences (less than 1\%), affected ARC and MMLU, were attributed mainly to changes in batch size, and task formatting. These variations emphasize the importance of maintaining consistent evaluation practices to ensure accurate model comparisons across different versions. Details can be found in \cref{appendix:evaluation:accvar}.

\section{Ethical \& Broader Impact}

The ability to evaluate large language models (LLMs) across a wide range of European languages, particularly underrepresented ones, is a critical step toward enhancing inclusivity and accessibility in natural language processing (NLP). By ensuring that LLMs can perform well in languages beyond English or other high-resource languages, we contribute to a more equitable digital landscape where speakers of less widely spoken languages have equal access to advanced language technologies. However, this inclusivity brings unique challenges, particularly in achieving benchmarks that are comparable across diverse linguistic and cultural contexts.

\section*{Acknowledgments}
This work was funded by the Federal Ministry of Education and Research of Germany and the state of North-Rhine Westphalia as part of the Lamarr-Institute for Machine Learning and Artificial Intelligence, LAMARR22B as well as by the German Federal Ministry for Economic Affairs and Climate Action (BMWK) through the project OpenGPT-X (project no. 68GX21007D) and by the European Union’s Horizon 2020 research and innovation program under grant agreement No 101135671 (TrustLLM).
The authors gratefully acknowledge the Gauss Centre for Supercomputing e.V. (www.gauss-centre.eu) for funding this project by providing computing time on the GCS Supercomputer JUWELS at Jülich Supercomputing Centre (JSC) as well as the Center for Information Services and High Performance Computing [Zentrum für Informationsdienste und Hochleistungsrechnen (ZIH)] at TU Dresden for providing its facilities for automatic evaluation computations.

\bibliography{literature}

\begin{thebibliography}{33}
\providecommand{\natexlab}[1]{#1}

\bibitem[{Abdin et~al.(2024)Abdin, Aneja, Awadalla, Awadallah, Awan, Bach, Bahree, Bakhtiari, Bao, Behl, Benhaim, Bilenko, Bjorck, Bubeck, Cai, Cai, Chaudhary, Chen, Chen, Chen, Chen, Chen, Cheng, Chopra, Dai, Dixon, Eldan, Fragoso, Gao, Gao, Gao, Garg, Giorno, Goswami, Gunasekar, Haider, Hao, Hewett, Hu, Huynh, Iter, Jacobs, Javaheripi, Jin, Karampatziakis, Kauffmann, Khademi, Kim, Kim, Kurilenko, Lee, Lee, Li, Li, Liang, Liden, Lin, Lin, Liu, Liu, Liu, Liu, Liu, Luo, Madan, Mahmoudzadeh, Majercak, Mazzola, Mendes, Mitra, Modi, Nguyen, Norick, Patra, Perez-Becker, Portet, Pryzant, Qin, Radmilac, Ren, de~Rosa, Rosset, Roy, Ruwase, Saarikivi, Saied, Salim, Santacroce, Shah, Shang, Sharma, Shen, Shukla, Song, Tanaka, Tupini, Vaddamanu, Wang, Wang, Wang, Wang, Wang, Wang, Ward, Wen, Witte, Wu, Wu, Wyatt, Xiao, Xu, Xu, Xu, Xue, Yadav, Yang, Yang, Yang, Yang, Yu, Yuan, Zhang, Zhang, Zhang, Zhang, Zhang, Zhang, Zhang, and Zhou}]{abdin2024phi3technicalreporthighly}
Marah Abdin, Jyoti Aneja, Hany Awadalla, Ahmed Awadallah, Ammar~Ahmad Awan, Nguyen Bach, Amit Bahree, Arash Bakhtiari, Jianmin Bao, Harkirat Behl, Alon Benhaim, Misha Bilenko, Johan Bjorck, Sébastien Bubeck, Martin Cai, Qin Cai, Vishrav Chaudhary, Dong Chen, Dongdong Chen, Weizhu Chen, Yen-Chun Chen, Yi-Ling Chen, Hao Cheng, Parul Chopra, Xiyang Dai, Matthew Dixon, Ronen Eldan, Victor Fragoso, Jianfeng Gao, Mei Gao, Min Gao, Amit Garg, Allie~Del Giorno, Abhishek Goswami, Suriya Gunasekar, Emman Haider, Junheng Hao, Russell~J. Hewett, Wenxiang Hu, Jamie Huynh, Dan Iter, Sam~Ade Jacobs, Mojan Javaheripi, Xin Jin, Nikos Karampatziakis, Piero Kauffmann, Mahoud Khademi, Dongwoo Kim, Young~Jin Kim, Lev Kurilenko, James~R. Lee, Yin~Tat Lee, Yuanzhi Li, Yunsheng Li, Chen Liang, Lars Liden, Xihui Lin, Zeqi Lin, Ce~Liu, Liyuan Liu, Mengchen Liu, Weishung Liu, Xiaodong Liu, Chong Luo, Piyush Madan, Ali Mahmoudzadeh, David Majercak, Matt Mazzola, Caio César~Teodoro Mendes, Arindam Mitra, Hardik Modi, Anh Nguyen,
  Brandon Norick, Barun Patra, Daniel Perez-Becker, Thomas Portet, Reid Pryzant, Heyang Qin, Marko Radmilac, Liliang Ren, Gustavo de~Rosa, Corby Rosset, Sambudha Roy, Olatunji Ruwase, Olli Saarikivi, Amin Saied, Adil Salim, Michael Santacroce, Shital Shah, Ning Shang, Hiteshi Sharma, Yelong Shen, Swadheen Shukla, Xia Song, Masahiro Tanaka, Andrea Tupini, Praneetha Vaddamanu, Chunyu Wang, Guanhua Wang, Lijuan Wang, Shuohang Wang, Xin Wang, Yu~Wang, Rachel Ward, Wen Wen, Philipp Witte, Haiping Wu, Xiaoxia Wu, Michael Wyatt, Bin Xiao, Can Xu, Jiahang Xu, Weijian Xu, Jilong Xue, Sonali Yadav, Fan Yang, Jianwei Yang, Yifan Yang, Ziyi Yang, Donghan Yu, Lu~Yuan, Chenruidong Zhang, Cyril Zhang, Jianwen Zhang, Li~Lyna Zhang, Yi~Zhang, Yue Zhang, Yunan Zhang, and Xiren Zhou. 2024.
\newblock \href {https://arxiv.org/abs/2404.14219} {Phi-3 technical report: A highly capable language model locally on your phone}.
\newblock \emph{Preprint}, arXiv:2404.14219.

\bibitem[{Aryabumi et~al.(2024)Aryabumi, Dang, Talupuru, Dash, Cairuz, Lin, Venkitesh, Smith, Campos, Tan, Marchisio, Bartolo, Ruder, Locatelli, Kreutzer, Frosst, Gomez, Blunsom, Fadaee, Üstün, and Hooker}]{aryabumi2024aya23openweight}
Viraat Aryabumi, John Dang, Dwarak Talupuru, Saurabh Dash, David Cairuz, Hangyu Lin, Bharat Venkitesh, Madeline Smith, Jon~Ander Campos, Yi~Chern Tan, Kelly Marchisio, Max Bartolo, Sebastian Ruder, Acyr Locatelli, Julia Kreutzer, Nick Frosst, Aidan Gomez, Phil Blunsom, Marzieh Fadaee, Ahmet Üstün, and Sara Hooker. 2024.
\newblock \href {https://arxiv.org/abs/2405.15032} {Aya 23: Open weight releases to further multilingual progress}.
\newblock \emph{Preprint}, arXiv:2405.15032.

\bibitem[{Chiang et~al.(2024)Chiang, Zheng, Sheng, Angelopoulos, Li, Li, Zhang, Zhu, Jordan, Gonzalez, and Stoica}]{Chiang:2024}
Wei-Lin Chiang, Lianmin Zheng, Ying Sheng, Anastasios~Nikolas Angelopoulos, Tianle Li, Dacheng Li, Hao Zhang, Banghua Zhu, Michael Jordan, Joseph~E. Gonzalez, and Ion Stoica. 2024.
\newblock \href {https://arxiv.org/abs/2403.04132} {Chatbot arena: An open platform for evaluating llms by human preference}.
\newblock \emph{Preprint}, arXiv:2403.04132.

\bibitem[{Clark et~al.(2018)Clark, Cowhey, Etzioni, Khot, Sabharwal, Schoenick, and Tafjord}]{DBLP:journals/corr/abs-1803-05457}
Peter Clark, Isaac Cowhey, Oren Etzioni, Tushar Khot, Ashish Sabharwal, Carissa Schoenick, and Oyvind Tafjord. 2018.
\newblock Think you have solved question answering? try arc, the {AI2} reasoning challenge.
\newblock \emph{CoRR}, abs/1803.05457.

\bibitem[{Cobbe et~al.(2021)Cobbe, Kosaraju, Bavarian, Chen, Jun, Kaiser, Plappert, Tworek, Hilton, Nakano, Hesse, and Schulman}]{cobbe2021gsm8k}
Karl Cobbe, Vineet Kosaraju, Mohammad Bavarian, Mark Chen, Heewoo Jun, Lukasz Kaiser, Matthias Plappert, Jerry Tworek, Jacob Hilton, Reiichiro Nakano, Christopher Hesse, and John Schulman. 2021.
\newblock Training verifiers to solve math word problems.
\newblock \emph{arXiv preprint arXiv:2110.14168}.

\bibitem[{Conneau et~al.(2018)Conneau, Rinott, Lample, Williams, Bowman, Schwenk, and Stoyanov}]{conneau-etal-2018-xnli}
Alexis Conneau, Ruty Rinott, Guillaume Lample, Adina Williams, Samuel Bowman, Holger Schwenk, and Veselin Stoyanov. 2018.
\newblock \href {https://doi.org/10.18653/v1/D18-1269} {{XNLI}: Evaluating cross-lingual sentence representations}.
\newblock In \emph{Proceedings of the 2018 Conference on Empirical Methods in Natural Language Processing}, pages 2475--2485, Brussels, Belgium. Association for Computational Linguistics.

\bibitem[{Dubey et~al.(2024)Dubey, Jauhri, Pandey, Kadian, Al-Dahle, Letman, Mathur, Schelten, Yang, Fan, Goyal, Hartshorn, Yang, Mitra, Sravankumar, Korenev, Hinsvark, Rao, Zhang, Rodriguez, Gregerson, Spataru, Roziere, Biron, Tang, Chern, Caucheteux, Nayak, Bi, Marra, McConnell, Keller, Touret, Wu, Wong, Ferrer, Nikolaidis, Allonsius, Song, Pintz, Livshits, Esiobu, Choudhary, Mahajan, Garcia-Olano, Perino, Hupkes, Lakomkin, AlBadawy, Lobanova, Dinan, Smith, Radenovic, Zhang, Synnaeve, Lee, Anderson, Nail, Mialon, Pang, Cucurell, Nguyen, Korevaar, Xu, Touvron, Zarov, Ibarra, Kloumann, Misra, Evtimov, Copet, Lee, Geffert, Vranes, Park, Mahadeokar, Shah, van~der Linde, Billock, Hong, Lee, Fu, Chi, Huang, Liu, Wang, Yu, Bitton, Spisak, Park, Rocca, Johnstun, Saxe, Jia, Alwala, Upasani, Plawiak, Li, Heafield, Stone, El-Arini, Iyer, Malik, Chiu, Bhalla, Rantala-Yeary, van~der Maaten, Chen, Tan, Jenkins, Martin, Madaan, Malo, Blecher, Landzaat, de~Oliveira, Muzzi, Pasupuleti, Singh, Paluri, Kardas, Oldham, Rita,
  Pavlova, Kambadur, Lewis, Si, Singh, Hassan, Goyal, Torabi, Bashlykov, Bogoychev, Chatterji, Duchenne, Çelebi, Alrassy, Zhang, Li, Vasic, Weng, Bhargava, Dubal, Krishnan, Koura, Xu, He, Dong, Srinivasan, Ganapathy, Calderer, Cabral, Stojnic, Raileanu, Girdhar, Patel, Sauvestre, Polidoro, Sumbaly, Taylor, Silva, Hou, Wang, Hosseini, Chennabasappa, Singh, Bell, Kim, Edunov, Nie, Narang, Raparthy, Shen, Wan, Bhosale, Zhang, Vandenhende, Batra, Whitman, Sootla, Collot, Gururangan, Borodinsky, Herman, Fowler, Sheasha, Georgiou, Scialom, Speckbacher, Mihaylov, Xiao, Karn, Goswami, Gupta, Ramanathan, Kerkez, Gonguet, Do, Vogeti, Petrovic, Chu, Xiong, Fu, Meers, Martinet, Wang, Tan, Xie, Jia, Wang, Goldschlag, Gaur, Babaei, Wen, Song, Zhang, Li, Mao, Coudert, Yan, Chen, Papakipos, Singh, Grattafiori, Jain, Kelsey, Shajnfeld, Gangidi, Victoria, Goldstand, Menon, Sharma, Boesenberg, Vaughan, Baevski, Feinstein, Kallet, Sangani, Yunus, Lupu, Alvarado, Caples, Gu, Ho, Poulton, Ryan, Ramchandani, Franco, Saraf,
  Chowdhury, Gabriel, Bharambe, Eisenman, Yazdan, James, Maurer, Leonhardi, Huang, Loyd, Paola, Paranjape, Liu, Wu, Ni, Hancock, Wasti, Spence, Stojkovic, Gamido, Montalvo, Parker, Burton, Mejia, Wang, Kim, Zhou, Hu, Chu, Cai, Tindal, Feichtenhofer, Civin, Beaty, Kreymer, Li, Wyatt, Adkins, Xu, Testuggine, David, Parikh, Liskovich, Foss, Wang, Le, Holland, Dowling, Jamil, Montgomery, Presani, Hahn, Wood, Brinkman, Arcaute, Dunbar, Smothers, Sun, Kreuk, Tian, Ozgenel, Caggioni, Guzmán, Kanayet, Seide, Florez, Schwarz, Badeer, Swee, Halpern, Thattai, Herman, Sizov, Guangyi, Zhang, Lakshminarayanan, Shojanazeri, Zou, Wang, Zha, Habeeb, Rudolph, Suk, Aspegren, Goldman, Damlaj, Molybog, Tufanov, Veliche, Gat, Weissman, Geboski, Kohli, Asher, Gaya, Marcus, Tang, Chan, Zhen, Reizenstein, Teboul, Zhong, Jin, Yang, Cummings, Carvill, Shepard, McPhie, Torres, Ginsburg, Wang, Wu, U, Saxena, Prasad, Khandelwal, Zand, Matosich, Veeraraghavan, Michelena, Li, Huang, Chawla, Lakhotia, Huang, Chen, Garg, A, Silva, Bell,
  Zhang, Guo, Yu, Moshkovich, Wehrstedt, Khabsa, Avalani, Bhatt, Tsimpoukelli, Mankus, Hasson, Lennie, Reso, Groshev, Naumov, Lathi, Keneally, Seltzer, Valko, Restrepo, Patel, Vyatskov, Samvelyan, Clark, Macey, Wang, Hermoso, Metanat, Rastegari, Bansal, Santhanam, Parks, White, Bawa, Singhal, Egebo, Usunier, Laptev, Dong, Zhang, Cheng, Chernoguz, Hart, Salpekar, Kalinli, Kent, Parekh, Saab, Balaji, Rittner, Bontrager, Roux, Dollar, Zvyagina, Ratanchandani, Yuvraj, Liang, Alao, Rodriguez, Ayub, Murthy, Nayani, Mitra, Li, Hogan, Battey, Wang, Maheswari, Howes, Rinott, Bondu, Datta, Chugh, Hunt, Dhillon, Sidorov, Pan, Verma, Yamamoto, Ramaswamy, Lindsay, Lindsay, Feng, Lin, Zha, Shankar, Zhang, Zhang, Wang, Agarwal, Sajuyigbe, Chintala, Max, Chen, Kehoe, Satterfield, Govindaprasad, Gupta, Cho, Virk, Subramanian, Choudhury, Goldman, Remez, Glaser, Best, Kohler, Robinson, Li, Zhang, Matthews, Chou, Shaked, Vontimitta, Ajayi, Montanez, Mohan, Kumar, Mangla, Albiero, Ionescu, Poenaru, Mihailescu, Ivanov, Li, Wang,
  Jiang, Bouaziz, Constable, Tang, Wang, Wu, Wang, Xia, Wu, Gao, Chen, Hu, Jia, Qi, Li, Zhang, Zhang, Adi, Nam, Yu, Wang, Hao, Qian, He, Rait, DeVito, Rosnbrick, Wen, Yang, and Zhao}]{dubey2024llama3herdmodels}
Abhimanyu Dubey, Abhinav Jauhri, Abhinav Pandey, Abhishek Kadian, Ahmad Al-Dahle, Aiesha Letman, Akhil Mathur, Alan Schelten, Amy Yang, Angela Fan, Anirudh Goyal, Anthony Hartshorn, Aobo Yang, Archi Mitra, Archie Sravankumar, Artem Korenev, Arthur Hinsvark, Arun Rao, Aston Zhang, Aurelien Rodriguez, Austen Gregerson, Ava Spataru, Baptiste Roziere, Bethany Biron, Binh Tang, Bobbie Chern, Charlotte Caucheteux, Chaya Nayak, Chloe Bi, Chris Marra, Chris McConnell, Christian Keller, Christophe Touret, Chunyang Wu, Corinne Wong, Cristian~Canton Ferrer, Cyrus Nikolaidis, Damien Allonsius, Daniel Song, Danielle Pintz, Danny Livshits, David Esiobu, Dhruv Choudhary, Dhruv Mahajan, Diego Garcia-Olano, Diego Perino, Dieuwke Hupkes, Egor Lakomkin, Ehab AlBadawy, Elina Lobanova, Emily Dinan, Eric~Michael Smith, Filip Radenovic, Frank Zhang, Gabriel Synnaeve, Gabrielle Lee, Georgia~Lewis Anderson, Graeme Nail, Gregoire Mialon, Guan Pang, Guillem Cucurell, Hailey Nguyen, Hannah Korevaar, Hu~Xu, Hugo Touvron, Iliyan Zarov,
  Imanol~Arrieta Ibarra, Isabel Kloumann, Ishan Misra, Ivan Evtimov, Jade Copet, Jaewon Lee, Jan Geffert, Jana Vranes, Jason Park, Jay Mahadeokar, Jeet Shah, Jelmer van~der Linde, Jennifer Billock, Jenny Hong, Jenya Lee, Jeremy Fu, Jianfeng Chi, Jianyu Huang, Jiawen Liu, Jie Wang, Jiecao Yu, Joanna Bitton, Joe Spisak, Jongsoo Park, Joseph Rocca, Joshua Johnstun, Joshua Saxe, Junteng Jia, Kalyan~Vasuden Alwala, Kartikeya Upasani, Kate Plawiak, Ke~Li, Kenneth Heafield, Kevin Stone, Khalid El-Arini, Krithika Iyer, Kshitiz Malik, Kuenley Chiu, Kunal Bhalla, Lauren Rantala-Yeary, Laurens van~der Maaten, Lawrence Chen, Liang Tan, Liz Jenkins, Louis Martin, Lovish Madaan, Lubo Malo, Lukas Blecher, Lukas Landzaat, Luke de~Oliveira, Madeline Muzzi, Mahesh Pasupuleti, Mannat Singh, Manohar Paluri, Marcin Kardas, Mathew Oldham, Mathieu Rita, Maya Pavlova, Melanie Kambadur, Mike Lewis, Min Si, Mitesh~Kumar Singh, Mona Hassan, Naman Goyal, Narjes Torabi, Nikolay Bashlykov, Nikolay Bogoychev, Niladri Chatterji, Olivier
  Duchenne, Onur Çelebi, Patrick Alrassy, Pengchuan Zhang, Pengwei Li, Petar Vasic, Peter Weng, Prajjwal Bhargava, Pratik Dubal, Praveen Krishnan, Punit~Singh Koura, Puxin Xu, Qing He, Qingxiao Dong, Ragavan Srinivasan, Raj Ganapathy, Ramon Calderer, Ricardo~Silveira Cabral, Robert Stojnic, Roberta Raileanu, Rohit Girdhar, Rohit Patel, Romain Sauvestre, Ronnie Polidoro, Roshan Sumbaly, Ross Taylor, Ruan Silva, Rui Hou, Rui Wang, Saghar Hosseini, Sahana Chennabasappa, Sanjay Singh, Sean Bell, Seohyun~Sonia Kim, Sergey Edunov, Shaoliang Nie, Sharan Narang, Sharath Raparthy, Sheng Shen, Shengye Wan, Shruti Bhosale, Shun Zhang, Simon Vandenhende, Soumya Batra, Spencer Whitman, Sten Sootla, Stephane Collot, Suchin Gururangan, Sydney Borodinsky, Tamar Herman, Tara Fowler, Tarek Sheasha, Thomas Georgiou, Thomas Scialom, Tobias Speckbacher, Todor Mihaylov, Tong Xiao, Ujjwal Karn, Vedanuj Goswami, Vibhor Gupta, Vignesh Ramanathan, Viktor Kerkez, Vincent Gonguet, Virginie Do, Vish Vogeti, Vladan Petrovic, Weiwei Chu,
  Wenhan Xiong, Wenyin Fu, Whitney Meers, Xavier Martinet, Xiaodong Wang, Xiaoqing~Ellen Tan, Xinfeng Xie, Xuchao Jia, Xuewei Wang, Yaelle Goldschlag, Yashesh Gaur, Yasmine Babaei, Yi~Wen, Yiwen Song, Yuchen Zhang, Yue Li, Yuning Mao, Zacharie~Delpierre Coudert, Zheng Yan, Zhengxing Chen, Zoe Papakipos, Aaditya Singh, Aaron Grattafiori, Abha Jain, Adam Kelsey, Adam Shajnfeld, Adithya Gangidi, Adolfo Victoria, Ahuva Goldstand, Ajay Menon, Ajay Sharma, Alex Boesenberg, Alex Vaughan, Alexei Baevski, Allie Feinstein, Amanda Kallet, Amit Sangani, Anam Yunus, Andrei Lupu, Andres Alvarado, Andrew Caples, Andrew Gu, Andrew Ho, Andrew Poulton, Andrew Ryan, Ankit Ramchandani, Annie Franco, Aparajita Saraf, Arkabandhu Chowdhury, Ashley Gabriel, Ashwin Bharambe, Assaf Eisenman, Azadeh Yazdan, Beau James, Ben Maurer, Benjamin Leonhardi, Bernie Huang, Beth Loyd, Beto~De Paola, Bhargavi Paranjape, Bing Liu, Bo~Wu, Boyu Ni, Braden Hancock, Bram Wasti, Brandon Spence, Brani Stojkovic, Brian Gamido, Britt Montalvo, Carl
  Parker, Carly Burton, Catalina Mejia, Changhan Wang, Changkyu Kim, Chao Zhou, Chester Hu, Ching-Hsiang Chu, Chris Cai, Chris Tindal, Christoph Feichtenhofer, Damon Civin, Dana Beaty, Daniel Kreymer, Daniel Li, Danny Wyatt, David Adkins, David Xu, Davide Testuggine, Delia David, Devi Parikh, Diana Liskovich, Didem Foss, Dingkang Wang, Duc Le, Dustin Holland, Edward Dowling, Eissa Jamil, Elaine Montgomery, Eleonora Presani, Emily Hahn, Emily Wood, Erik Brinkman, Esteban Arcaute, Evan Dunbar, Evan Smothers, Fei Sun, Felix Kreuk, Feng Tian, Firat Ozgenel, Francesco Caggioni, Francisco Guzmán, Frank Kanayet, Frank Seide, Gabriela~Medina Florez, Gabriella Schwarz, Gada Badeer, Georgia Swee, Gil Halpern, Govind Thattai, Grant Herman, Grigory Sizov, Guangyi, Zhang, Guna Lakshminarayanan, Hamid Shojanazeri, Han Zou, Hannah Wang, Hanwen Zha, Haroun Habeeb, Harrison Rudolph, Helen Suk, Henry Aspegren, Hunter Goldman, Ibrahim Damlaj, Igor Molybog, Igor Tufanov, Irina-Elena Veliche, Itai Gat, Jake Weissman, James
  Geboski, James Kohli, Japhet Asher, Jean-Baptiste Gaya, Jeff Marcus, Jeff Tang, Jennifer Chan, Jenny Zhen, Jeremy Reizenstein, Jeremy Teboul, Jessica Zhong, Jian Jin, Jingyi Yang, Joe Cummings, Jon Carvill, Jon Shepard, Jonathan McPhie, Jonathan Torres, Josh Ginsburg, Junjie Wang, Kai Wu, Kam~Hou U, Karan Saxena, Karthik Prasad, Kartikay Khandelwal, Katayoun Zand, Kathy Matosich, Kaushik Veeraraghavan, Kelly Michelena, Keqian Li, Kun Huang, Kunal Chawla, Kushal Lakhotia, Kyle Huang, Lailin Chen, Lakshya Garg, Lavender A, Leandro Silva, Lee Bell, Lei Zhang, Liangpeng Guo, Licheng Yu, Liron Moshkovich, Luca Wehrstedt, Madian Khabsa, Manav Avalani, Manish Bhatt, Maria Tsimpoukelli, Martynas Mankus, Matan Hasson, Matthew Lennie, Matthias Reso, Maxim Groshev, Maxim Naumov, Maya Lathi, Meghan Keneally, Michael~L. Seltzer, Michal Valko, Michelle Restrepo, Mihir Patel, Mik Vyatskov, Mikayel Samvelyan, Mike Clark, Mike Macey, Mike Wang, Miquel~Jubert Hermoso, Mo~Metanat, Mohammad Rastegari, Munish Bansal, Nandhini
  Santhanam, Natascha Parks, Natasha White, Navyata Bawa, Nayan Singhal, Nick Egebo, Nicolas Usunier, Nikolay~Pavlovich Laptev, Ning Dong, Ning Zhang, Norman Cheng, Oleg Chernoguz, Olivia Hart, Omkar Salpekar, Ozlem Kalinli, Parkin Kent, Parth Parekh, Paul Saab, Pavan Balaji, Pedro Rittner, Philip Bontrager, Pierre Roux, Piotr Dollar, Polina Zvyagina, Prashant Ratanchandani, Pritish Yuvraj, Qian Liang, Rachad Alao, Rachel Rodriguez, Rafi Ayub, Raghotham Murthy, Raghu Nayani, Rahul Mitra, Raymond Li, Rebekkah Hogan, Robin Battey, Rocky Wang, Rohan Maheswari, Russ Howes, Ruty Rinott, Sai~Jayesh Bondu, Samyak Datta, Sara Chugh, Sara Hunt, Sargun Dhillon, Sasha Sidorov, Satadru Pan, Saurabh Verma, Seiji Yamamoto, Sharadh Ramaswamy, Shaun Lindsay, Shaun Lindsay, Sheng Feng, Shenghao Lin, Shengxin~Cindy Zha, Shiva Shankar, Shuqiang Zhang, Shuqiang Zhang, Sinong Wang, Sneha Agarwal, Soji Sajuyigbe, Soumith Chintala, Stephanie Max, Stephen Chen, Steve Kehoe, Steve Satterfield, Sudarshan Govindaprasad, Sumit Gupta,
  Sungmin Cho, Sunny Virk, Suraj Subramanian, Sy~Choudhury, Sydney Goldman, Tal Remez, Tamar Glaser, Tamara Best, Thilo Kohler, Thomas Robinson, Tianhe Li, Tianjun Zhang, Tim Matthews, Timothy Chou, Tzook Shaked, Varun Vontimitta, Victoria Ajayi, Victoria Montanez, Vijai Mohan, Vinay~Satish Kumar, Vishal Mangla, Vítor Albiero, Vlad Ionescu, Vlad Poenaru, Vlad~Tiberiu Mihailescu, Vladimir Ivanov, Wei Li, Wenchen Wang, Wenwen Jiang, Wes Bouaziz, Will Constable, Xiaocheng Tang, Xiaofang Wang, Xiaojian Wu, Xiaolan Wang, Xide Xia, Xilun Wu, Xinbo Gao, Yanjun Chen, Ye~Hu, Ye~Jia, Ye~Qi, Yenda Li, Yilin Zhang, Ying Zhang, Yossi Adi, Youngjin Nam, Yu, Wang, Yuchen Hao, Yundi Qian, Yuzi He, Zach Rait, Zachary DeVito, Zef Rosnbrick, Zhaoduo Wen, Zhenyu Yang, and Zhiwei Zhao. 2024.
\newblock \href {https://arxiv.org/abs/2407.21783} {The llama 3 herd of models}.
\newblock \emph{Preprint}, arXiv:2407.21783.

\bibitem[{Goyal et~al.(2022)Goyal, Gao, Chaudhary, Chen, Wenzek, Ju, Krishnan, Ranzato, Guzm{\'a}n, and Fan}]{goyal-etal-2022-flores}
Naman Goyal, Cynthia Gao, Vishrav Chaudhary, Peng-Jen Chen, Guillaume Wenzek, Da~Ju, Sanjana Krishnan, Marc{'}Aurelio Ranzato, Francisco Guzm{\'a}n, and Angela Fan. 2022.
\newblock \href {https://doi.org/10.1162/tacl_a_00474} {The {F}lores-101 evaluation benchmark for low-resource and multilingual machine translation}.
\newblock \emph{Transactions of the Association for Computational Linguistics}, 10:522--538.

\bibitem[{Hendrycks et~al.(2020)Hendrycks, Burns, Basart, Zou, Mazeika, Song, and Steinhardt}]{hendrycks2020measuring}
Dan Hendrycks, Collin Burns, Steven Basart, Andy Zou, Mantas Mazeika, Dawn Song, and Jacob Steinhardt. 2020.
\newblock Measuring massive multitask language understanding.
\newblock \emph{arXiv preprint arXiv:2009.03300}.

\bibitem[{Jiang et~al.(2023)Jiang, Sablayrolles, Mensch, Bamford, Chaplot, de~las Casas, Bressand, Lengyel, Lample, Saulnier, Lavaud, Lachaux, Stock, Scao, Lavril, Wang, Lacroix, and Sayed}]{jiang2023mistral7b}
Albert~Q. Jiang, Alexandre Sablayrolles, Arthur Mensch, Chris Bamford, Devendra~Singh Chaplot, Diego de~las Casas, Florian Bressand, Gianna Lengyel, Guillaume Lample, Lucile Saulnier, Lélio~Renard Lavaud, Marie-Anne Lachaux, Pierre Stock, Teven~Le Scao, Thibaut Lavril, Thomas Wang, Timothée Lacroix, and William~El Sayed. 2023.
\newblock \href {https://arxiv.org/abs/2310.06825} {Mistral 7b}.
\newblock \emph{Preprint}, arXiv:2310.06825.

\bibitem[{Jiang et~al.(2024)Jiang, Sablayrolles, Roux, Mensch, Savary, Bamford, Chaplot, de~las Casas, Hanna, Bressand, Lengyel, Bour, Lample, Lavaud, Saulnier, Lachaux, Stock, Subramanian, Yang, Antoniak, Scao, Gervet, Lavril, Wang, Lacroix, and Sayed}]{jiang2024mixtralexperts}
Albert~Q. Jiang, Alexandre Sablayrolles, Antoine Roux, Arthur Mensch, Blanche Savary, Chris Bamford, Devendra~Singh Chaplot, Diego de~las Casas, Emma~Bou Hanna, Florian Bressand, Gianna Lengyel, Guillaume Bour, Guillaume Lample, Lélio~Renard Lavaud, Lucile Saulnier, Marie-Anne Lachaux, Pierre Stock, Sandeep Subramanian, Sophia Yang, Szymon Antoniak, Teven~Le Scao, Théophile Gervet, Thibaut Lavril, Thomas Wang, Timothée Lacroix, and William~El Sayed. 2024.
\newblock \href {https://arxiv.org/abs/2401.04088} {Mixtral of experts}.
\newblock \emph{Preprint}, arXiv:2401.04088.

\bibitem[{Kocmi et~al.(2023)Kocmi, Avramidis, Bawden, Bojar, Dvorkovich, Federmann, Fishel, Freitag, Gowda, Grundkiewicz, Haddow, Koehn, Marie, Monz, Morishita, Murray, Nagata, Nakazawa, Popel, Popovi{\'c}, and Shmatova}]{kocmi-etal-2023-findings}
Tom Kocmi, Eleftherios Avramidis, Rachel Bawden, Ond{\v{r}}ej Bojar, Anton Dvorkovich, Christian Federmann, Mark Fishel, Markus Freitag, Thamme Gowda, Roman Grundkiewicz, Barry Haddow, Philipp Koehn, Benjamin Marie, Christof Monz, Makoto Morishita, Kenton Murray, Makoto Nagata, Toshiaki Nakazawa, Martin Popel, Maja Popovi{\'c}, and Mariya Shmatova. 2023.
\newblock \href {https://doi.org/10.18653/v1/2023.wmt-1.1} {Findings of the 2023 conference on machine translation ({WMT}23): {LLM}s are here but not quite there yet}.
\newblock In \emph{Proceedings of the Eighth Conference on Machine Translation}, pages 1--42, Singapore. Association for Computational Linguistics.

\bibitem[{Lai et~al.(2023)Lai, Nguyen, Ngo, Nguyen, Dernoncourt, Rossi, and Nguyen}]{lai-etal-2023-okapi}
Viet Lai, Chien Nguyen, Nghia Ngo, Thuat Nguyen, Franck Dernoncourt, Ryan Rossi, and Thien Nguyen. 2023.
\newblock \href {https://doi.org/10.18653/v1/2023.emnlp-demo.28} {Okapi: Instruction-tuned large language models in multiple languages with reinforcement learning from human feedback}.
\newblock In \emph{Proceedings of the 2023 Conference on Empirical Methods in Natural Language Processing: System Demonstrations}, pages 318--327, Singapore. Association for Computational Linguistics.

\bibitem[{Lewis et~al.(2020)Lewis, Oguz, Rinott, Riedel, and Schwenk}]{lewis-etal-2020-mlqa}
Patrick Lewis, Barlas Oguz, Ruty Rinott, Sebastian Riedel, and Holger Schwenk. 2020.
\newblock \href {https://doi.org/10.18653/v1/2020.acl-main.653} {{MLQA}: Evaluating cross-lingual extractive question answering}.
\newblock In \emph{Proceedings of the 58th Annual Meeting of the Association for Computational Linguistics}, pages 7315--7330, Online. Association for Computational Linguistics.

\bibitem[{Lin et~al.(2022)Lin, Hilton, and Evans}]{lin-etal-2022-truthfulqa}
Stephanie Lin, Jacob Hilton, and Owain Evans. 2022.
\newblock \href {https://doi.org/10.18653/v1/2022.acl-long.229} {{T}ruthful{QA}: Measuring how models mimic human falsehoods}.
\newblock In \emph{Proceedings of the 60th Annual Meeting of the Association for Computational Linguistics (Volume 1: Long Papers)}, pages 3214--3252, Dublin, Ireland. Association for Computational Linguistics.

\bibitem[{Martins et~al.(2024)Martins, Fernandes, Alves, Guerreiro, Rei, Alves, Pombal, Farajian, Faysse, Klimaszewski, Colombo, Haddow, de~Souza, Birch, and Martins}]{martins2024eurollmmultilinguallanguagemodels}
Pedro~Henrique Martins, Patrick Fernandes, João Alves, Nuno~M. Guerreiro, Ricardo Rei, Duarte~M. Alves, José Pombal, Amin Farajian, Manuel Faysse, Mateusz Klimaszewski, Pierre Colombo, Barry Haddow, José G.~C. de~Souza, Alexandra Birch, and André F.~T. Martins. 2024.
\newblock \href {https://arxiv.org/abs/2409.16235} {Eurollm: Multilingual language models for europe}.
\newblock \emph{Preprint}, arXiv:2409.16235.

\bibitem[{Meng et~al.(2022)Meng, Huang, Zhang, and Han}]{meng2022generating}
Yu~Meng, Jiaxin Huang, Yu~Zhang, and Jiawei Han. 2022.
\newblock Generating training data with language models: Towards zero-shot language understanding.
\newblock \emph{Advances in Neural Information Processing Systems}, 35:462--477.

\bibitem[{Muennighoff et~al.(2023)Muennighoff, Wang, Sutawika, Roberts, Biderman, Scao, Bari, Shen, Yong, Schoelkopf, Tang, Radev, Aji, Almubarak, Albanie, Alyafeai, Webson, Raff, and Raffel}]{muennighoff2023crosslingualgeneralizationmultitaskfinetuning}
Niklas Muennighoff, Thomas Wang, Lintang Sutawika, Adam Roberts, Stella Biderman, Teven~Le Scao, M~Saiful Bari, Sheng Shen, Zheng-Xin Yong, Hailey Schoelkopf, Xiangru Tang, Dragomir Radev, Alham~Fikri Aji, Khalid Almubarak, Samuel Albanie, Zaid Alyafeai, Albert Webson, Edward Raff, and Colin Raffel. 2023.
\newblock \href {https://arxiv.org/abs/2211.01786} {Crosslingual generalization through multitask finetuning}.
\newblock \emph{Preprint}, arXiv:2211.01786.

\bibitem[{{NLLB Team} et~al.(2022){NLLB Team}, Costa-jussà, Cross, Çelebi, Elbayad, Heafield, Heffernan, Kalbassi, Lam, Licht, Maillard, Sun, Wang, Wenzek, Youngblood, Akula, Barrault, Gonzalez, Hansanti, Hoffman, Jarrett, Sadagopan, Rowe, Spruit, Tran, Andrews, Ayan, Bhosale, Edunov, Fan, Gao, Goswami, Guzmán, Koehn, Mourachko, Ropers, Saleem, Schwenk, and Wang}]{nllbteam2022language}
{NLLB Team}, Marta~R. Costa-jussà, James Cross, Onur Çelebi, Maha Elbayad, Kenneth Heafield, Kevin Heffernan, Elahe Kalbassi, Janice Lam, Daniel Licht, Jean Maillard, Anna Sun, Skyler Wang, Guillaume Wenzek, Al~Youngblood, Bapi Akula, Loic Barrault, Gabriel~Mejia Gonzalez, Prangthip Hansanti, John Hoffman, Semarley Jarrett, Kaushik~Ram Sadagopan, Dirk Rowe, Shannon Spruit, Chau Tran, Pierre Andrews, Necip~Fazil Ayan, Shruti Bhosale, Sergey Edunov, Angela Fan, Cynthia Gao, Vedanuj Goswami, Francisco Guzmán, Philipp Koehn, Alexandre Mourachko, Christophe Ropers, Safiyyah Saleem, Holger Schwenk, and Jeff Wang. 2022.
\newblock \href {https://arxiv.org/abs/2207.04672} {No language left behind: Scaling human-centered machine translation}.
\newblock \emph{Preprint}, arXiv:2207.04672.

\bibitem[{Ott et~al.(2022)Ott, Barbosa-Silva, Blagec, Brauner, and Samwald}]{ott2022mapping}
Simon Ott, Adriano Barbosa-Silva, Kathrin Blagec, Jan Brauner, and Matthias Samwald. 2022.
\newblock Mapping global dynamics of benchmark creation and saturation in artificial intelligence.
\newblock \emph{Nature Communications}, 13(1):6793.

\bibitem[{Peng et~al.(2023)Peng, Li, He, Galley, and Gao}]{peng2023instructiontuninggpt4}
Baolin Peng, Chunyuan Li, Pengcheng He, Michel Galley, and Jianfeng Gao. 2023.
\newblock \href {https://arxiv.org/abs/2304.03277} {Instruction tuning with gpt-4}.
\newblock \emph{Preprint}, arXiv:2304.03277.

\bibitem[{Rei et~al.(2023)Rei, Guerreiro, Pombal, van Stigt, Treviso, Coheur, C.~de Souza, and Martins}]{rei-etal-2023-scaling}
Ricardo Rei, Nuno~M. Guerreiro, Jos{\~A}{\copyright} Pombal, Daan van Stigt, Marcos Treviso, Luisa Coheur, Jos{\'e}~G. C.~de Souza, and Andr{\'e} Martins. 2023.
\newblock \href {https://doi.org/10.18653/v1/2023.wmt-1.73} {Scaling up {C}omet{K}iwi: Unbabel-{IST} 2023 submission for the quality estimation shared task}.
\newblock In \emph{Proceedings of the Eighth Conference on Machine Translation}, pages 841--848, Singapore. Association for Computational Linguistics.

\bibitem[{Sakaguchi et~al.(2019)Sakaguchi, Bras, Bhagavatula, and Choi}]{winogrande}
Keisuke Sakaguchi, Ronan~Le Bras, Chandra Bhagavatula, and Yejin Choi. 2019.
\newblock \href {https://arxiv.org/abs/1907.10641} {Winogrande: An adversarial winograd schema challenge at scale}.
\newblock \emph{Preprint}, arXiv:1907.10641.

\bibitem[{Sreenivas et~al.(2024)Sreenivas, Muralidharan, Joshi, Chochowski, Patwary, Shoeybi, Catanzaro, Kautz, and Molchanov}]{sreenivas2024llmpruningdistillationpractice}
Sharath~Turuvekere Sreenivas, Saurav Muralidharan, Raviraj Joshi, Marcin Chochowski, Mostofa Patwary, Mohammad Shoeybi, Bryan Catanzaro, Jan Kautz, and Pavlo Molchanov. 2024.
\newblock \href {https://arxiv.org/abs/2408.11796} {Llm pruning and distillation in practice: The minitron approach}.
\newblock \emph{Preprint}, arXiv:2408.11796.

\bibitem[{Srivastava et~al.(2023)Srivastava, Rastogi, Rao, Shoeb, Abid, Fisch, Brown, Santoro, Gupta, Garriga-Alonso, Kluska, Lewkowycz, Agarwal, Power, Ray, Warstadt, Kocurek, Safaya, Tazarv, Xiang, Parrish, Nie, Hussain, Askell, Dsouza, Slone, Rahane, Iyer, Andreassen, Madotto, Santilli, Stuhlm{\"u}ller, Dai, La, Lampinen, Zou, Jiang, Chen, Vuong, Gupta, Gottardi, Norelli, Venkatesh, Gholamidavoodi, Tabassum, Menezes, Kirubarajan, Mullokandov, Sabharwal, Herrick, Efrat, Erdem, Karaka{\c{s}}, Roberts, Loe, Zoph, Bojanowski, {\"O}zyurt, Hedayatnia, Neyshabur, Inden, Stein, Ekmekci, Lin, Howald, Orinion, Diao, Dour, Stinson, Argueta, Ferri, Singh, Rathkopf, Meng, Baral, Wu, Callison-Burch, Waites, Voigt, Manning, Potts, Ramirez, Rivera, Siro, Raffel, Ashcraft, Garbacea, Sileo, Garrette, Hendrycks, Kilman, Roth, Freeman, Khashabi, Levy, Gonz{\'a}lez, Perszyk, Hernandez, Chen, Ippolito, Gilboa, Dohan, Drakard, Jurgens, Datta, Ganguli, Emelin, Kleyko, Yuret, Chen, Tam, Hupkes, Misra, Buzan, Mollo, Yang, Lee,
  Schrader, Shutova, Cubuk, Segal, Hagerman, Barnes, Donoway, Pavlick, Rodol{\`a}, Lam, Chu, Tang, Erdem, Chang, Chi, Dyer, Jerzak, Kim, Manyasi, Zheltonozhskii, Xia, Siar, Mart{\'\i}nez-Plumed, Happ{\'e}, Chollet, Rong, Mishra, Winata, de~Melo, Kruszewski, Parascandolo, Mariani, Wang, Jaimovitch-Lopez, Betz, Gur-Ari, Galijasevic, Kim, Rashkin, Hajishirzi, Mehta, Bogar, Shevlin, Schuetze, Yakura, Zhang, Wong, Ng, Noble, Jumelet, Geissinger, Kernion, Hilton, Lee, Fisac, Simon, Koppel, Zheng, Zou, Kocon, Thompson, Wingfield, Kaplan, Radom, Sohl-Dickstein, Phang, Wei, Yosinski, Novikova, Bosscher, Marsh, Kim, Taal, Engel, Alabi, Xu, Song, Tang, Waweru, Burden, Miller, Balis, Batchelder, Berant, Frohberg, Rozen, Hernandez-Orallo, Boudeman, Guerr, Jones, Tenenbaum, Rule, Chua, Kanclerz, Livescu, Krauth, Gopalakrishnan, Ignatyeva, Markert, Dhole, Gimpel, Omondi, Mathewson, Chiafullo, Shkaruta, Shridhar, McDonell, Richardson, Reynolds, Gao, Zhang, Dugan, Qin, Contreras-Ochando, Morency, Moschella, Lam, Noble,
  Schmidt, He, Oliveros-Col{\'o}n, Metz, Senel, Bosma, Sap, Hoeve, Farooqi, Faruqui, Mazeika, Baturan, Marelli, Maru, Ramirez-Quintana, Tolkiehn, Giulianelli, Lewis, Potthast, Leavitt, Hagen, Schubert, Baitemirova, Arnaud, McElrath, Yee, Cohen, Gu, Ivanitskiy, Starritt, Strube, Sw{\k{e}}drowski, Bevilacqua, Yasunaga, Kale, Cain, Xu, Suzgun, Walker, Tiwari, Bansal, Aminnaseri, Geva, Gheini, T, Peng, Chi, Lee, Krakover, Cameron, Roberts, Doiron, Martinez, Nangia, Deckers, Muennighoff, Keskar, Iyer, Constant, Fiedel, Wen, Zhang, Agha, Elbaghdadi, Levy, Evans, Casares, Doshi, Fung, Liang, Vicol, Alipoormolabashi, Liao, Liang, Chang, Eckersley, Htut, Hwang, Mi{\l}kowski, Patil, Pezeshkpour, Oli, Mei, Lyu, Chen, Banjade, Rudolph, Gabriel, Habacker, Risco, Milli{\`e}re, Garg, Barnes, Saurous, Arakawa, Raymaekers, Frank, Sikand, Novak, Sitelew, Bras, Liu, Jacobs, Zhang, Salakhutdinov, Chi, Lee, Stovall, Teehan, Yang, Singh, Mohammad, Anand, Dillavou, Shleifer, Wiseman, Gruetter, Bowman, Schoenholz, Han, Kwatra, Rous,
  Ghazarian, Ghosh, Casey, Bischoff, Gehrmann, Schuster, Sadeghi, Hamdan, Zhou, Srivastava, Shi, Singh, Asaadi, Gu, Pachchigar, Toshniwal, Upadhyay, Debnath, Shakeri, Thormeyer, Melzi, Reddy, Makini, Lee, Torene, Hatwar, Dehaene, Divic, Ermon, Biderman, Lin, Prasad, Piantadosi, Shieber, Misherghi, Kiritchenko, Mishra, Linzen, Schuster, Li, Yu, Ali, Hashimoto, Wu, Desbordes, Rothschild, Phan, Wang, Nkinyili, Schick, Kornev, Tunduny, Gerstenberg, Chang, Neeraj, Khot, Shultz, Shaham, Misra, Demberg, Nyamai, Raunak, Ramasesh, vinay~uday prabhu, Padmakumar, Srikumar, Fedus, Saunders, Zhang, Vossen, Ren, Tong, Zhao, Wu, Shen, Yaghoobzadeh, Lakretz, Song, Bahri, Choi, Yang, Hao, Chen, Belinkov, Hou, Hou, Bai, Seid, Zhao, Wang, Wang, Wang, and Wu}]{51569}
Aarohi Srivastava, Abhinav Rastogi, Abhishek Rao, Abu Awal~Md Shoeb, Abubakar Abid, Adam Fisch, Adam~R. Brown, Adam Santoro, Aditya Gupta, Adri{\`a} Garriga-Alonso, Agnieszka Kluska, Aitor Lewkowycz, Akshat Agarwal, Alethea Power, Alex Ray, Alex Warstadt, Alexander~W. Kocurek, Ali Safaya, Ali Tazarv, Alice Xiang, Alicia Parrish, Allen Nie, Aman Hussain, Amanda Askell, Amanda Dsouza, Ambrose Slone, Ameet Rahane, Anantharaman~S. Iyer, Anders~Johan Andreassen, Andrea Madotto, Andrea Santilli, Andreas Stuhlm{\"u}ller, Andrew~M. Dai, Andrew La, Andrew Lampinen, Andy Zou, Angela Jiang, Angelica Chen, Anh Vuong, Animesh Gupta, Anna Gottardi, Antonio Norelli, Anu Venkatesh, Arash Gholamidavoodi, Arfa Tabassum, Arul Menezes, Arun Kirubarajan, Asher Mullokandov, Ashish Sabharwal, Austin Herrick, Avia Efrat, Aykut Erdem, Ayla Karaka{\c{s}}, B.~Ryan Roberts, Bao~Sheng Loe, Barret Zoph, Bart{\l}omiej Bojanowski, Batuhan {\"O}zyurt, Behnam Hedayatnia, Behnam Neyshabur, Benjamin Inden, Benno Stein, Berk Ekmekci, Bill~Yuchen
  Lin, Blake Howald, Bryan Orinion, Cameron Diao, Cameron Dour, Catherine Stinson, Cedrick Argueta, Cesar Ferri, Chandan Singh, Charles Rathkopf, Chenlin Meng, Chitta Baral, Chiyu Wu, Chris Callison-Burch, Christopher Waites, Christian Voigt, Christopher~D Manning, Christopher Potts, Cindy Ramirez, Clara~E. Rivera, Clemencia Siro, Colin Raffel, Courtney Ashcraft, Cristina Garbacea, Damien Sileo, Dan Garrette, Dan Hendrycks, Dan Kilman, Dan Roth, C.~Daniel Freeman, Daniel Khashabi, Daniel Levy, Daniel~Mosegu{\'\i} Gonz{\'a}lez, Danielle Perszyk, Danny Hernandez, Danqi Chen, Daphne Ippolito, Dar Gilboa, David Dohan, David Drakard, David Jurgens, Debajyoti Datta, Deep Ganguli, Denis Emelin, Denis Kleyko, Deniz Yuret, Derek Chen, Derek Tam, Dieuwke Hupkes, Diganta Misra, Dilyar Buzan, Dimitri~Coelho Mollo, Diyi Yang, Dong-Ho Lee, Dylan Schrader, Ekaterina Shutova, Ekin~Dogus Cubuk, Elad Segal, Eleanor Hagerman, Elizabeth Barnes, Elizabeth Donoway, Ellie Pavlick, Emanuele Rodol{\`a}, Emma Lam, Eric Chu, Eric Tang,
  Erkut Erdem, Ernie Chang, Ethan~A Chi, Ethan Dyer, Ethan Jerzak, Ethan Kim, Eunice~Engefu Manyasi, Evgenii Zheltonozhskii, Fanyue Xia, Fatemeh Siar, Fernando Mart{\'\i}nez-Plumed, Francesca Happ{\'e}, Francois Chollet, Frieda Rong, Gaurav Mishra, Genta~Indra Winata, Gerard de~Melo, Germ{\'a}n Kruszewski, Giambattista Parascandolo, Giorgio Mariani, Gloria~Xinyue Wang, Gonzalo Jaimovitch-Lopez, Gregor Betz, Guy Gur-Ari, Hana Galijasevic, Hannah Kim, Hannah Rashkin, Hannaneh Hajishirzi, Harsh Mehta, Hayden Bogar, Henry Francis~Anthony Shevlin, Hinrich Schuetze, Hiromu Yakura, Hongming Zhang, Hugh~Mee Wong, Ian Ng, Isaac Noble, Jaap Jumelet, Jack Geissinger, Jackson Kernion, Jacob Hilton, Jaehoon Lee, Jaime~Fern{\'a}ndez Fisac, James~B Simon, James Koppel, James Zheng, James Zou, Jan Kocon, Jana Thompson, Janelle Wingfield, Jared Kaplan, Jarema Radom, Jascha Sohl-Dickstein, Jason Phang, Jason Wei, Jason Yosinski, Jekaterina Novikova, Jelle Bosscher, Jennifer Marsh, Jeremy Kim, Jeroen Taal, Jesse Engel, Jesujoba
  Alabi, Jiacheng Xu, Jiaming Song, Jillian Tang, Joan Waweru, John Burden, John Miller, John~U. Balis, Jonathan Batchelder, Jonathan Berant, J{\"o}rg Frohberg, Jos Rozen, Jose Hernandez-Orallo, Joseph Boudeman, Joseph Guerr, Joseph Jones, Joshua~B. Tenenbaum, Joshua~S. Rule, Joyce Chua, Kamil Kanclerz, Karen Livescu, Karl Krauth, Karthik Gopalakrishnan, Katerina Ignatyeva, Katja Markert, Kaustubh Dhole, Kevin Gimpel, Kevin Omondi, Kory~Wallace Mathewson, Kristen Chiafullo, Ksenia Shkaruta, Kumar Shridhar, Kyle McDonell, Kyle Richardson, Laria Reynolds, Leo Gao, Li~Zhang, Liam Dugan, Lianhui Qin, Lidia Contreras-Ochando, Louis-Philippe Morency, Luca Moschella, Lucas Lam, Lucy Noble, Ludwig Schmidt, Luheng He, Luis Oliveros-Col{\'o}n, Luke Metz, L{\"u}tfi~Kerem Senel, Maarten Bosma, Maarten Sap, Maartje~Ter Hoeve, Maheen Farooqi, Manaal Faruqui, Mantas Mazeika, Marco Baturan, Marco Marelli, Marco Maru, Maria~Jose Ramirez-Quintana, Marie Tolkiehn, Mario Giulianelli, Martha Lewis, Martin Potthast, Matthew~L
  Leavitt, Matthias Hagen, M{\'a}ty{\'a}s Schubert, Medina~Orduna Baitemirova, Melody Arnaud, Melvin McElrath, Michael~Andrew Yee, Michael Cohen, Michael Gu, Michael Ivanitskiy, Michael Starritt, Michael Strube, Micha{\l} Sw{\k{e}}drowski, Michele Bevilacqua, Michihiro Yasunaga, Mihir Kale, Mike Cain, Mimee Xu, Mirac Suzgun, Mitch Walker, Mo~Tiwari, Mohit Bansal, Moin Aminnaseri, Mor Geva, Mozhdeh Gheini, Mukund~Varma T, Nanyun Peng, Nathan~Andrew Chi, Nayeon Lee, Neta Gur-Ari Krakover, Nicholas Cameron, Nicholas Roberts, Nick Doiron, Nicole Martinez, Nikita Nangia, Niklas Deckers, Niklas Muennighoff, Nitish~Shirish Keskar, Niveditha~S. Iyer, Noah Constant, Noah Fiedel, Nuan Wen, Oliver Zhang, Omar Agha, Omar Elbaghdadi, Omer Levy, Owain Evans, Pablo Antonio~Moreno Casares, Parth Doshi, Pascale Fung, Paul~Pu Liang, Paul Vicol, Pegah Alipoormolabashi, Peiyuan Liao, Percy Liang, Peter~W Chang, Peter Eckersley, Phu~Mon Htut, Pinyu Hwang, Piotr Mi{\l}kowski, Piyush Patil, Pouya Pezeshkpour, Priti Oli, Qiaozhu
  Mei, Qing Lyu, Qinlang Chen, Rabin Banjade, Rachel~Etta Rudolph, Raefer Gabriel, Rahel Habacker, Ramon Risco, Rapha{\"e}l Milli{\`e}re, Rhythm Garg, Richard Barnes, Rif~A. Saurous, Riku Arakawa, Robbe Raymaekers, Robert Frank, Rohan Sikand, Roman Novak, Roman Sitelew, Ronan~Le Bras, Rosanne Liu, Rowan Jacobs, Rui Zhang, Russ Salakhutdinov, Ryan~Andrew Chi, Seungjae~Ryan Lee, Ryan Stovall, Ryan Teehan, Rylan Yang, Sahib Singh, Saif~M. Mohammad, Sajant Anand, Sam Dillavou, Sam Shleifer, Sam Wiseman, Samuel Gruetter, Samuel~R. Bowman, Samuel~Stern Schoenholz, Sanghyun Han, Sanjeev Kwatra, Sarah~A. Rous, Sarik Ghazarian, Sayan Ghosh, Sean Casey, Sebastian Bischoff, Sebastian Gehrmann, Sebastian Schuster, Sepideh Sadeghi, Shadi Hamdan, Sharon Zhou, Shashank Srivastava, Sherry Shi, Shikhar Singh, Shima Asaadi, Shixiang~Shane Gu, Shubh Pachchigar, Shubham Toshniwal, Shyam Upadhyay, Shyamolima~Shammie Debnath, Siamak Shakeri, Simon Thormeyer, Simone Melzi, Siva Reddy, Sneha~Priscilla Makini, Soo-Hwan Lee, Spencer
  Torene, Sriharsha Hatwar, Stanislas Dehaene, Stefan Divic, Stefano Ermon, Stella Biderman, Stephanie Lin, Stephen Prasad, Steven Piantadosi, Stuart Shieber, Summer Misherghi, Svetlana Kiritchenko, Swaroop Mishra, Tal Linzen, Tal Schuster, Tao Li, Tao Yu, Tariq Ali, Tatsunori Hashimoto, Te-Lin Wu, Th{\'e}o Desbordes, Theodore Rothschild, Thomas Phan, Tianle Wang, Tiberius Nkinyili, Timo Schick, Timofei Kornev, Titus Tunduny, Tobias Gerstenberg, Trenton Chang, Trishala Neeraj, Tushar Khot, Tyler Shultz, Uri Shaham, Vedant Misra, Vera Demberg, Victoria Nyamai, Vikas Raunak, Vinay~Venkatesh Ramasesh, vinay~uday prabhu, Vishakh Padmakumar, Vivek Srikumar, William Fedus, William Saunders, William Zhang, Wout Vossen, Xiang Ren, Xiaoyu Tong, Xinran Zhao, Xinyi Wu, Xudong Shen, Yadollah Yaghoobzadeh, Yair Lakretz, Yangqiu Song, Yasaman Bahri, Yejin Choi, Yichi Yang, Yiding Hao, Yifu Chen, Yonatan Belinkov, Yu~Hou, Yufang Hou, Yuntao Bai, Zachary Seid, Zhuoye Zhao, Zijian Wang, Zijie~J. Wang, Zirui Wang, and Ziyi Wu.
  2023.
\newblock \href {https://openreview.net/forum?id=uyTL5Bvosj} {Beyond the imitation game: Quantifying and extrapolating the capabilities of language models}.
\newblock \emph{Transactions on Machine Learning Research}.

\bibitem[{Team et~al.(2024{\natexlab{a}})Team, Mesnard, Hardin, Dadashi, Bhupatiraju, Pathak, Sifre, Rivière, Kale, Love, Tafti, Hussenot, Sessa, Chowdhery, Roberts, Barua, Botev, Castro-Ros, Slone, Héliou, Tacchetti, Bulanova, Paterson, Tsai, Shahriari, Lan, Choquette-Choo, Crepy, Cer, Ippolito, Reid, Buchatskaya, Ni, Noland, Yan, Tucker, Muraru, Rozhdestvenskiy, Michalewski, Tenney, Grishchenko, Austin, Keeling, Labanowski, Lespiau, Stanway, Brennan, Chen, Ferret, Chiu, Mao-Jones, Lee, Yu, Millican, Sjoesund, Lee, Dixon, Reid, Mikuła, Wirth, Sharman, Chinaev, Thain, Bachem, Chang, Wahltinez, Bailey, Michel, Yotov, Chaabouni, Comanescu, Jana, Anil, McIlroy, Liu, Mullins, Smith, Borgeaud, Girgin, Douglas, Pandya, Shakeri, De, Klimenko, Hennigan, Feinberg, Stokowiec, hui Chen, Ahmed, Gong, Warkentin, Peran, Giang, Farabet, Vinyals, Dean, Kavukcuoglu, Hassabis, Ghahramani, Eck, Barral, Pereira, Collins, Joulin, Fiedel, Senter, Andreev, and Kenealy}]{gemmateam2024gemmaopenmodelsbased}
Gemma Team, Thomas Mesnard, Cassidy Hardin, Robert Dadashi, Surya Bhupatiraju, Shreya Pathak, Laurent Sifre, Morgane Rivière, Mihir~Sanjay Kale, Juliette Love, Pouya Tafti, Léonard Hussenot, Pier~Giuseppe Sessa, Aakanksha Chowdhery, Adam Roberts, Aditya Barua, Alex Botev, Alex Castro-Ros, Ambrose Slone, Amélie Héliou, Andrea Tacchetti, Anna Bulanova, Antonia Paterson, Beth Tsai, Bobak Shahriari, Charline~Le Lan, Christopher~A. Choquette-Choo, Clément Crepy, Daniel Cer, Daphne Ippolito, David Reid, Elena Buchatskaya, Eric Ni, Eric Noland, Geng Yan, George Tucker, George-Christian Muraru, Grigory Rozhdestvenskiy, Henryk Michalewski, Ian Tenney, Ivan Grishchenko, Jacob Austin, James Keeling, Jane Labanowski, Jean-Baptiste Lespiau, Jeff Stanway, Jenny Brennan, Jeremy Chen, Johan Ferret, Justin Chiu, Justin Mao-Jones, Katherine Lee, Kathy Yu, Katie Millican, Lars~Lowe Sjoesund, Lisa Lee, Lucas Dixon, Machel Reid, Maciej Mikuła, Mateo Wirth, Michael Sharman, Nikolai Chinaev, Nithum Thain, Olivier Bachem,
  Oscar Chang, Oscar Wahltinez, Paige Bailey, Paul Michel, Petko Yotov, Rahma Chaabouni, Ramona Comanescu, Reena Jana, Rohan Anil, Ross McIlroy, Ruibo Liu, Ryan Mullins, Samuel~L Smith, Sebastian Borgeaud, Sertan Girgin, Sholto Douglas, Shree Pandya, Siamak Shakeri, Soham De, Ted Klimenko, Tom Hennigan, Vlad Feinberg, Wojciech Stokowiec, Yu~hui Chen, Zafarali Ahmed, Zhitao Gong, Tris Warkentin, Ludovic Peran, Minh Giang, Clément Farabet, Oriol Vinyals, Jeff Dean, Koray Kavukcuoglu, Demis Hassabis, Zoubin Ghahramani, Douglas Eck, Joelle Barral, Fernando Pereira, Eli Collins, Armand Joulin, Noah Fiedel, Evan Senter, Alek Andreev, and Kathleen Kenealy. 2024{\natexlab{a}}.
\newblock \href {https://arxiv.org/abs/2403.08295} {Gemma: Open models based on gemini research and technology}.
\newblock \emph{Preprint}, arXiv:2403.08295.

\bibitem[{Team et~al.(2024{\natexlab{b}})Team, Riviere, Pathak, Sessa, Hardin, Bhupatiraju, Hussenot, Mesnard, Shahriari, Ramé, Ferret, Liu, Tafti, Friesen, Casbon, Ramos, Kumar, Lan, Jerome, Tsitsulin, Vieillard, Stanczyk, Girgin, Momchev, Hoffman, Thakoor, Grill, Neyshabur, Bachem, Walton, Severyn, Parrish, Ahmad, Hutchison, Abdagic, Carl, Shen, Brock, Coenen, Laforge, Paterson, Bastian, Piot, Wu, Royal, Chen, Kumar, Perry, Welty, Choquette-Choo, Sinopalnikov, Weinberger, Vijaykumar, Rogozińska, Herbison, Bandy, Wang, Noland, Moreira, Senter, Eltyshev, Visin, Rasskin, Wei, Cameron, Martins, Hashemi, Klimczak-Plucińska, Batra, Dhand, Nardini, Mein, Zhou, Svensson, Stanway, Chan, Zhou, Carrasqueira, Iljazi, Becker, Fernandez, van Amersfoort, Gordon, Lipschultz, Newlan, yeong Ji, Mohamed, Badola, Black, Millican, McDonell, Nguyen, Sodhia, Greene, Sjoesund, Usui, Sifre, Heuermann, Lago, McNealus, Soares, Kilpatrick, Dixon, Martins, Reid, Singh, Iverson, Görner, Velloso, Wirth, Davidow, Miller, Rahtz,
  Watson, Risdal, Kazemi, Moynihan, Zhang, Kahng, Park, Rahman, Khatwani, Dao, Bardoliwalla, Devanathan, Dumai, Chauhan, Wahltinez, Botarda, Barnes, Barham, Michel, Jin, Georgiev, Culliton, Kuppala, Comanescu, Merhej, Jana, Rokni, Agarwal, Mullins, Saadat, Carthy, Cogan, Perrin, Arnold, Krause, Dai, Garg, Sheth, Ronstrom, Chan, Jordan, Yu, Eccles, Hennigan, Kocisky, Doshi, Jain, Yadav, Meshram, Dharmadhikari, Barkley, Wei, Ye, Han, Kwon, Xu, Shen, Gong, Wei, Cotruta, Kirk, Rao, Giang, Peran, Warkentin, Collins, Barral, Ghahramani, Hadsell, Sculley, Banks, Dragan, Petrov, Vinyals, Dean, Hassabis, Kavukcuoglu, Farabet, Buchatskaya, Borgeaud, Fiedel, Joulin, Kenealy, Dadashi, and Andreev}]{gemmateam2024gemma2improvingopen}
Gemma Team, Morgane Riviere, Shreya Pathak, Pier~Giuseppe Sessa, Cassidy Hardin, Surya Bhupatiraju, Léonard Hussenot, Thomas Mesnard, Bobak Shahriari, Alexandre Ramé, Johan Ferret, Peter Liu, Pouya Tafti, Abe Friesen, Michelle Casbon, Sabela Ramos, Ravin Kumar, Charline~Le Lan, Sammy Jerome, Anton Tsitsulin, Nino Vieillard, Piotr Stanczyk, Sertan Girgin, Nikola Momchev, Matt Hoffman, Shantanu Thakoor, Jean-Bastien Grill, Behnam Neyshabur, Olivier Bachem, Alanna Walton, Aliaksei Severyn, Alicia Parrish, Aliya Ahmad, Allen Hutchison, Alvin Abdagic, Amanda Carl, Amy Shen, Andy Brock, Andy Coenen, Anthony Laforge, Antonia Paterson, Ben Bastian, Bilal Piot, Bo~Wu, Brandon Royal, Charlie Chen, Chintu Kumar, Chris Perry, Chris Welty, Christopher~A. Choquette-Choo, Danila Sinopalnikov, David Weinberger, Dimple Vijaykumar, Dominika Rogozińska, Dustin Herbison, Elisa Bandy, Emma Wang, Eric Noland, Erica Moreira, Evan Senter, Evgenii Eltyshev, Francesco Visin, Gabriel Rasskin, Gary Wei, Glenn Cameron, Gus Martins,
  Hadi Hashemi, Hanna Klimczak-Plucińska, Harleen Batra, Harsh Dhand, Ivan Nardini, Jacinda Mein, Jack Zhou, James Svensson, Jeff Stanway, Jetha Chan, Jin~Peng Zhou, Joana Carrasqueira, Joana Iljazi, Jocelyn Becker, Joe Fernandez, Joost van Amersfoort, Josh Gordon, Josh Lipschultz, Josh Newlan, Ju~yeong Ji, Kareem Mohamed, Kartikeya Badola, Kat Black, Katie Millican, Keelin McDonell, Kelvin Nguyen, Kiranbir Sodhia, Kish Greene, Lars~Lowe Sjoesund, Lauren Usui, Laurent Sifre, Lena Heuermann, Leticia Lago, Lilly McNealus, Livio~Baldini Soares, Logan Kilpatrick, Lucas Dixon, Luciano Martins, Machel Reid, Manvinder Singh, Mark Iverson, Martin Görner, Mat Velloso, Mateo Wirth, Matt Davidow, Matt Miller, Matthew Rahtz, Matthew Watson, Meg Risdal, Mehran Kazemi, Michael Moynihan, Ming Zhang, Minsuk Kahng, Minwoo Park, Mofi Rahman, Mohit Khatwani, Natalie Dao, Nenshad Bardoliwalla, Nesh Devanathan, Neta Dumai, Nilay Chauhan, Oscar Wahltinez, Pankil Botarda, Parker Barnes, Paul Barham, Paul Michel, Pengchong Jin,
  Petko Georgiev, Phil Culliton, Pradeep Kuppala, Ramona Comanescu, Ramona Merhej, Reena Jana, Reza~Ardeshir Rokni, Rishabh Agarwal, Ryan Mullins, Samaneh Saadat, Sara~Mc Carthy, Sarah Cogan, Sarah Perrin, Sébastien M.~R. Arnold, Sebastian Krause, Shengyang Dai, Shruti Garg, Shruti Sheth, Sue Ronstrom, Susan Chan, Timothy Jordan, Ting Yu, Tom Eccles, Tom Hennigan, Tomas Kocisky, Tulsee Doshi, Vihan Jain, Vikas Yadav, Vilobh Meshram, Vishal Dharmadhikari, Warren Barkley, Wei Wei, Wenming Ye, Woohyun Han, Woosuk Kwon, Xiang Xu, Zhe Shen, Zhitao Gong, Zichuan Wei, Victor Cotruta, Phoebe Kirk, Anand Rao, Minh Giang, Ludovic Peran, Tris Warkentin, Eli Collins, Joelle Barral, Zoubin Ghahramani, Raia Hadsell, D.~Sculley, Jeanine Banks, Anca Dragan, Slav Petrov, Oriol Vinyals, Jeff Dean, Demis Hassabis, Koray Kavukcuoglu, Clement Farabet, Elena Buchatskaya, Sebastian Borgeaud, Noah Fiedel, Armand Joulin, Kathleen Kenealy, Robert Dadashi, and Alek Andreev. 2024{\natexlab{b}}.
\newblock \href {https://arxiv.org/abs/2408.00118} {Gemma 2: Improving open language models at a practical size}.
\newblock \emph{Preprint}, arXiv:2408.00118.

\bibitem[{Tiedemann(2012)}]{tiedemann-2012-parallel}
J{\"o}rg Tiedemann. 2012.
\newblock \href {http://www.lrec-conf.org/proceedings/lrec2012/pdf/463_Paper.pdf} {Parallel data, tools and interfaces in {OPUS}}.
\newblock In \emph{Proceedings of the Eighth International Conference on Language Resources and Evaluation ({LREC}'12)}, pages 2214--2218, Istanbul, Turkey. European Language Resources Association (ELRA).

\bibitem[{Touvron et~al.(2023)Touvron, Martin, Stone, Albert, Almahairi, Babaei, Bashlykov, Batra, Bhargava, Bhosale, Bikel, Blecher, Ferrer, Chen, Cucurull, Esiobu, Fernandes, Fu, Fu, Fuller, Gao, Goswami, Goyal, Hartshorn, Hosseini, Hou, Inan, Kardas, Kerkez, Khabsa, Kloumann, Korenev, Koura, Lachaux, Lavril, Lee, Liskovich, Lu, Mao, Martinet, Mihaylov, Mishra, Molybog, Nie, Poulton, Reizenstein, Rungta, Saladi, Schelten, Silva, Smith, Subramanian, Tan, Tang, Taylor, Williams, Kuan, Xu, Yan, Zarov, Zhang, Fan, Kambadur, Narang, Rodriguez, Stojnic, Edunov, and Scialom}]{touvron2023llama2openfoundation}
Hugo Touvron, Louis Martin, Kevin Stone, Peter Albert, Amjad Almahairi, Yasmine Babaei, Nikolay Bashlykov, Soumya Batra, Prajjwal Bhargava, Shruti Bhosale, Dan Bikel, Lukas Blecher, Cristian~Canton Ferrer, Moya Chen, Guillem Cucurull, David Esiobu, Jude Fernandes, Jeremy Fu, Wenyin Fu, Brian Fuller, Cynthia Gao, Vedanuj Goswami, Naman Goyal, Anthony Hartshorn, Saghar Hosseini, Rui Hou, Hakan Inan, Marcin Kardas, Viktor Kerkez, Madian Khabsa, Isabel Kloumann, Artem Korenev, Punit~Singh Koura, Marie-Anne Lachaux, Thibaut Lavril, Jenya Lee, Diana Liskovich, Yinghai Lu, Yuning Mao, Xavier Martinet, Todor Mihaylov, Pushkar Mishra, Igor Molybog, Yixin Nie, Andrew Poulton, Jeremy Reizenstein, Rashi Rungta, Kalyan Saladi, Alan Schelten, Ruan Silva, Eric~Michael Smith, Ranjan Subramanian, Xiaoqing~Ellen Tan, Binh Tang, Ross Taylor, Adina Williams, Jian~Xiang Kuan, Puxin Xu, Zheng Yan, Iliyan Zarov, Yuchen Zhang, Angela Fan, Melanie Kambadur, Sharan Narang, Aurelien Rodriguez, Robert Stojnic, Sergey Edunov, and Thomas
  Scialom. 2023.
\newblock \href {https://arxiv.org/abs/2307.09288} {Llama 2: Open foundation and fine-tuned chat models}.
\newblock \emph{Preprint}, arXiv:2307.09288.

\bibitem[{Workshop et~al.(2023)Workshop, :, Scao, Fan, Akiki, Pavlick, Ilić, Hesslow, Castagné, Luccioni, Yvon, Gallé, Tow, Rush, Biderman, Webson, Ammanamanchi, Wang, Sagot, Muennighoff, del Moral, Ruwase, Bawden, Bekman, McMillan-Major, Beltagy, Nguyen, Saulnier, Tan, Suarez, Sanh, Laurençon, Jernite, Launay, Mitchell, Raffel, Gokaslan, Simhi, Soroa, Aji, Alfassy, Rogers, Nitzav, Xu, Mou, Emezue, Klamm, Leong, van Strien, Adelani, Radev, Ponferrada, Levkovizh, Kim, Natan, Toni, Dupont, Kruszewski, Pistilli, Elsahar, Benyamina, Tran, Yu, Abdulmumin, Johnson, Gonzalez-Dios, de~la Rosa, Chim, Dodge, Zhu, Chang, Frohberg, Tobing, Bhattacharjee, Almubarak, Chen, Lo, Werra, Weber, Phan, allal, Tanguy, Dey, Muñoz, Masoud, Grandury, Šaško, Huang, Coavoux, Singh, Jiang, Vu, Jauhar, Ghaleb, Subramani, Kassner, Khamis, Nguyen, Espejel, de~Gibert, Villegas, Henderson, Colombo, Amuok, Lhoest, Harliman, Bommasani, López, Ribeiro, Osei, Pyysalo, Nagel, Bose, Muhammad, Sharma, Longpre, Nikpoor, Silberberg, Pai,
  Zink, Torrent, Schick, Thrush, Danchev, Nikoulina, Laippala, Lepercq, Prabhu, Alyafeai, Talat, Raja, Heinzerling, Si, Taşar, Salesky, Mielke, Lee, Sharma, Santilli, Chaffin, Stiegler, Datta, Szczechla, Chhablani, Wang, Pandey, Strobelt, Fries, Rozen, Gao, Sutawika, Bari, Al-shaibani, Manica, Nayak, Teehan, Albanie, Shen, Ben-David, Bach, Kim, Bers, Fevry, Neeraj, Thakker, Raunak, Tang, Yong, Sun, Brody, Uri, Tojarieh, Roberts, Chung, Tae, Phang, Press, Li, Narayanan, Bourfoune, Casper, Rasley, Ryabinin, Mishra, Zhang, Shoeybi, Peyrounette, Patry, Tazi, Sanseviero, von Platen, Cornette, Lavallée, Lacroix, Rajbhandari, Gandhi, Smith, Requena, Patil, Dettmers, Baruwa, Singh, Cheveleva, Ligozat, Subramonian, Névéol, Lovering, Garrette, Tunuguntla, Reiter, Taktasheva, Voloshina, Bogdanov, Winata, Schoelkopf, Kalo, Novikova, Forde, Clive, Kasai, Kawamura, Hazan, Carpuat, Clinciu, Kim, Cheng, Serikov, Antverg, van~der Wal, Zhang, Zhang, Gehrmann, Mirkin, Pais, Shavrina, Scialom, Yun, Limisiewicz, Rieser,
  Protasov, Mikhailov, Pruksachatkun, Belinkov, Bamberger, Kasner, Rueda, Pestana, Feizpour, Khan, Faranak, Santos, Hevia, Unldreaj, Aghagol, Abdollahi, Tammour, HajiHosseini, Behroozi, Ajibade, Saxena, Ferrandis, McDuff, Contractor, Lansky, David, Kiela, Nguyen, Tan, Baylor, Ozoani, Mirza, Ononiwu, Rezanejad, Jones, Bhattacharya, Solaiman, Sedenko, Nejadgholi, Passmore, Seltzer, Sanz, Dutra, Samagaio, Elbadri, Mieskes, Gerchick, Akinlolu, McKenna, Qiu, Ghauri, Burynok, Abrar, Rajani, Elkott, Fahmy, Samuel, An, Kromann, Hao, Alizadeh, Shubber, Wang, Roy, Viguier, Le, Oyebade, Le, Yang, Nguyen, Kashyap, Palasciano, Callahan, Shukla, Miranda-Escalada, Singh, Beilharz, Wang, Brito, Zhou, Jain, Xu, Fourrier, Periñán, Molano, Yu, Manjavacas, Barth, Fuhrimann, Altay, Bayrak, Burns, Vrabec, Bello, Dash, Kang, Giorgi, Golde, Posada, Sivaraman, Bulchandani, Liu, Shinzato, de~Bykhovetz, Takeuchi, Pàmies, Castillo, Nezhurina, Sänger, Samwald, Cullan, Weinberg, Wolf, Mihaljcic, Liu, Freidank, Kang, Seelam, Dahlberg,
  Broad, Muellner, Fung, Haller, Chandrasekhar, Eisenberg, Martin, Canalli, Su, Su, Cahyawijaya, Garda, Deshmukh, Mishra, Kiblawi, Ott, Sang-aroonsiri, Kumar, Schweter, Bharati, Laud, Gigant, Kainuma, Kusa, Labrak, Bajaj, Venkatraman, Xu, Xu, Xu, Tan, Xie, Ye, Bras, Belkada, and Wolf}]{workshop2023bloom176bparameteropenaccessmultilingual}
BigScience Workshop, :, Teven~Le Scao, Angela Fan, Christopher Akiki, Ellie Pavlick, Suzana Ilić, Daniel Hesslow, Roman Castagné, Alexandra~Sasha Luccioni, François Yvon, Matthias Gallé, Jonathan Tow, Alexander~M. Rush, Stella Biderman, Albert Webson, Pawan~Sasanka Ammanamanchi, Thomas Wang, Benoît Sagot, Niklas Muennighoff, Albert~Villanova del Moral, Olatunji Ruwase, Rachel Bawden, Stas Bekman, Angelina McMillan-Major, Iz~Beltagy, Huu Nguyen, Lucile Saulnier, Samson Tan, Pedro~Ortiz Suarez, Victor Sanh, Hugo Laurençon, Yacine Jernite, Julien Launay, Margaret Mitchell, Colin Raffel, Aaron Gokaslan, Adi Simhi, Aitor Soroa, Alham~Fikri Aji, Amit Alfassy, Anna Rogers, Ariel~Kreisberg Nitzav, Canwen Xu, Chenghao Mou, Chris Emezue, Christopher Klamm, Colin Leong, Daniel van Strien, David~Ifeoluwa Adelani, Dragomir Radev, Eduardo~González Ponferrada, Efrat Levkovizh, Ethan Kim, Eyal~Bar Natan, Francesco~De Toni, Gérard Dupont, Germán Kruszewski, Giada Pistilli, Hady Elsahar, Hamza Benyamina, Hieu Tran,
  Ian Yu, Idris Abdulmumin, Isaac Johnson, Itziar Gonzalez-Dios, Javier de~la Rosa, Jenny Chim, Jesse Dodge, Jian Zhu, Jonathan Chang, Jörg Frohberg, Joseph Tobing, Joydeep Bhattacharjee, Khalid Almubarak, Kimbo Chen, Kyle Lo, Leandro~Von Werra, Leon Weber, Long Phan, Loubna~Ben allal, Ludovic Tanguy, Manan Dey, Manuel~Romero Muñoz, Maraim Masoud, María Grandury, Mario Šaško, Max Huang, Maximin Coavoux, Mayank Singh, Mike Tian-Jian Jiang, Minh~Chien Vu, Mohammad~A. Jauhar, Mustafa Ghaleb, Nishant Subramani, Nora Kassner, Nurulaqilla Khamis, Olivier Nguyen, Omar Espejel, Ona de~Gibert, Paulo Villegas, Peter Henderson, Pierre Colombo, Priscilla Amuok, Quentin Lhoest, Rheza Harliman, Rishi Bommasani, Roberto~Luis López, Rui Ribeiro, Salomey Osei, Sampo Pyysalo, Sebastian Nagel, Shamik Bose, Shamsuddeen~Hassan Muhammad, Shanya Sharma, Shayne Longpre, Somaieh Nikpoor, Stanislav Silberberg, Suhas Pai, Sydney Zink, Tiago~Timponi Torrent, Timo Schick, Tristan Thrush, Valentin Danchev, Vassilina Nikoulina,
  Veronika Laippala, Violette Lepercq, Vrinda Prabhu, Zaid Alyafeai, Zeerak Talat, Arun Raja, Benjamin Heinzerling, Chenglei Si, Davut~Emre Taşar, Elizabeth Salesky, Sabrina~J. Mielke, Wilson~Y. Lee, Abheesht Sharma, Andrea Santilli, Antoine Chaffin, Arnaud Stiegler, Debajyoti Datta, Eliza Szczechla, Gunjan Chhablani, Han Wang, Harshit Pandey, Hendrik Strobelt, Jason~Alan Fries, Jos Rozen, Leo Gao, Lintang Sutawika, M~Saiful Bari, Maged~S. Al-shaibani, Matteo Manica, Nihal Nayak, Ryan Teehan, Samuel Albanie, Sheng Shen, Srulik Ben-David, Stephen~H. Bach, Taewoon Kim, Tali Bers, Thibault Fevry, Trishala Neeraj, Urmish Thakker, Vikas Raunak, Xiangru Tang, Zheng-Xin Yong, Zhiqing Sun, Shaked Brody, Yallow Uri, Hadar Tojarieh, Adam Roberts, Hyung~Won Chung, Jaesung Tae, Jason Phang, Ofir Press, Conglong Li, Deepak Narayanan, Hatim Bourfoune, Jared Casper, Jeff Rasley, Max Ryabinin, Mayank Mishra, Minjia Zhang, Mohammad Shoeybi, Myriam Peyrounette, Nicolas Patry, Nouamane Tazi, Omar Sanseviero, Patrick von
  Platen, Pierre Cornette, Pierre~François Lavallée, Rémi Lacroix, Samyam Rajbhandari, Sanchit Gandhi, Shaden Smith, Stéphane Requena, Suraj Patil, Tim Dettmers, Ahmed Baruwa, Amanpreet Singh, Anastasia Cheveleva, Anne-Laure Ligozat, Arjun Subramonian, Aurélie Névéol, Charles Lovering, Dan Garrette, Deepak Tunuguntla, Ehud Reiter, Ekaterina Taktasheva, Ekaterina Voloshina, Eli Bogdanov, Genta~Indra Winata, Hailey Schoelkopf, Jan-Christoph Kalo, Jekaterina Novikova, Jessica~Zosa Forde, Jordan Clive, Jungo Kasai, Ken Kawamura, Liam Hazan, Marine Carpuat, Miruna Clinciu, Najoung Kim, Newton Cheng, Oleg Serikov, Omer Antverg, Oskar van~der Wal, Rui Zhang, Ruochen Zhang, Sebastian Gehrmann, Shachar Mirkin, Shani Pais, Tatiana Shavrina, Thomas Scialom, Tian Yun, Tomasz Limisiewicz, Verena Rieser, Vitaly Protasov, Vladislav Mikhailov, Yada Pruksachatkun, Yonatan Belinkov, Zachary Bamberger, Zdeněk Kasner, Alice Rueda, Amanda Pestana, Amir Feizpour, Ammar Khan, Amy Faranak, Ana Santos, Anthony Hevia, Antigona
  Unldreaj, Arash Aghagol, Arezoo Abdollahi, Aycha Tammour, Azadeh HajiHosseini, Bahareh Behroozi, Benjamin Ajibade, Bharat Saxena, Carlos~Muñoz Ferrandis, Daniel McDuff, Danish Contractor, David Lansky, Davis David, Douwe Kiela, Duong~A. Nguyen, Edward Tan, Emi Baylor, Ezinwanne Ozoani, Fatima Mirza, Frankline Ononiwu, Habib Rezanejad, Hessie Jones, Indrani Bhattacharya, Irene Solaiman, Irina Sedenko, Isar Nejadgholi, Jesse Passmore, Josh Seltzer, Julio~Bonis Sanz, Livia Dutra, Mairon Samagaio, Maraim Elbadri, Margot Mieskes, Marissa Gerchick, Martha Akinlolu, Michael McKenna, Mike Qiu, Muhammed Ghauri, Mykola Burynok, Nafis Abrar, Nazneen Rajani, Nour Elkott, Nour Fahmy, Olanrewaju Samuel, Ran An, Rasmus Kromann, Ryan Hao, Samira Alizadeh, Sarmad Shubber, Silas Wang, Sourav Roy, Sylvain Viguier, Thanh Le, Tobi Oyebade, Trieu Le, Yoyo Yang, Zach Nguyen, Abhinav~Ramesh Kashyap, Alfredo Palasciano, Alison Callahan, Anima Shukla, Antonio Miranda-Escalada, Ayush Singh, Benjamin Beilharz, Bo~Wang, Caio Brito,
  Chenxi Zhou, Chirag Jain, Chuxin Xu, Clémentine Fourrier, Daniel~León Periñán, Daniel Molano, Dian Yu, Enrique Manjavacas, Fabio Barth, Florian Fuhrimann, Gabriel Altay, Giyaseddin Bayrak, Gully Burns, Helena~U. Vrabec, Imane Bello, Ishani Dash, Jihyun Kang, John Giorgi, Jonas Golde, Jose~David Posada, Karthik~Rangasai Sivaraman, Lokesh Bulchandani, Lu~Liu, Luisa Shinzato, Madeleine~Hahn de~Bykhovetz, Maiko Takeuchi, Marc Pàmies, Maria~A Castillo, Marianna Nezhurina, Mario Sänger, Matthias Samwald, Michael Cullan, Michael Weinberg, Michiel~De Wolf, Mina Mihaljcic, Minna Liu, Moritz Freidank, Myungsun Kang, Natasha Seelam, Nathan Dahlberg, Nicholas~Michio Broad, Nikolaus Muellner, Pascale Fung, Patrick Haller, Ramya Chandrasekhar, Renata Eisenberg, Robert Martin, Rodrigo Canalli, Rosaline Su, Ruisi Su, Samuel Cahyawijaya, Samuele Garda, Shlok~S Deshmukh, Shubhanshu Mishra, Sid Kiblawi, Simon Ott, Sinee Sang-aroonsiri, Srishti Kumar, Stefan Schweter, Sushil Bharati, Tanmay Laud, Théo Gigant, Tomoya
  Kainuma, Wojciech Kusa, Yanis Labrak, Yash~Shailesh Bajaj, Yash Venkatraman, Yifan Xu, Yingxin Xu, Yu~Xu, Zhe Tan, Zhongli Xie, Zifan Ye, Mathilde Bras, Younes Belkada, and Thomas Wolf. 2023.
\newblock \href {https://arxiv.org/abs/2211.05100} {Bloom: A 176b-parameter open-access multilingual language model}.
\newblock \emph{Preprint}, arXiv:2211.05100.

\bibitem[{Yang et~al.(2024)Yang, Yang, Hui, Zheng, Yu, Zhou, Li, Li, Liu, Huang, Dong, Wei, Lin, Tang, Wang, Yang, Tu, Zhang, Ma, Yang, Xu, Zhou, Bai, He, Lin, Dang, Lu, Chen, Yang, Li, Xue, Ni, Zhang, Wang, Peng, Men, Gao, Lin, Wang, Bai, Tan, Zhu, Li, Liu, Ge, Deng, Zhou, Ren, Zhang, Wei, Ren, Liu, Fan, Yao, Zhang, Wan, Chu, Liu, Cui, Zhang, Guo, and Fan}]{yang2024qwen2technicalreport}
An~Yang, Baosong Yang, Binyuan Hui, Bo~Zheng, Bowen Yu, Chang Zhou, Chengpeng Li, Chengyuan Li, Dayiheng Liu, Fei Huang, Guanting Dong, Haoran Wei, Huan Lin, Jialong Tang, Jialin Wang, Jian Yang, Jianhong Tu, Jianwei Zhang, Jianxin Ma, Jianxin Yang, Jin Xu, Jingren Zhou, Jinze Bai, Jinzheng He, Junyang Lin, Kai Dang, Keming Lu, Keqin Chen, Kexin Yang, Mei Li, Mingfeng Xue, Na~Ni, Pei Zhang, Peng Wang, Ru~Peng, Rui Men, Ruize Gao, Runji Lin, Shijie Wang, Shuai Bai, Sinan Tan, Tianhang Zhu, Tianhao Li, Tianyu Liu, Wenbin Ge, Xiaodong Deng, Xiaohuan Zhou, Xingzhang Ren, Xinyu Zhang, Xipin Wei, Xuancheng Ren, Xuejing Liu, Yang Fan, Yang Yao, Yichang Zhang, Yu~Wan, Yunfei Chu, Yuqiong Liu, Zeyu Cui, Zhenru Zhang, Zhifang Guo, and Zhihao Fan. 2024.
\newblock \href {https://arxiv.org/abs/2407.10671} {Qwen2 technical report}.
\newblock \emph{Preprint}, arXiv:2407.10671.

\bibitem[{Yang et~al.(2019)Yang, Zhang, Tar, and Baldridge}]{yang-etal-2019-paws}
Yinfei Yang, Yuan Zhang, Chris Tar, and Jason Baldridge. 2019.
\newblock \href {https://doi.org/10.18653/v1/D19-1382} {{PAWS}-{X}: A cross-lingual adversarial dataset for paraphrase identification}.
\newblock In \emph{Proceedings of the 2019 Conference on Empirical Methods in Natural Language Processing and the 9th International Joint Conference on Natural Language Processing (EMNLP-IJCNLP)}, pages 3687--3692, Hong Kong, China. Association for Computational Linguistics.

\bibitem[{Zellers et~al.(2019)Zellers, Holtzman, Bisk, Farhadi, and Choi}]{DBLP:conf/acl/ZellersHBFC19}
Rowan Zellers, Ari Holtzman, Yonatan Bisk, Ali Farhadi, and Yejin Choi. 2019.
\newblock Hellaswag: Can a machine really finish your sentence?
\newblock In \emph{{ACL} {(1)}}, pages 4791--4800. Association for Computational Linguistics.

\end{thebibliography}
\clearpage

\appendix
\section{Experimental Setup}\label{appendix:experiment}
\subsection{Problematic Examples}\label{appendix:experiment:problematic_examples}
Tables \ref{fig:winogrande_base} and \ref{fig:winogrande_spa} contain a sample from the development split of WinoGrande-XL and its translation, illustrating challenges in 1:1 translation of evaluation datasets.

\begin{table}[h]
\begin{tabularx}{\linewidth}{Xcp{2cm}}
\hline
\textbf{Sentence:} The juice from the mango fruit could not fill up the cup because the \_ is small. \\
\hline
\textbf{Choices:} \\
\hline
mango \\ 
\hline
cup \\
\hline
\end{tabularx}
\caption{Development split instance of the WinoGrande-XL dataset.}
\label{fig:winogrande_base}
\end{table}

\begin{table}[h]
\begin{tabularx}{\linewidth}{Xcp{2cm}}
\hline
\textbf{Choices:} \\
\hline
El zumo de la fruta de mango no podía llenar la taza porque el mango es pequeño. \\ 
\hline
El zumo de la fruta de mango no podía llenar la taza porque \textcolor{red}{la} taza es pequeñ\textcolor{red}{a}. \\
\hline
\end{tabularx}
\caption{Potential Spanish translation of the instance in Table \ref{fig:winogrande_base}. Highlighted in red are differences between the two choices beyond replacement of the subject.}
\label{fig:winogrande_spa}
\end{table}

\begin{table}[!htb]
\begin{tabularx}{\linewidth}{Xcp{2cm}}
\hline
\textbf{Context:} A cartoon animation video is shown with people wandering around and rockets being shot. two men\\
\hline
\textbf{Endings:} \\
\hline
1. fight robots of evil and ends with a to be continued. \\ 
\hline
2. are then shown in closeups shooting a shot put. \\ 
\hline
3. push a child in a speedboat in the water. \\ 
\hline
4. look in the cameraman's eye and smile. \\ 
\hline
\end{tabularx}
\caption{An English HellaSwag sample, which includes a context (a cartoon animation video) and four possible endings. The task requires models to select the most appropriate ending based on the context.}
\label{fig:hellaswag_base}
\end{table}

\begin{table}[!htb]
\begin{tabularx}{\linewidth}{Xcp{2cm}}
\hline
\textbf{Context:} Es wird ein Zeichentrickvideo gezeigt, in dem Menschen umherwandern und Raketen geschossen werden. zwei Männer \\
\hline
\textbf{Endings:} \\
\hline
1. Kampfroboter des Bösen und endet mit einem "to be continued" \\ 
\hline
2. werden dann in Nahaufnahmen beim Kugelstoßen gezeigt. \\ 
\hline
3. ein Kind in einem Schnellboot ins Wasser schieben. \\ 
\hline
4. in die Augen des Kameramanns schauen und lächeln. \\ 
\hline
\end{tabularx}
\caption{Naive, context-unaware translation into German of the sample shown in Table \ref{fig:hellaswag_base}.  This literal translation demonstrates that without considering the context, certain phrases such as "Raketen geschossen" (rockets shot) lose fluency and clarity in translation, affecting the overall coherence of the text.}
\label{fig:hellaswag_de_naive}
\end{table}

\begin{figure}[!htb]
    \centering
    \begin{verbatim}
    'A cartoon animation video is shown 
    with people wandering around and
    rockets being shot. two men
    <x>SEP</x>fight robots
    of evil and ends with a to be
    continued.<x>SEP</x>are then shown
    in closeupsshooting a shot put.
    <x>SEP</x>look in the cameraman's
    eye and smile.'
\end{verbatim}
    \caption{Processed HellaSwag instance, as supplied to the DeepL-API.The original context and endings are formatted in a machine-readable way, ensuring consistent input for translation.}
    \label{fig:hellaswag_string}
\end{figure}

\begin{table}[!htb]
\begin{tabularx}{\linewidth}{Xcp{2cm}}
\hline
\textbf{Context:} Es wird ein Zeichentrickfilm gezeigt, in dem Menschen umherlaufen
und Raketen abgeschossen werden. zwei Männer' \\
\hline
\textbf{Endings:} \\
\hline
1. kämpfen gegen Roboter des Bösen und enden mit einem "to be continued". \\ 
\hline
2. werden dann in Großaufnahme beim Kugelstoßen gezeigt. \\ 
\hline
3. schieben ein Kind in einem Schnellboot im Wasser.\\ 
\hline
4. schauen dem Kameramann in die Augen und lächeln. \\ 
\hline
\end{tabularx}
\caption{Context-aware translation into German of the sample shown in Table \ref{fig:hellaswag_base}.  In contrast to the naive translation, this version properly adapts phrases, such as translating "rockets being shot" to "Raketen abgeschossen" (rockets launched), maintaining fluency and better capturing the meaning of the original text.}
\label{fig:hellaswag_de_proper}
\end{table}
\twocolumn

\onecolumn
\section{Evaluation Results}\label{appendix:evaluation}
\subsection{Model overview}\label{appendix:evaluation:model_overview}
\begin{longtable}[htbp]{llccl}
\caption{Model Overview; Legend: D: Used in downstream evaluation (cf. \Cref{sec:evaluation}); T: Used in OKAPI translation comparison (cf. \Cref{sec:okapi}); C: Used in LMSYS correlation analysis (cf. \Cref{sec:human_pref})}
\label{tab:eval_model_overview}
\\
\toprule
\textbf{Model}                    & D         & T      & C      & \textbf{Hugging Face Link} (\texttt{https://hf.co/...}) \\
\midrule
\endhead
\midrule
\multicolumn{2}{r}{Continued on next page} \\
\midrule
\endfoot
\bottomrule
\endlastfoot

Aya-23-8B                         & \checkmark & \checkmark &            & \href{https://hf.co/CohereForAI/aya-23-8B}{CohereForAI/aya-23-8B} \\
Bloom-7b1                         & \checkmark & \checkmark &            & \href{https://hf.co/bigscience/bloom-7b1}{bigscience/bloom-7b1} \\
Bloomz-7b1                        & \checkmark & \checkmark &            & \href{https://hf.co/bigscience/bloomz-7b1}{bigscience/bloomz-7b1} \\
c4ai-command-r-35B-v01            & \checkmark &            & \checkmark & \href{https://hf.co/CohereForAI/c4ai-command-r-v01}{CohereForAI/c4ai-command-r-v01} \\
EuroLLM-1.7B                      & \checkmark &            &            & \href{https://hf.co/utter-project/EuroLLM-1.7B}{utter-project/EuroLLM-1.7B} \\
EuroLLM-1.7B-Instruct             & \checkmark &            &            & \href{https://hf.co/utter-project/EuroLLM-1.7B-Instruct}{utter-project/EuroLLM-1.7B-Instruct} \\
Gemma-7b                          & \checkmark & \checkmark &            & \href{https://hf.co/google/gemma-7b-it}{google/gemma-7b-it} \\
Gemma-1.1-7b-Instruct             & \checkmark & \checkmark & \checkmark & \href{https://hf.co/google/gemma-1.1-7b-it}{google/gemma-1.1-7b-it} \\
Gemma-2-9b-Instruct               & \checkmark &            & \checkmark & \href{https://hf.co/google/gemma-2-9b-it}{google/gemma-2-9b-it} \\
Gemma-2-27b-Instruct              & \checkmark &            & \checkmark & \href{https://hf.co/google/gemma-2-27b-it}{google/gemma-2-27b-it} \\
Meta-Llama-2-7B                   & \checkmark & \checkmark &            & \href{https://hf.co/meta-llama/Llama-2-7b-hf}{meta-llama/Llama-2-7b-hf} \\
Meta-Llama-2-7B-Chat              & \checkmark & \checkmark & \checkmark & \href{https://hf.co/meta-llama/Llama-2-7b-chat-hf}{meta-llama/Llama-2-7b-chat-hf} \\
Meta-Llama-2-13B-Chat             & \checkmark &            & \checkmark & \href{https://hf.co/meta-llama/Llama-2-13b-chat-hf}{meta-llama/Llama-2-13b-chat-hf} \\
Meta-Llama-3-8B                   & \checkmark & \checkmark &            & \href{https://hf.co/meta-llama/Meta-Llama-3-8B}{meta-llama/Meta-Llama-3-8B} \\
Meta-Llama-3-8B-Instruct          & \checkmark & \checkmark & \checkmark & \href{https://hf.co/meta-llama/Meta-Llama-3-8B-Instruct}{meta-llama/Meta-Llama-3-8B-Instruct} \\
Meta-Llama-3.1-8B                 & \checkmark & \checkmark &            & \href{https://hf.co/meta-llama/Llama-3.1-8B}{meta-llama/Llama-3.1-8B} \\
Meta-Llama-3.1-8B-Instruct        & \checkmark & \checkmark & \checkmark & \href{https://hf.co/meta-llama/Llama-3.1-8B-Instruct}{meta-llama/Llama-3.1-8B-Instruct} \\
Meta-Llama-3.1-70B-Instruct       & \checkmark &            & \checkmark & \href{https://hf.co/meta-llama/Llama-3.1-70B-Instruct}{meta-llama/Llama-3.1-70B-Instruct} \\
Mistral-7B-v0.1                   & \checkmark & \checkmark &            & \href{https://hf.co/mistralai/Mistral-7B-v0.3}{mistralai/Mistral-7B-v0.3} \\
Mistral-7B-v0.3                   & \checkmark & \checkmark &            & \href{https://hf.co/mistralai/Mistral-7B-v0.3}{mistralai/Mistral-7B-v0.3} \\
Mistral-7B-Instruct-v0.1          & \checkmark & \checkmark & \checkmark & \href{https://hf.co/mistralai/Mistral-7B-Instruct-v0.1}{mistralai/Mistral-7B-Instruct-v0.1} \\
Mistral-7B-Instruct-v0.2          & \checkmark & \checkmark & \checkmark & \href{https://hf.co/mistralai/Mistral-7B-Instruct-v0.2}{mistralai/Mistral-7B-Instruct-v0.2} \\
Mistral-7B-Instruct-v0.3          & \checkmark & \checkmark &            & \href{https://hf.co/mistralai/Mistral-7B-Instruct-v0.3}{mistralai/Mistral-7B-Instruct-v0.3} \\
Mistral-Nemo-Base-12.2B\_2407     & \checkmark &            &            & \href{https://hf.co/mistralai/Mistral-Nemo-Instruct-2407}{mistralai/Mistral-Nemo-Instruct-2407} \\
Mistral-Nemo-Instruct-12.2B\_2407 & \checkmark &            &            & \href{https://hf.co/mistralai/Mistral-Nemo-Instruct-2407}{mistralai/Mistral-Nemo-Instruct-2407} \\
Mistral-NeMo-Minitron-8B-Base     & \checkmark &            &            & \href{https://hf.co/nvidia/Mistral-NeMo-Minitron-8B-Base}{nvidia/Mistral-NeMo-Minitron-8B-Base} \\
Mixtral-8x7B-v0.1                 & \checkmark &            &            & \href{https://hf.co/mistralai/Mixtral-8x7B-v0.1}{mistralai/Mixtral-8x7B-v0.1} \\
Mixtral-8x7B-Instruct-v0.1        & \checkmark &            & \checkmark & \href{https://hf.co/mistralai/Mixtral-8x7B-Instruct-v0.1}{mistralai/Mixtral-8x7B-Instruct-v0.1} \\
Occiglot-7b-eu5                   & \checkmark & \checkmark &            & \href{https://hf.co/occiglot/occiglot-7b-eu5}{occiglot/occiglot-7b-eu5} \\
Occiglot-7b-eu5-Instruct          & \checkmark & \checkmark &            & \href{https://hf.co/occiglot/occiglot-7b-eu5-instruct}{occiglot/occiglot-7b-eu5-instruct} \\
Pharia-1-LLM-7B-control           & \checkmark &            &            & \href{https://hf.co/Aleph-Alpha/Pharia-1-LLM-7B-control-hf}{Aleph-Alpha/Pharia-1-LLM-7B-control-hf} \\
Pharia-1-LLM-7B-control-aligned   & \checkmark &            &            & \href{https://hf.co/Aleph-Alpha/Pharia-1-LLM-7B-control-aligned-hf}{Aleph-Alpha/Pharia-1-LLM-7B-control-aligned-hf} \\
Phi-3-medium-14B-4k-Instruct      & \checkmark &            & \checkmark & \href{https://hf.co/microsoft/Phi-3-medium-4k-instruct}{microsoft/Phi-3-medium-4k-instruct} \\
Phi-3-medium-14B-128k-Instruct    & \checkmark &            &            & \href{https://hf.co/microsoft/Phi-3-mini-128k-instruct}{microsoft/Phi-3-mini-128k-instruct} \\
Phi-3-mini-4k-Instruct            & \checkmark & \checkmark & \checkmark & \href{https://hf.co/microsoft/Phi-3-mini-4k-instruct}{microsoft/Phi-3-mini-4k-instruct} \\
Phi-3-mini-3.8B-128k-Instruct     & \checkmark &            & \checkmark & \href{https://hf.co/microsoft/Phi-3-mini-128k-instruct}{microsoft/Phi-3-mini-128k-instruct} \\
Qwen2-7B                          & \checkmark &            &            & \href{https://hf.co/Qwen/Qwen2-7B}{Qwen/Qwen2-7B} \\
Qwen2-7B-Instruct                 & \checkmark &            &            & \href{https://hf.co/Qwen/Qwen2-7B-Instruct}{Qwen/Qwen2-7B-Instruct} \\
Vicuna-13b-v1.5                   & \checkmark &            & \checkmark & \href{https://hf.co/lmsys/vicuna-13b-v1.5}{lmsys/vicuna-13b-v1.5} \\
Vicuna-33b-v1.3                   & \checkmark &            & \checkmark & \href{https://hf.co/lmsys/vicuna-33b-v1.3}{lmsys/vicuna-33b-v1.3} \\
\bottomrule
\end{longtable}

\twocolumn
Table~\ref{tab:eval_model_overview} contains an overview of the models used in the different evaluations performed in this paper.
The columns D, T and C indicate whether the respective model was used in \cref{sec:evaluation}, \ref{sec:okapi} and \ref{sec:human_pref}, respectively.
The URLs for to the Hugging Face repositories of the models had to be shortened due to page width constraints and have to be prefixed with \texttt{https://hf.co}.

\subsection{Accuracy Variations}
\label{appendix:evaluation:accvar}
Notable accuracy variations were observed when evaluating the performance of language models (LMs) using different versions of the LM Eval Harness (v0.4.1, v0.4.3, and 42dc24/main).
The evaluations described in the following focused on the French version of the ARC-challenge dataset, and the findings highlight key differences and potential causes of these discrepancies.
The results were conducted using Mistral-7B-v0.1, and accuracy was assessed across different batch sizes and floating-point formats.
The model evaluations used batch sizes of 30, 32, and 60.
These are the key observations and the numbers that underline the findings:

\paragraph{Accuracy Variations Across Versions:}
Small but consistent discrepancies in accuracies were noted across the different versions.
For instance, the accuracies in v0.4.1 and v0.4.3 for a batch size of 30 were both 40.27\% for the ARC challenge, while the normalized accuracy varied slightly, i.e.~v0.4.1 46.59\% and v0.4.3 46.41\%.
When using a larger batch size of 60, both versions again reported an accuracy of 40.27\%, with a normalized accuracy of v0.4.1 46.50\% and v0.4.3 46.59\%

\begin{table}[ht]
    \centering
    \begin{tabular}{llclcc}
        \hline
        \textbf{Base Version}  & \textbf{BS}  & \textbf{Acc\_Norm} & \textbf{Acc} \\
        \hline
        v0.4.1  & 30 &  0.46587 & 0.40273 \\
        v0.4.1  & 32 &  0.46587 & 0.40188 \\
        v0.4.1  & 60 &  0.46502 & 0.40273 \\
        v0.4.3  & 30 &  0.46502 & 0.40273 \\
        v0.4.3  & 32 &  0.46416 & 0.40188 \\
        v0.4.3  & 60 &  0.46587 & 0.40273 \\
        42dc24  & 30 &  0.46502 & 0.40273 \\
        \hline
    \end{tabular}
    \caption{Evaluation results with different LM Eval Harness base versions, batch sizes, and metrics.}
    \label{tab:eval_results}
\end{table}

\paragraph{Potential Causes for Discrepancies:}
These differences suggest that changes in batch size and GPU counts influenced how padding samples were handled, causing minor variations in accuracy. This observation is consistent with the findings of other research groups using the LM Eval Harness for benchmarking \footnote{\url{https://huggingface.co/spaces/open-llm-leaderboard-old/open_llm_leaderboard}}.
Differences in the usage of random number generators in few-shot settings were noted, particularly between v0.4.1 and v0.4.3, which could affect model accuracy.
In zero-shot settings, the accuracy difference was minimal, and evaluating in a few-shot setting did not significantly impact the outcome.
Based on a manual code inspection, we suspect that few-shot results may vary depending on what other tasks are evaluated in the same run.
Furthermore, sample logs revealed that in data-parallel mode, the same sequence of few-shot contexts is used in each process, such that only the actual test sample differed between the $i$-th samples of each process.
This might possibly increase the standard error depending on how many processes (GPUs) are used as statistical noise caused by the specific selection of few-shot contexts might be amplified.
We also compared float16 and float32 runs, and results showed no significant differences in accuracy between these formats, e.g.~float16 accuracy 40.53\% (in both versions) and float32 accuracy 40.44\% (in both versions).

\paragraph{Key Differences Identified:}
A primary source of accuracy discrepancies was found to be the changed reordering and collation of batches in v0.4.3.
The accuracy results for reverse-patched reordering matched the v0.4.1 results, confirming that the reordering was the cause in this setting.
Specifically, the v0.4.3 reverse-patched reordering exhibited 40.27\% Accuracy and a Normalized Accuracy of 46.59\% (exactly matching v0.4.1)
Task specification fields previously treated as plain texts were re-interpreted as Jinja2 templates in v0.4.3, so trailing newlines that had been present in v0.4.1 were removed, leading to a slight accuracy shift in MMLU tasks.
After patching, accuracy matched across both versions.

\paragraph{Further Investigation:}
Few-shot settings showed matching results after reordering and newline patches were applied, ensuring consistency between v0.4.1 and v0.4.3 in few-shot evaluations.
When multiplying the accuracies by the number of samples in the arc\_challenge\_FR test split (1172 samples), the higher accuracy resulted in 472 correct answers, while the lower accuracy yielded 471 correct answers.
This marginal difference (0.085\%) may be explained by numerical differences between log-probabilities during batching and ultimately the discontinuous nature of the Accuracy metric.

\paragraph{Conclusion:}
The differences in batching and Jinja2 templating in v0.4.3 explain the accuracy variations observed in this experiment.
The patched version of v0.4.3 aligns with v0.4.1 when these issues are addressed, suggesting that changes in batching or task description formatting may lead to non-deterministic results across different hardware or LM Eval Harness versions.
While the accuracy variations were small (less than 1\%), the discrepancies were linked to batching behavior and templating in v0.4.3.
Ensuring consistency across these factors is crucial for accurate model evaluations, especially when comparing models across different versions.
It should be a development goal for LLM evaluation frameworks that, except for padding, the inputs actually passed to the model under test should not vary depending on batch size, degree of data or model parallelism or other tasks evaluated at the same time.

\subsection{Downstream results tables}
\label{appendix:evaluation:downstream_results}
Tables~\ref{tab:acc_std_21lang_1} and \ref{tab:acc_std_21lang_2} contain model- and task-wise arithmetic means and population standard deviations across the EU21 languages, and the average of these means.
Tables~\ref{tab:acc_std_lang_group_germanic_1}, \ref{tab:acc_std_lang_group_germanic_2}, \ref{tab:acc_std_lang_group_romance_1}, \ref{tab:acc_std_lang_group_romance_2}, \ref{tab:acc_std_lang_group_slavic_1}, and \ref{tab:acc_std_lang_group_slavic_2} contain the same information across the languages of the Germanic, Romance and Slavic language families instead of the EU21 languages.
\Cref{tab:diff_med_high} lists the differences of the per-task mean accuracies of the high-resource languages minus the corresponding mean accuracies of the medium-resource languages according to \cref{tab:language-statistics}.

\onecolumn
\begin{longtable}[c]{llll}
\caption{Performance of Models Across Tasks: Accuracy and Standard Deviation for 21 Languages}\\
\label{tab:acc_std_21lang_1}\\
\hline
                    & \multicolumn{3}{c}{\textbf{Task Group}} \\
\textbf{Model Name} & \textbf{Avg.} & \textbf{EU21-ARC} & \textbf{EU21-HeSw} \\
\hline
\endfirsthead

\hline
\multicolumn{4}{c}%
{{\bfseries \tablename\ \thetable{} -- continued from previous page}} \\
\hline
                    & \multicolumn{3}{c}{\textbf{Task Group}} \\
\textbf{Model Name} & \textbf{Avg.} & \textbf{EU21-ARC} & \textbf{EU21-HeSw} \\
\hline
\endhead

\hline
\multicolumn{4}{c}{{Continued on next page}} \\
\hline
\endfoot

\hline
\endlastfoot
Aya-23-8B \cite{aryabumi2024aya23openweight} & .441 & .475 ± .134 & .535 ± .145 \\
Bloom-7b1 \cite{workshop2023bloom176bparameteropenaccessmultilingual}& .280 & .319 ± .098 & .355 ± .112 \\
Bloomz-7b1 \cite{muennighoff2023crosslingualgeneralizationmultitaskfinetuning}& .287 & .316 ± .101 & .354 ± .115 \\
c4ai-command-r-35B-v01 & .557 & .592 ± .130 & .646 ± .135 \\
EuroLLM-1.7B \cite{martins2024eurollmmultilinguallanguagemodels} & .338 & .469 ± .036 & .485 ± .038 \\
EuroLLM-1.7B-Instruct \cite{martins2024eurollmmultilinguallanguagemodels} & .358 & .486 ± .038 & .495 ± .039 \\
Gemma-1.1-7b-Instruct \cite{gemmateam2024gemmaopenmodelsbased} & .373 & .443 ± .072 & .437 ± .085 \\
Gemma-2-27b-Instruct \cite{yang2024qwen2technicalreport} & .698 & .747 ± .034 & .711 ± .048 \\
Gemma-2-9b-Instruct \cite{yang2024qwen2technicalreport} & .581 & .668 ± .061 & .608 ± .069 \\
Gemma-7b \cite{gemmateam2024gemmaopenmodelsbased} & .536 & .607 ± .052 & .629 ± .066 \\
Meta-Llama-2-13B-Chat \cite{touvron2023llama2openfoundation} & .407 & .462 ± .116 & .504 ± .125 \\
Meta-Llama-2-7B \cite{touvron2023llama2openfoundation}& .362 & .447 ± .108 & .482 ± .114 \\
Meta-Llama-2-7B-Chat \cite{touvron2023llama2openfoundation} & .370 & .444 ± .110 & .457 ± .114 \\
Meta-Llama-3-8B \cite{dubey2024llama3herdmodels} & .503 & .550 ± .074 & .587 ± .089 \\
Meta-Llama-3-8B-Instruct \cite{dubey2024llama3herdmodels} & .547 & .558 ± .080 & .539 ± .091 \\
Meta-Llama-3.1-70B-Instruct \cite{dubey2024llama3herdmodels} & .706 & .712 ± .040 & .728 ± .059 \\
Meta-Llama-3.1-8B \cite{dubey2024llama3herdmodels} & .512 & .554 ± .071 & .588 ± .090 \\
Meta-Llama-3.1-8B-Instruct \cite{dubey2024llama3herdmodels} & .562 & .563 ± .075 & .579 ± .089 \\
Mistral-7B-Instruct-v0.1 \cite{jiang2023mistral7b} & .400 & .446 ± .121 & .454 ± .113 \\
Mistral-7B-Instruct-v0.2 \cite{jiang2023mistral7b} & .475 & .524 ± .135 & .539 ± .133 \\
Mistral-7B-Instruct-v0.3 \cite{jiang2023mistral7b} & .482 & .530 ± .136 & .538 ± .129 \\
Mistral-7B-v0.1 \cite{jiang2023mistral7b} & .458 & .521 ± .130 & .539 ± .125 \\
Mistral-7B-v0.3 \cite{jiang2023mistral7b} & .450 & .513 ± .128 & .534 ± .124 \\
Mistral-NeMo-Minitron-8B-Base \cite{sreenivas2024llmpruningdistillationpractice} & .502 & .522 ± .114 & .546 ± .120 \\
Mistral-Nemo-Base-12.2B\_2407 & .562 & .615 ± .071 & .641 ± .094 \\
Mistral-Nemo-Instruct-12.2B\_2407 & .598 & .621 ± .071 & .624 ± .093 \\
Mixtral-8x7B-Instruct-v0.1 \cite{jiang2024mixtralexperts} & .591 & .625 ± .138 & .641 ± .137 \\
Mixtral-8x7B-v0.1 \cite{jiang2024mixtralexperts} & .554 & .608 ± .132 & .636 ± .132 \\
Occiglot-7b-eu5 & .404 & .470 ± .130 & .511 ± .143 \\
Occiglot-7b-eu5-Instruct & .406 & .484 ± .132 & .519 ± .146 \\
Pharia-1-LLM-7B-control & .333 & .393 ± .174 & .433 ± .181 \\
Pharia-1-LLM-7B-control-aligned & .346 & .396 ± .177 & .438 ± .185 \\
Phi-3-medium-14B-128k-Instruct \cite{abdin2024phi3technicalreporthighly} & .508 & .497 ± .168 & .492 ± .166 \\
Phi-3-medium-14B-4k-Instruct \cite{abdin2024phi3technicalreporthighly} & .519 & .500 ± .170 & .500 ± .169 \\
Phi-3-mini-3.8B-128k-Instruct \cite{abdin2024phi3technicalreporthighly} & .424 & .415 ± .161 & .420 ± .153 \\
Phi-3-mini-4k-Instruct \cite{abdin2024phi3technicalreporthighly} & .415 & .411 ± .161 & .422 ± .158 \\
Qwen2-7B \cite{yang2024qwen2technicalreport}& .542 & .481 ± .113 & .519 ± .116 \\
Qwen2-7B-Instruct \cite{yang2024qwen2technicalreport} & .528 & .506 ± .115 & .530 ± .120 \\
Vicuna-13b-v1.5 \cite{peng2023instructiontuninggpt4} & .426 & .467 ± .111 & .519 ± .124 \\
Vicuna-33b-v1.3 \cite{peng2023instructiontuninggpt4} & .447 & .527 ± .140 & .547 ± .135 \\
\end{longtable}

\begin{longtable}[c]{llll}
\caption{Performance of Models Across Tasks: Accuracy and Standard Deviation for 21 Languages}\\
\label{tab:acc_std_21lang_2}\\
\hline
                    & \multicolumn{3}{c}{\textbf{Task Group}} \\
\textbf{Model Name} & \textbf{EU21-MMLU} & \textbf{EU21-TQA} & \textbf{EU21-GSM8k} \\
\hline
\endfirsthead

\hline
\multicolumn{4}{c}%
{{\bfseries \tablename\ \thetable{} -- continued from previous page}} \\
\hline
                    & \multicolumn{3}{c}{\textbf{Task Group}} \\
\textbf{Model Name} & \textbf{EU21-MMLU} & \textbf{EU21-TQA} & \textbf{EU21-GSM8k} \\
\hline
\endhead

\hline
\multicolumn{4}{c}{{Continued on next page}} \\
\hline
\endfoot

\hline
\endlastfoot
Aya-23-8B & .455 ± .070 & .476 ± .028 & .266 ± .105 \\
Bloom-7b1 & .256 ± .005 & .464 ± .039 & .010 ± .005 \\
Bloomz-7b1 & .302 ± .039 & .461 ± .037 & .000 ± .001 \\
c4ai-command-r-35B-v01 & .561 ± .074 & .543 ± .032 & .440 ± .085 \\
EuroLLM-1.7B & .257 ± .006 & .441 ± .027 & .035 ± .006 \\
EuroLLM-1.7B-Instruct & .266 ± .007 & .462 ± .028 & .082 ± .021 \\
Gemma-1.1-7b-Instruct & .414 ± .044 & .466 ± .026 & .106 ± .049 \\
Gemma-2-27b-Instruct & .679 ± .027 & .605 ± .021 & .748 ± .023 \\
Gemma-2-9b-Instruct & .590 ± .042 & .588 ± .022 & .448 ± .042 \\
Gemma-7b & .568 ± .032 & .479 ± .027 & .394 ± .036 \\
Meta-Llama-2-13B-Chat & .429 ± .063 & .469 ± .028 & .169 ± .081 \\
Meta-Llama-2-7B & .364 ± .050 & .438 ± .031 & .079 ± .033 \\
Meta-Llama-2-7B-Chat & .370 ± .052 & .478 ± .029 & .100 ± .052 \\
Meta-Llama-3-8B & .554 ± .049 & .477 ± .027 & .346 ± .058 \\
Meta-Llama-3-8B-Instruct & .548 ± .055 & .532 ± .024 & .560 ± .085 \\
Meta-Llama-3.1-70B-Instruct & .771 ± .026 & .573 ± .026 & .746 ± .063 \\
Meta-Llama-3.1-8B & .556 ± .047 & .495 ± .026 & .365 ± .057 \\
Meta-Llama-3.1-8B-Instruct & .576 ± .051 & .532 ± .023 & .560 ± .071 \\
Mistral-7B-Instruct-v0.1 & .424 ± .070 & .512 ± .051 & .165 ± .095 \\
Mistral-7B-Instruct-v0.2 & .476 ± .072 & .585 ± .066 & .251 ± .089 \\
Mistral-7B-Instruct-v0.3 & .491 ± .075 & .548 ± .048 & .303 ± .101 \\
Mistral-7B-v0.1 & .506 ± .077 & .475 ± .035 & .248 ± .082 \\
Mistral-7B-v0.3 & .501 ± .075 & .472 ± .034 & .231 ± .076 \\
Mistral-NeMo-Minitron-8B-Base & .555 ± .069 & .481 ± .026 & .407 ± .076 \\
Mistral-Nemo-Base-12.2B\_2407 & .602 ± .047 & .513 ± .031 & .441 ± .058 \\
Mistral-Nemo-Instruct-12.2B\_2407 & .595 ± .047 & .577 ± .027 & .571 ± .071 \\
Mixtral-8x7B-Instruct-v0.1 & .608 ± .080 & .605 ± .048 & .477 ± .108 \\
Mixtral-8x7B-v0.1 & .615 ± .079 & .488 ± .030 & .424 ± .102 \\
Occiglot-7b-eu5 & .426 ± .073 & .448 ± .030 & .163 ± .065 \\
Occiglot-7b-eu5-Instruct & .428 ± .074 & .471 ± .032 & .131 ± .055 \\
Pharia-1-LLM-7B-control & .353 ± .080 & .456 ± .029 & .030 ± .029 \\
Pharia-1-LLM-7B-control-aligned & .366 ± .088 & .469 ± .034 & .060 ± .065 \\
Phi-3-medium-14B-128k-Instruct & .565 ± .123 & .494 ± .047 & .491 ± .206 \\
Phi-3-medium-14B-4k-Instruct & .574 ± .127 & .499 ± .053 & .523 ± .202 \\
Phi-3-mini-3.8B-128k-Instruct & .462 ± .125 & .501 ± .043 & .320 ± .244 \\
Phi-3-mini-4k-Instruct & .449 ± .118 & .498 ± .052 & .292 ± .230 \\
Qwen2-7B & .593 ± .066 & .527 ± .025 & .592 ± .105 \\
Qwen2-7B-Instruct & .584 ± .065 & .553 ± .030 & .468 ± .116 \\
Vicuna-13b-v1.5 & .448 ± .067 & .505 ± .030 & .188 ± .071 \\
Vicuna-33b-v1.3 & .467 ± .078 & .524 ± .039 & .169 ± .082 \\
\end{longtable}

\begin{longtable}[c]{llll}
\caption{Standard Deviation of Task Accuracies for Germanic Languages}\\
\label{tab:acc_std_lang_group_germanic_1}\\
\hline
                    & \multicolumn{3}{c}{\textbf{Task Group}} \\
\textbf{Model Name} & \textbf{Avg.} & \textbf{EU21-ARC} & \textbf{EU21-HeSw} \\
\hline
\endfirsthead

\hline
\multicolumn{4}{c}\%
{{\bfseries \tablename\ \thetable{} -- continued from previous page}} \\
\hline
                    & \multicolumn{3}{c}{\textbf{Task Group}} \\
\textbf{Model Name} & \textbf{Avg.} & \textbf{EU21-ARC} & \textbf{EU21-HeSw} \\
\hline
\endhead

\hline
\multicolumn{4}{c}{{Continued on next page}} \\
\hline
\endfoot

\hline
\endlastfoot
Aya-23-8B & .509 & .563 ± .105 & .636 ± .121 \\
Bloom-7b1 & .291 & .348 ± .132 & .381 ± .150 \\
Bloomz-7b1 & .303 & .344 ± .138 & .379 ± .154 \\
c4ai-command-r-35B-v01 & .627 & .681 ± .076 & .750 ± .087 \\
EuroLLM-1.7B & .352 & .507 ± .046 & .518 ± .058 \\
EuroLLM-1.7B-Instruct & .377 & .520 ± .051 & .530 ± .058 \\
Gemma-1.1-7b-Instruct & .422 & .509 ± .090 & .517 ± .134 \\
Gemma-2-27b-Instruct & .727 & .776 ± .037 & .757 ± .060 \\
Gemma-2-9b-Instruct & .624 & .723 ± .053 & .674 ± .082 \\
Gemma-7b & .574 & .655 ± .053 & .694 ± .084 \\
Meta-Llama-2-13B-Chat & .485 & .566 ± .079 & .635 ± .125 \\
Meta-Llama-2-7B & .420 & .544 ± .087 & .599 ± .125 \\
Meta-Llama-2-7B-Chat & .438 & .542 ± .089 & .578 ± .140 \\
Meta-Llama-3-8B & .559 & .615 ± .069 & .679 ± .095 \\
Meta-Llama-3-8B-Instruct & .614 & .627 ± .076 & .636 ± .102 \\
Meta-Llama-3.1-70B-Instruct & .745 & .748 ± .037 & .787 ± .052 \\
Meta-Llama-3.1-8B & .568 & .618 ± .064 & .682 ± .092 \\
Meta-Llama-3.1-8B-Instruct & .624 & .633 ± .069 & .672 ± .087 \\
Mistral-7B-Instruct-v0.1 & .483 & .552 ± .088 & .566 ± .128 \\
Mistral-7B-Instruct-v0.2 & .569 & .634 ± .072 & .667 ± .120 \\
Mistral-7B-Instruct-v0.3 & .571 & .642 ± .075 & .663 ± .122 \\
Mistral-7B-v0.1 & .534 & .629 ± .072 & .658 ± .119 \\
Mistral-7B-v0.3 & .522 & .616 ± .075 & .652 ± .118 \\
Mistral-NeMo-Minitron-8B-Base & .577 & .620 ± .093 & .657 ± .119 \\
Mistral-Nemo-Base-12.2B\_2407 & .614 & .671 ± .068 & .724 ± .091 \\
Mistral-Nemo-Instruct-12.2B\_2407 & .654 & .680 ± .068 & .708 ± .094 \\
Mixtral-8x7B-Instruct-v0.1 & .676 & .727 ± .054 & .760 ± .080 \\
Mixtral-8x7B-v0.1 & .632 & .706 ± .050 & .751 ± .079 \\
Occiglot-7b-eu5 & .475 & .577 ± .087 & .637 ± .121 \\
Occiglot-7b-eu5-Instruct & .473 & .591 ± .087 & .650 ± .117 \\
Pharia-1-LLM-7B-control & .428 & .558 ± .165 & .605 ± .181 \\
Pharia-1-LLM-7B-control-aligned & .450 & .562 ± .167 & .612 ± .182 \\
Phi-3-medium-14B-128k-Instruct & .633 & .634 ± .129 & .634 ± .166 \\
Phi-3-medium-14B-4k-Instruct & .646 & .637 ± .129 & .644 ± .167 \\
Phi-3-mini-3.8B-128k-Instruct & .545 & .540 ± .174 & .538 ± .193 \\
Phi-3-mini-4k-Instruct & .536 & .543 ± .170 & .545 ± .198 \\
Qwen2-7B & .622 & .580 ± .111 & .629 ± .121 \\
Qwen2-7B-Instruct & .597 & .607 ± .101 & .644 ± .118 \\
Vicuna-13b-v1.5 & .502 & .566 ± .066 & .645 ± .111 \\
Vicuna-33b-v1.3 & .532 & .644 ± .072 & .678 ± .103 \\
\end{longtable}

\begin{longtable}[c]{llll}
\caption{Standard Deviation of Task Accuracies for Germanic Languages}\\
\label{tab:acc_std_lang_group_germanic_2}\\
\hline
                    & \multicolumn{3}{c}{\textbf{Task Group}} \\
\textbf{Model Name} & \textbf{EU21-MMLU} & \textbf{EU21-TQA} & \textbf{EU21-GSM8k} \\
\hline
\endfirsthead

\hline
\multicolumn{4}{c}\%
{{\bfseries \tablename\ \thetable{} -- continued from previous page}} \\
\hline
                    & \multicolumn{3}{c}{\textbf{Task Group}} \\
\textbf{Model Name} & \textbf{EU21-MMLU} & \textbf{EU21-TQA} & \textbf{EU21-GSM8k} \\
\hline
\endhead

\hline
\multicolumn{4}{c}{{Continued on next page}} \\
\hline
\endfoot

\hline
\endlastfoot
Aya-23-8B & .509 ± .050 & .490 ± .026 & .343 ± .077 \\
Bloom-7b1 & .257 ± .004 & .450 ± .051 & .016 ± .005 \\
Bloomz-7b1 & .321 ± .034 & .471 ± .028 & .001 ± .001 \\
EuroLLM-1.7B & .261 ± .007 & .432 ± .036 & .040 ± .005 \\
EuroLLM-1.7B-Instruct & .272 ± .007 & .458 ± .036 & .106 ± .030 \\
Gemma-1.1-7b-Instruct & .453 ± .054 & .467 ± .016 & .163 ± .075 \\
Gemma-2-27b-Instruct & .704 ± .024 & .618 ± .020 & .778 ± .004 \\
Gemma-2-9b-Instruct & .630 ± .038 & .601 ± .021 & .492 ± .043 \\
Gemma-7b & .599 ± .029 & .486 ± .022 & .438 ± .041 \\
Meta-Llama-2-13B-Chat & .490 ± .038 & .481 ± .030 & .251 ± .058 \\
Meta-Llama-2-7B & .411 ± .039 & .436 ± .035 & .108 ± .023 \\
Meta-Llama-2-7B-Chat & .422 ± .041 & .497 ± .033 & .150 ± .043 \\
Meta-Llama-3-8B & .603 ± .042 & .486 ± .040 & .414 ± .059 \\
Meta-Llama-3-8B-Instruct & .601 ± .047 & .551 ± .025 & .656 ± .070 \\
Meta-Llama-3.1-70B-Instruct & .789 ± .019 & .599 ± .023 & .804 ± .055 \\
Meta-Llama-3.1-8B & .604 ± .038 & .507 ± .044 & .431 ± .047 \\
Meta-Llama-3.1-8B-Instruct & .627 ± .042 & .550 ± .025 & .636 ± .079 \\
Mistral-7B-Instruct-v0.1 & .488 ± .046 & .555 ± .013 & .254 ± .051 \\
Mistral-7B-Instruct-v0.2 & .542 ± .042 & .657 ± .012 & .343 ± .050 \\
Mistral-7B-Instruct-v0.3 & .557 ± .047 & .595 ± .019 & .399 ± .040 \\
Mistral-7B-v0.1 & .575 ± .041 & .486 ± .048 & .323 ± .043 \\
Mistral-7B-v0.3 & .568 ± .044 & .478 ± .040 & .295 ± .039 \\
Mistral-NeMo-Minitron-8B-Base & .622 ± .045 & .497 ± .023 & .487 ± .067 \\
Mistral-Nemo-Base-12.2B\_2407 & .646 ± .036 & .529 ± .038 & .500 ± .045 \\
Mistral-Nemo-Instruct-12.2B\_2407 & .639 ± .037 & .587 ± .041 & .654 ± .060 \\
Mixtral-8x7B-Instruct-v0.1 & .670 ± .029 & .646 ± .002 & .576 ± .040 \\
Mixtral-8x7B-v0.1 & .678 ± .028 & .507 ± .019 & .519 ± .043 \\
Occiglot-7b-eu5 & .491 ± .044 & .449 ± .035 & .222 ± .037 \\
Occiglot-7b-eu5-Instruct & .496 ± .045 & .483 ± .033 & .144 ± .020 \\
Pharia-1-LLM-7B-control & .435 ± .069 & .477 ± .025 & .067 ± .034 \\
Pharia-1-LLM-7B-control-aligned & .458 ± .069 & .492 ± .015 & .128 ± .063 \\
Phi-3-medium-14B-128k-Instruct & .675 ± .078 & .533 ± .018 & .687 ± .120 \\
Phi-3-medium-14B-4k-Instruct & .688 ± .077 & .549 ± .026 & .715 ± .109 \\
Phi-3-mini-3.8B-128k-Instruct & .575 ± .102 & .532 ± .019 & .540 ± .205 \\
Phi-3-mini-4k-Instruct & .559 ± .103 & .534 ± .048 & .500 ± .216 \\
Qwen2-7B & .654 ± .045 & .547 ± .012 & .698 ± .057 \\
Qwen2-7B-Instruct & .646 ± .047 & .585 ± .014 & .504 ± .190 \\
Vicuna-13b-v1.5 & .508 ± .038 & .533 ± .022 & .257 ± .026 \\
Vicuna-33b-v1.3 & .537 ± .038 & .570 ± .025 & .233 ± .064 \\
c4ai-command-r-35B-v01 & .620 ± .044 & .570 ± .017 & .512 ± .053 \\
\end{longtable}

\begin{longtable}[c]{llll}
\caption{Standard Deviation of Task Accuracies for Romance Languages}\\
\label{tab:acc_std_lang_group_romance_1}\\
\hline
                    & \multicolumn{3}{c}{\textbf{Task Group}} \\
\textbf{Model Name} & \textbf{Avg.} & \textbf{EU21-ARC} & \textbf{EU21-HeSw} \\
\hline
\endfirsthead

\hline
\multicolumn{4}{c}\%
{{\bfseries \tablename\ \thetable{} -- continued from previous page}} \\
\hline
                    & \multicolumn{3}{c}{\textbf{Task Group}} \\
\textbf{Model Name} & \textbf{Avg.} & \textbf{EU21-ARC} & \textbf{EU21-HeSw} \\
\hline
\endhead

\hline
\multicolumn{4}{c}{{Continued on next page}} \\
\hline
\endfoot

\hline
\endlastfoot

Aya-23-8B & .522 & .600 ± .013 & .668 ± .019 \\
Bloom-7b1 & .323 & .427 ± .103 & .478 ± .125 \\
Bloomz-7b1 & .337 & .427 ± .106 & .480 ± .130 \\
c4ai-command-r-35B-v01 & .634 & .712 ± .022 & .766 ± .023 \\
EuroLLM-1.7B & .346 & .495 ± .019 & .511 ± .024 \\
EuroLLM-1.7B-Instruct & .368 & .515 ± .020 & .524 ± .025 \\
Gemma-1.1-7b-Instruct & .403 & .495 ± .031 & .488 ± .038 \\
Gemma-2-27b-Instruct & .716 & .772 ± .011 & .743 ± .015 \\
Gemma-2-9b-Instruct & .617 & .719 ± .018 & .658 ± .028 \\
Gemma-7b & .557 & .648 ± .020 & .672 ± .029 \\
Meta-Llama-2-13B-Chat & .461 & .543 ± .031 & .584 ± .045 \\
Meta-Llama-2-7B & .400 & .527 ± .028 & .557 ± .042 \\
Meta-Llama-2-7B-Chat & .414 & .529 ± .033 & .528 ± .042 \\
Meta-Llama-3-8B & .538 & .611 ± .026 & .649 ± .034 \\
Meta-Llama-3-8B-Instruct & .593 & .628 ± .026 & .601 ± .038 \\
Meta-Llama-3.1-70B-Instruct & .726 & .745 ± .013 & .772 ± .017 \\
Meta-Llama-3.1-8B & .548 & .615 ± .022 & .654 ± .033 \\
Meta-Llama-3.1-8B-Instruct & .604 & .626 ± .021 & .646 ± .033 \\
Mistral-7B-Instruct-v0.1 & .464 & .542 ± .038 & .528 ± .052 \\
Mistral-7B-Instruct-v0.2 & .540 & .625 ± .030 & .629 ± .048 \\
Mistral-7B-Instruct-v0.3 & .544 & .627 ± .028 & .623 ± .044 \\
Mistral-7B-v0.1 & .514 & .614 ± .026 & .624 ± .042 \\
Mistral-7B-v0.3 & .508 & .606 ± .029 & .619 ± .041 \\
Mistral-NeMo-Minitron-8B-Base & .568 & .631 ± .057 & .649 ± .074 \\
Mistral-Nemo-Base-12.2B\_2407 & .608 & .680 ± .043 & .726 ± .061 \\
Mistral-Nemo-Instruct-12.2B\_2407 & .644 & .687 ± .035 & .706 ± .059 \\
Mixtral-8x7B-Instruct-v0.1 & .662 & .723 ± .028 & .748 ± .045 \\
Mixtral-8x7B-v0.1 & .616 & .702 ± .031 & .736 ± .042 \\
Occiglot-7b-eu5 & .477 & .596 ± .062 & .646 ± .097 \\
Occiglot-7b-eu5-Instruct & .471 & .614 ± .060 & .656 ± .098 \\
Pharia-1-LLM-7B-control & .421 & .561 ± .139 & .603 ± .152 \\
Pharia-1-LLM-7B-control-aligned & .450 & .570 ± .139 & .614 ± .156 \\
Phi-3-medium-14B-128k-Instruct & .648 & .677 ± .106 & .655 ± .129 \\
Phi-3-medium-14B-4k-Instruct & .661 & .683 ± .111 & .665 ± .133 \\
Phi-3-mini-3.8B-128k-Instruct & .571 & .584 ± .120 & .568 ± .127 \\
Phi-3-mini-4k-Instruct & .558 & .578 ± .118 & .574 ± .131 \\
Qwen2-7B & .612 & .584 ± .057 & .617 ± .072 \\
Qwen2-7B-Instruct & .592 & .615 ± .053 & .632 ± .077 \\
Vicuna-13b-v1.5 & .480 & .547 ± .029 & .603 ± .046 \\
Vicuna-33b-v1.3 & .512 & .627 ± .025 & .641 ± .034 \\
\end{longtable}

\begin{longtable}[c]{l rl rl}
\caption{Standard Deviation of Task Accuracies for Romance Languages}\\
\label{tab:acc_std_lang_group_romance_2}\\
\hline
                    & \multicolumn{3}{c}{\textbf{Task Group}} \\
\textbf{Model Name} & \textbf{EU21-MMLU} & \textbf{EU21-TQA} & \textbf{EU21-GSM8k} \\
\hline
\endfirsthead

\hline
\multicolumn{4}{c}\%
{{\bfseries \tablename\ \thetable{} -- continued from previous page}} \\
\hline
                    & \multicolumn{3}{c}{\textbf{Task Group}} \\
\textbf{Model Name} & \textbf{EU21-MMLU} & \textbf{EU21-TQA} & \textbf{EU21-GSM8k} \\
\hline
\endhead

\hline
\multicolumn{4}{c}{{Continued on next page}} \\
\hline
\endfoot

\hline
\endlastfoot
Aya-23-8B & .515 ± .007 & .476 ± .029 & .351 ± .026 \\
Bloom-7b1 & .260 ± .003 & .440 ± .051 & .012 ± .003 \\
Bloomz-7b1 & .349 ± .035 & .432 ± .040 & .000 ± .001 \\
c4ai-command-r-35B-v01 & .627 ± .009 & .566 ± .026 & .500 ± .023 \\
EuroLLM-1.7B & .256 ± .008 & .428 ± .030 & .038 ± .006 \\
EuroLLM-1.7B-Instruct & .267 ± .006 & .448 ± .029 & .085 ± .012 \\
Gemma-1.1-7b-Instruct & .449 ± .019 & .475 ± .023 & .108 ± .022 \\
Gemma-2-27b-Instruct & .701 ± .002 & .610 ± .010 & .752 ± .021 \\
Gemma-2-9b-Instruct & .626 ± .007 & .608 ± .010 & .474 ± .040 \\
Gemma-7b & .594 ± .006 & .465 ± .033 & .406 ± .016 \\
Meta-Llama-2-13B-Chat & .474 ± .013 & .474 ± .007 & .233 ± .037 \\
Meta-Llama-2-7B & .400 ± .013 & .416 ± .025 & .101 ± .018 \\
Meta-Llama-2-7B-Chat & .407 ± .020 & .472 ± .014 & .135 ± .036 \\
Meta-Llama-3-8B & .594 ± .013 & .459 ± .021 & .376 ± .021 \\
Meta-Llama-3-8B-Instruct & .591 ± .019 & .540 ± .007 & .602 ± .045 \\
Meta-Llama-3.1-70B-Instruct & .791 ± .007 & .570 ± .014 & .751 ± .021 \\
Meta-Llama-3.1-8B & .594 ± .010 & .483 ± .021 & .396 ± .023 \\
Meta-Llama-3.1-8B-Instruct & .616 ± .013 & .542 ± .019 & .592 ± .036 \\
Mistral-7B-Instruct-v0.1 & .476 ± .020 & .544 ± .010 & .230 ± .047 \\
Mistral-7B-Instruct-v0.2 & .527 ± .012 & .617 ± .018 & .302 ± .034 \\
Mistral-7B-Instruct-v0.3 & .542 ± .013 & .568 ± .017 & .358 ± .029 \\
Mistral-7B-v0.1 & .565 ± .012 & .465 ± .031 & .303 ± .025 \\
Mistral-7B-v0.3 & .555 ± .011 & .473 ± .038 & .284 ± .029 \\
Mistral-NeMo-Minitron-8B-Base & .615 ± .026 & .481 ± .014 & .462 ± .040 \\
Mistral-Nemo-Base-12.2B\_2407 & .645 ± .018 & .504 ± .020 & .485 ± .033 \\
Mistral-Nemo-Instruct-12.2B\_2407 & .637 ± .019 & .575 ± .024 & .615 ± .046 \\
Mixtral-8x7B-Instruct-v0.1 & .671 ± .014 & .626 ± .005 & .541 ± .037 \\
Mixtral-8x7B-v0.1 & .675 ± .011 & .481 ± .026 & .487 ± .041 \\
Occiglot-7b-eu5 & .493 ± .032 & .436 ± .032 & .214 ± .042 \\
Occiglot-7b-eu5-Instruct & .495 ± .037 & .468 ± .029 & .122 ± .066 \\
Pharia-1-LLM-7B-control & .433 ± .047 & .459 ± .026 & .049 ± .022 \\
Pharia-1-LLM-7B-control-aligned & .452 ± .056 & .499 ± .013 & .117 ± .055 \\
Phi-3-medium-14B-128k-Instruct & .689 ± .060 & .544 ± .023 & .677 ± .112 \\
Phi-3-medium-14B-4k-Instruct & .702 ± .060 & .551 ± .031 & .705 ± .116 \\
Phi-3-mini-3.8B-128k-Instruct & .590 ± .079 & .546 ± .007 & .567 ± .177 \\
Phi-3-mini-4k-Instruct & .567 ± .074 & .548 ± .028 & .521 ± .164 \\
Qwen2-7B & .650 ± .021 & .539 ± .020 & .669 ± .051 \\
Qwen2-7B-Instruct & .637 ± .022 & .578 ± .012 & .500 ± .080 \\
Vicuna-13b-v1.5 & .499 ± .013 & .515 ± .022 & .235 ± .014 \\
Vicuna-33b-v1.3 & .521 ± .010 & .531 ± .012 & .240 ± .024 \\
\bottomrule
\end{longtable}

\begin{longtable}[c]{lrlrlrlrl}
\caption{Standard Deviation of Task Accuracies for Slavic Languages}\\
\label{tab:acc_std_lang_group_slavic_1}\\
\hline
                    & \multicolumn{3}{c}{\textbf{Task Group}} \\
\textbf{Model Name} & \textbf{Avg.} & \textbf{EU21-ARC} & \textbf{EU21-HeSw} \\
\hline
\endfirsthead

\hline
\multicolumn{4}{c}\%
{{\bfseries \tablename\ \thetable{} -- continued from previous page}} \\
\hline
                    & \multicolumn{3}{c}{\textbf{Task Group}} \\
\textbf{Model Name} & \textbf{Avg.} & \textbf{EU21-ARC} & \textbf{EU21-HeSw} \\
\hline
\endhead

\hline
\multicolumn{4}{c}{{Continued on next page}} \\
\hline
\endfoot

\hline
\endlastfoot

Aya-23-8B & .412 & .427 ± .097 & .477 ± .105 \\
Bloom-7b1 & .259 & .263 ± .007 & .295 ± .002 \\
Bloomz-7b1 & .262 & .259 ± .013 & .294 ± .004 \\
c4ai-command-r-35B-v01 & .535 & .566 ± .079 & .603 ± .087 \\
EuroLLM-1.7B & .332 & .455 ± .013 & .472 ± .010 \\
EuroLLM-1.7B-Instruct & .354 & .474 ± .013 & .481 ± .012 \\
Gemma-1.1-7b-Instruct & .365 & .427 ± .017 & .417 ± .024 \\
Gemma-2-27b-Instruct & .694 & .742 ± .016 & .695 ± .020 \\
Gemma-2-9b-Instruct & .567 & .658 ± .027 & .586 ± .032 \\
Gemma-7b & .530 & .600 ± .015 & .607 ± .025 \\
Meta-Llama-2-13B-Chat & .392 & .450 ± .052 & .473 ± .045 \\
Meta-Llama-2-7B & .351 & .435 ± .043 & .449 ± .036 \\
Meta-Llama-2-7B-Chat & .355 & .424 ± .046 & .419 ± .043 \\
Meta-Llama-3-8B & .482 & .528 ± .029 & .541 ± .034 \\
Meta-Llama-3-8B-Instruct & .522 & .533 ± .029 & .493 ± .035 \\
Meta-Llama-3.1-70B-Instruct & .680 & .699 ± .019 & .696 ± .032 \\
Meta-Llama-3.1-8B & .492 & .535 ± .026 & .544 ± .036 \\
Meta-Llama-3.1-8B-Instruct & .541 & .543 ± .026 & .533 ± .037 \\
Mistral-7B-Instruct-v0.1 & .396 & .442 ± .052 & .432 ± .043 \\
Mistral-7B-Instruct-v0.2 & .478 & .541 ± .050 & .530 ± .050 \\
Mistral-7B-Instruct-v0.3 & .486 & .544 ± .055 & .528 ± .048 \\
Mistral-7B-v0.1 & .461 & .536 ± .052 & .528 ± .045 \\
Mistral-7B-v0.3 & .455 & .528 ± .053 & .523 ± .048 \\
Mistral-NeMo-Minitron-8B-Base & .475 & .488 ± .043 & .496 ± .039 \\
Mistral-Nemo-Base-12.2B\_2407 & .546 & .597 ± .024 & .601 ± .029 \\
Mistral-Nemo-Instruct-12.2B\_2407 & .581 & .605 ± .030 & .585 ± .033 \\
Mixtral-8x7B-Instruct-v0.1 & .598 & .644 ± .050 & .629 ± .052 \\
Mixtral-8x7B-v0.1 & .560 & .626 ± .052 & .625 ± .048 \\
Occiglot-7b-eu5 & .380 & .443 ± .038 & .457 ± .029 \\
Occiglot-7b-eu5-Instruct & .396 & .457 ± .040 & .464 ± .032 \\
Pharia-1-LLM-7B-control & .263 & .263 ± .009 & .305 ± .008 \\
Pharia-1-LLM-7B-control-aligned & .265 & .263 ± .014 & .307 ± .007 \\
Phi-3-medium-14B-128k-Instruct & .449 & .423 ± .035 & .402 ± .028 \\
Phi-3-medium-14B-4k-Instruct & .459 & .423 ± .033 & .406 ± .027 \\
Phi-3-mini-3.8B-128k-Instruct & .342 & .318 ± .026 & .329 ± .018 \\
Phi-3-mini-4k-Instruct & .330 & .311 ± .030 & .323 ± .015 \\
Qwen2-7B & .526 & .451 ± .049 & .481 ± .052 \\
Qwen2-7B-Instruct & .519 & .474 ± .053 & .489 ± .056 \\
Vicuna-13b-v1.5 & .414 & .454 ± .054 & .489 ± .044 \\
Vicuna-33b-v1.3 & .443 & .544 ± .058 & .531 ± .050 \\

\end{longtable}

\begin{longtable}[c]{l rl rl}
\caption{Standard Deviation of Task Accuracies for Slavic Languages}\\
\label{tab:acc_std_lang_group_slavic_2}\\
\hline
                    & \multicolumn{3}{c}{\textbf{Task Group}} \\
\textbf{Model Name} & \textbf{EU21-MMLU} & \textbf{EU21-TQA} & \textbf{EU21-GSM8k} \\
\hline
\endfirsthead

\hline
\multicolumn{4}{c}\%
{{\bfseries \tablename\ \thetable{} -- continued from previous page}} \\
\hline
                    & \multicolumn{3}{c}{\textbf{Task Group}} \\
\textbf{Model Name} & \textbf{EU21-MMLU} & \textbf{EU21-TQA} & \textbf{EU21-GSM8k} \\
\hline
\endhead

\hline
\multicolumn{4}{c}{{Continued on next page}} \\
\hline
\endfoot

\hline
\endlastfoot

Aya-23-8B & .436 ± .047 & .477 ± .030 & .244 ± .056 \\
Bloom-7b1 & .255 ± .006 & .475 ± .030 & .007 ± .001 \\
Bloomz-7b1 & .273 ± .016 & .483 ± .014 & .001 ± .000 \\
c4ai-command-r-35B-v01 & .540 ± .045 & .542 ± .021 & .424 ± .053 \\
EuroLLM-1.7B & .255 ± .005 & .445 ± .026 & .033 ± .004 \\
EuroLLM-1.7B-Instruct & .264 ± .009 & .469 ± .037 & .083 ± .006 \\
Gemma-1.1-7b-Instruct & .407 ± .010 & .458 ± .029 & .117 ± .020 \\
Gemma-2-27b-Instruct & .672 ± .012 & .611 ± .016 & .749 ± .019 \\
Gemma-2-9b-Instruct & .577 ± .009 & .587 ± .004 & .428 ± .021 \\
Gemma-7b & .558 ± .015 & .495 ± .030 & .391 ± .025 \\
Meta-Llama-2-13B-Chat & .416 ± .014 & .474 ± .046 & .147 ± .030 \\
Meta-Llama-2-7B & .346 ± .040 & .450 ± .033 & .074 ± .010 \\
Meta-Llama-2-7B-Chat & .355 ± .022 & .485 ± .038 & .093 ± .028 \\
Meta-Llama-3-8B & .531 ± .015 & .485 ± .015 & .323 ± .027 \\
Meta-Llama-3-8B-Instruct & .524 ± .019 & .525 ± .024 & .536 ± .040 \\
Meta-Llama-3.1-70B-Instruct & .764 ± .020 & .574 ± .017 & .666 ± .088 \\
Meta-Llama-3.1-8B & .534 ± .017 & .502 ± .021 & .346 ± .022 \\
Meta-Llama-3.1-8B-Instruct & .554 ± .015 & .532 ± .016 & .543 ± .039 \\
Mistral-7B-Instruct-v0.1 & .418 ± .017 & .516 ± .045 & .171 ± .029 \\
Mistral-7B-Instruct-v0.2 & .474 ± .018 & .589 ± .051 & .258 ± .023 \\
Mistral-7B-Instruct-v0.3 & .493 ± .013 & .548 ± .046 & .320 ± .019 \\
Mistral-7B-v0.1 & .497 ± .021 & .495 ± .046 & .250 ± .026 \\
Mistral-7B-v0.3 & .498 ± .016 & .486 ± .042 & .238 ± .034 \\
Mistral-NeMo-Minitron-8B-Base & .526 ± .032 & .481 ± .034 & .384 ± .028 \\
Mistral-Nemo-Base-12.2B\_2407 & .585 ± .011 & .524 ± .026 & .422 ± .034 \\
Mistral-Nemo-Instruct-12.2B\_2407 & .579 ± .009 & .588 ± .023 & .549 ± .025 \\
Mixtral-8x7B-Instruct-v0.1 & .609 ± .013 & .620 ± .045 & .488 ± .027 \\
Mixtral-8x7B-v0.1 & .615 ± .017 & .505 ± .033 & .429 ± .029 \\
Occiglot-7b-eu5 & .394 ± .036 & .451 ± .040 & .154 ± .018 \\
Occiglot-7b-eu5-Instruct & .402 ± .027 & .465 ± .051 & .194 ± .030 \\
Pharia-1-LLM-7B-control & .291 ± .014 & .448 ± .037 & .010 ± .005 \\
Pharia-1-LLM-7B-control-aligned & .297 ± .024 & .447 ± .040 & .013 ± .009 \\
Phi-3-medium-14B-128k-Instruct & .516 ± .016 & .472 ± .032 & .432 ± .042 \\
Phi-3-medium-14B-4k-Instruct & .526 ± .022 & .470 ± .036 & .470 ± .048 \\
Phi-3-mini-3.8B-128k-Instruct & .387 ± .033 & .485 ± .036 & .193 ± .045 \\
Phi-3-mini-4k-Instruct & .376 ± .033 & .476 ± .038 & .165 ± .048 \\
Qwen2-7B & .578 ± .028 & .534 ± .021 & .584 ± .058 \\
Qwen2-7B-Instruct & .572 ± .022 & .547 ± .024 & .512 ± .064 \\
Vicuna-13b-v1.5 & .437 ± .014 & .509 ± .039 & .182 ± .025 \\
Vicuna-33b-v1.3 & .465 ± .019 & .529 ± .047 & .149 ± .033 \\
\end{longtable}

\begin{table}[htbp]
\caption{Differences between high and medium resource languages}
\label{tab:diff_med_high}
\centering
\begin{tabular}{llllll}
      & EU21 & EU21- & EU21- & EU21- & EU21- \\
Model & ARC  & GSM8k & HeSw  & MMLU  & TQA \\
\midrule
Aya-23-8B & .216 & .155 & .230 & .109 & .006 \\
Bloom-7b1 & .126 & .006 & .139 & .006 & -0.047 \\
Bloomz-7b1 & .128 & -0.000 & .141 & .056 & -0.025 \\
c4ai-command-r-35B-v01 & .201 & .124 & .205 & .112 & .030 \\
EuroLLM-1.7B & .052 & .004 & .050 & -0.001 & -0.031 \\
EuroLLM-1.7B-Instruct & .056 & .023 & .052 & .003 & -0.030 \\
Gemma-1.1-7b-Instruct & .094 & .047 & .107 & .061 & .003 \\
Gemma-2-27b-Instruct & .042 & .032 & .058 & .035 & .015 \\
Gemma-2-9b-Instruct & .084 & .061 & .094 & .055 & .023 \\
Gemma-7b & .076 & .047 & .084 & .043 & -0.008 \\
Meta-Llama-2-13B-Chat & .152 & .120 & .160 & .082 & .002 \\
Meta-Llama-2-7B & .146 & .042 & .144 & .071 & -0.021 \\
Meta-Llama-2-7B-Chat & .149 & .073 & .146 & .073 & .005 \\
Meta-Llama-3-8B & .107 & .076 & .116 & .069 & -0.023 \\
Meta-Llama-3-8B-Instruct & .116 & .116 & .123 & .079 & .012 \\
Meta-Llama-3.1-70B-Instruct & .055 & .029 & .076 & .030 & .009 \\
Meta-Llama-3.1-8B & .104 & .073 & .122 & .065 & -0.012 \\
Meta-Llama-3.1-8B-Instruct & .108 & .096 & .122 & .071 & .013 \\
Mistral-7B-Instruct-v0.1 & .172 & .133 & .156 & .097 & .057 \\
Mistral-7B-Instruct-v0.2 & .177 & .119 & .179 & .096 & .073 \\
Mistral-7B-Instruct-v0.3 & .177 & .126 & .171 & .096 & .044 \\
Mistral-7B-v0.1 & .169 & .106 & .165 & .101 & -0.008 \\
Mistral-7B-v0.3 & .170 & .095 & .164 & .097 & -0.004 \\
Mistral-NeMo-Minitron-8B-Base & .175 & .111 & .176 & .103 & .012 \\
Mistral-Nemo-Base-12.2B\_2407 & .108 & .087 & .138 & .070 & .001 \\
Mistral-Nemo-Instruct-12.2B\_2407 & .108 & .107 & .138 & .069 & -0.005 \\
Mixtral-8x7B-Instruct-v0.1 & .171 & .140 & .183 & .103 & .053 \\
Mixtral-8x7B-v0.1 & .165 & .128 & .172 & .103 & .007 \\
Occiglot-7b-eu5 & .197 & .100 & .219 & .112 & -0.028 \\
Occiglot-7b-eu5-Instruct & .202 & -0.000 & .224 & .114 & -0.016 \\
Pharia-1-LLM-7B-control & .276 & .045 & .288 & .124 & .000 \\
Pharia-1-LLM-7B-control-aligned & .281 & .104 & .294 & .136 & .027 \\
Phi-3-medium-14B-128k-Instruct & .259 & .320 & .256 & .188 & .064 \\
Phi-3-medium-14B-4k-Instruct & .262 & .317 & .260 & .193 & .072 \\
Phi-3-mini-3.8B-128k-Instruct & .248 & .390 & .228 & .195 & .059 \\
Phi-3-mini-4k-Instruct & .245 & .363 & .233 & .183 & .070 \\
Qwen2-7B & .182 & .156 & .181 & .099 & .016 \\
Qwen2-7B-Instruct & .184 & .082 & .188 & .098 & .038 \\
Vicuna-13b-v1.5 & .148 & .092 & .161 & .088 & .015 \\
Vicuna-33b-v1.3 & .182 & .117 & .174 & .101 & .024 \\
\end{tabular}
\end{table}

\subsection{Downstream results heatmaps}
\label{appendix:evaluation:downstream_heatmaps}

\begin{figure*}
    \centering
    \includegraphics[width=\textwidth]{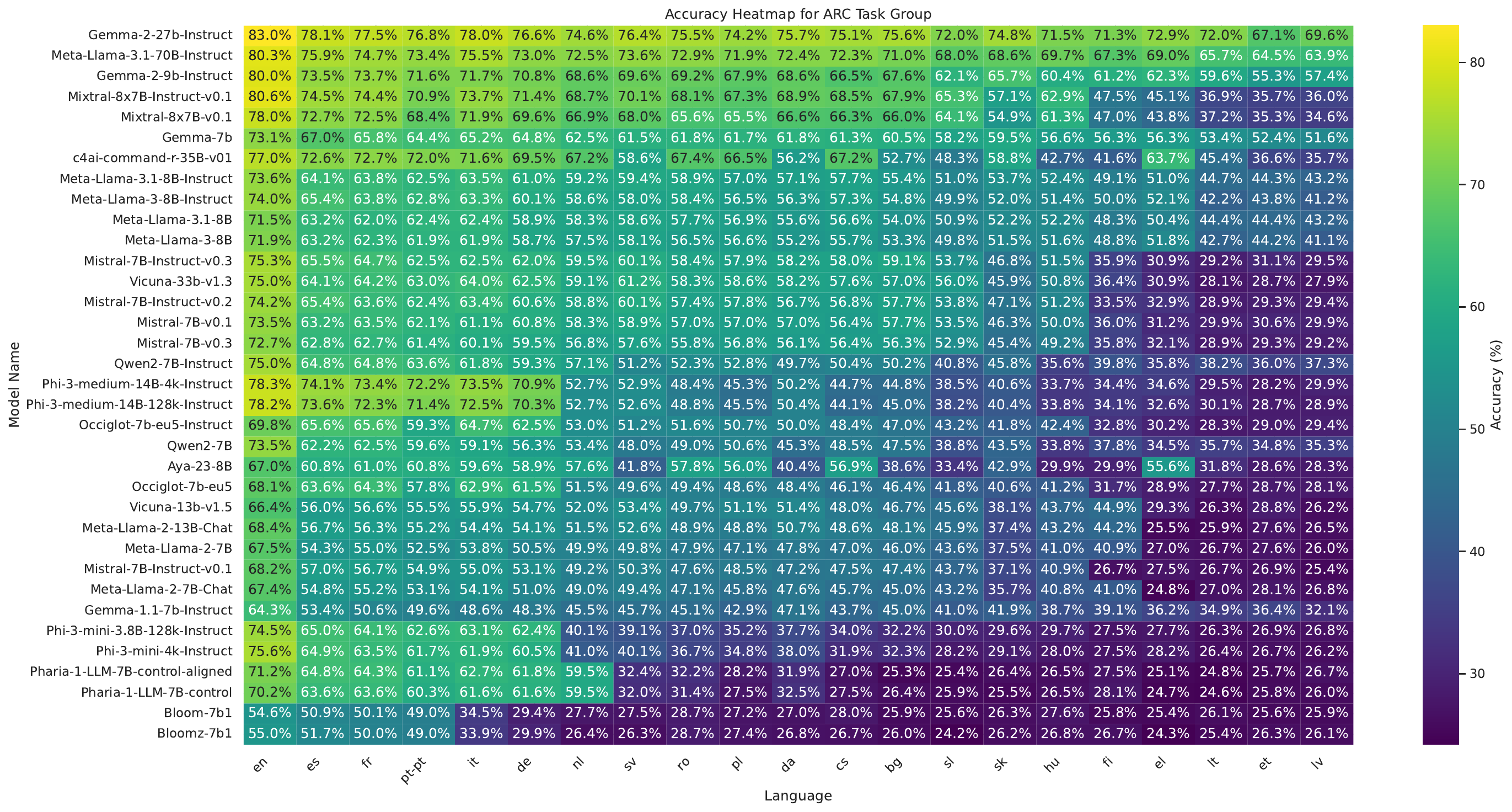}
    \caption{Task Accuracies per Language for EU21-ARC.}
    \label{fig:ds-taskacc-EU21-ARC}
\end{figure*}

\begin{figure*}
    \centering
    \includegraphics[width=\textwidth]{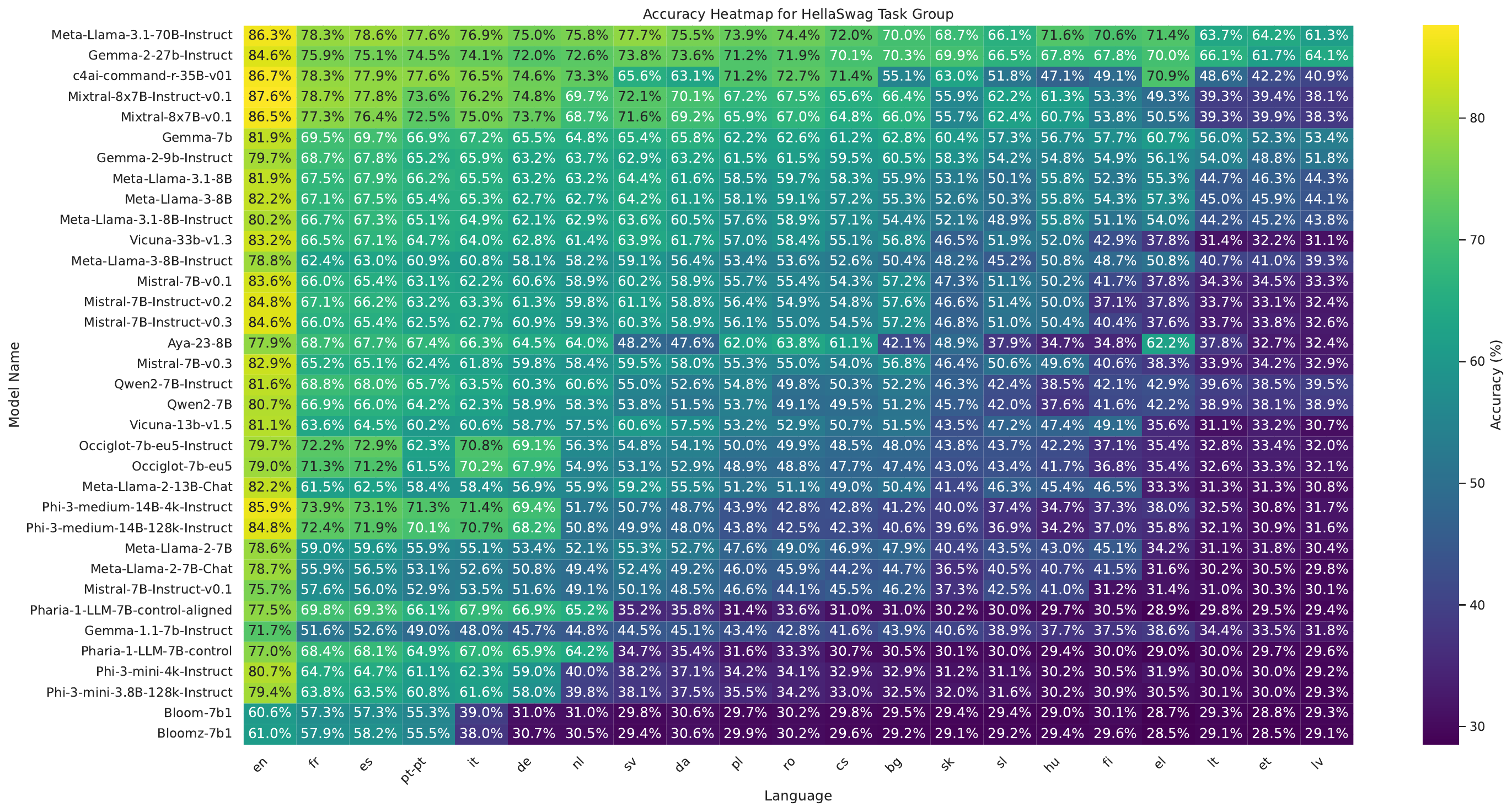}
    \caption{Task Accuracies per Language for EU21-HeSw.}
    \label{fig:ds-taskacc-EU21-HeSw}
\end{figure*}

\begin{figure*}
    \centering
    \includegraphics[width=\textwidth]{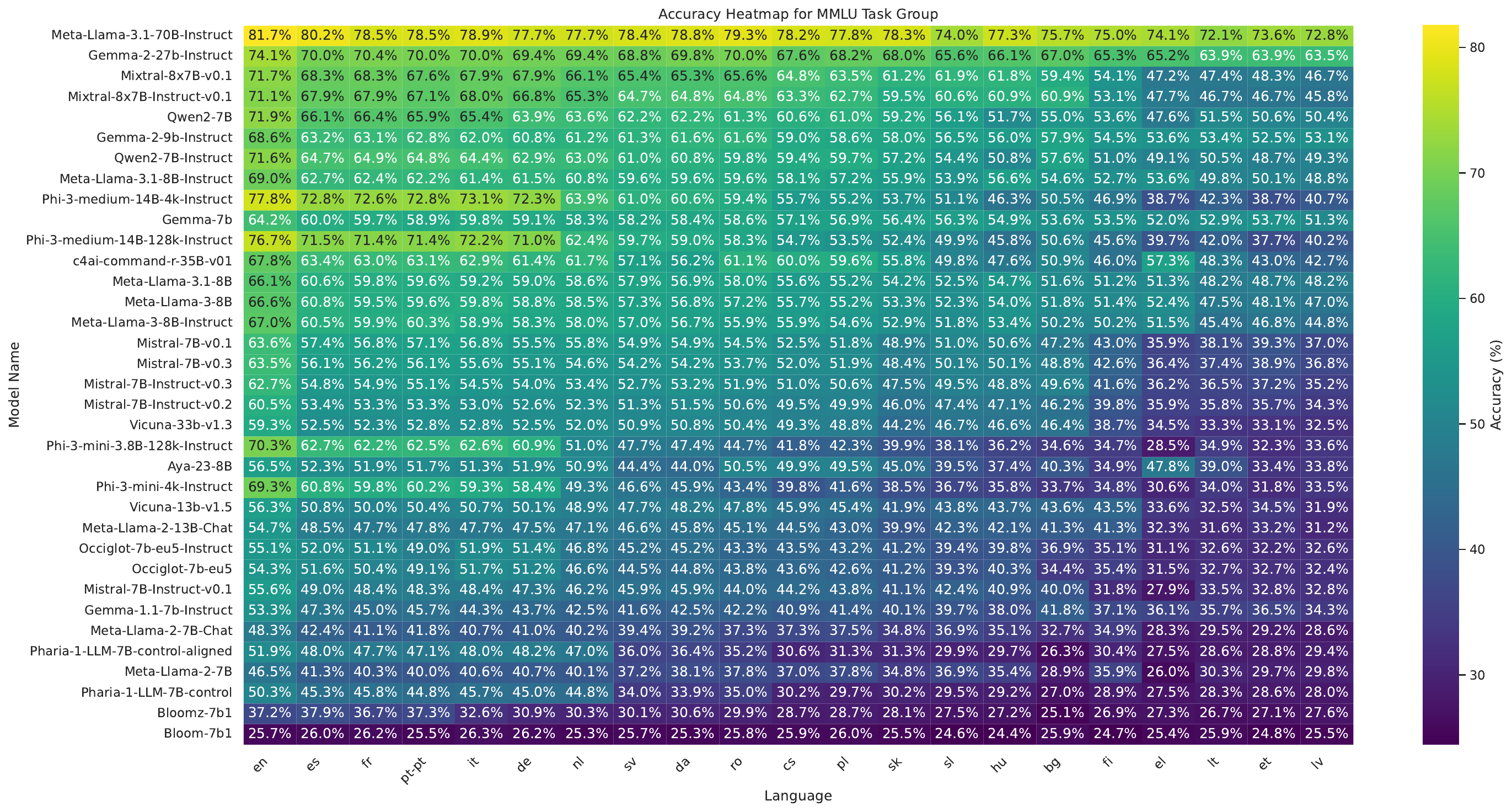}
    \caption{Task Accuracies per Language for EU21-MMLU.}
    \label{fig:ds-taskacc-EU21-MMLU}
\end{figure*}

\begin{figure*}
    \centering
    \includegraphics[width=\textwidth]{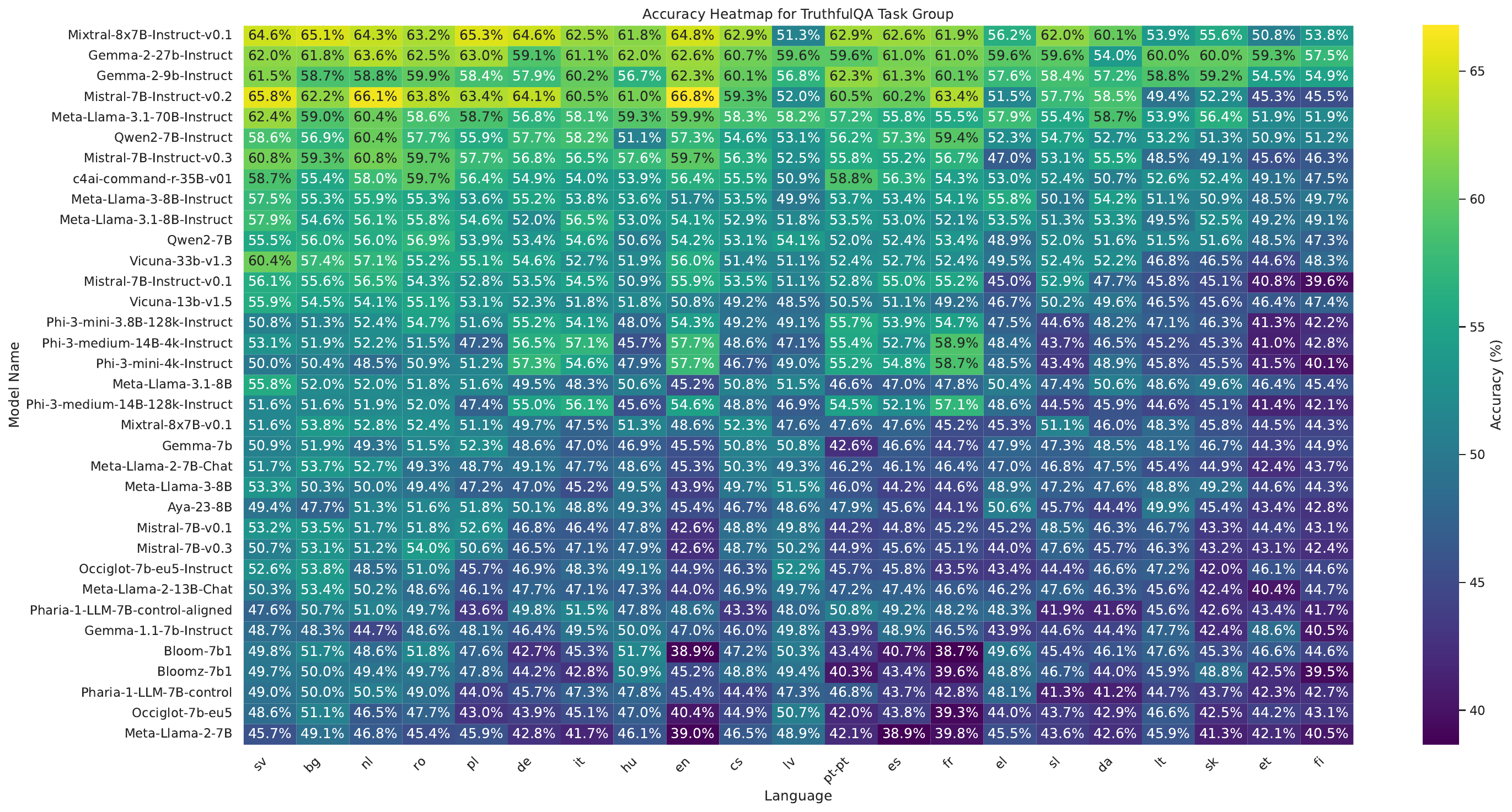}
    \caption{Task Accuracies per Language for EU21-TQA.}
    \label{fig:ds-taskacc-EU21-TQA}
\end{figure*}

\begin{figure*}
    \centering
    \includegraphics[width=\textwidth]{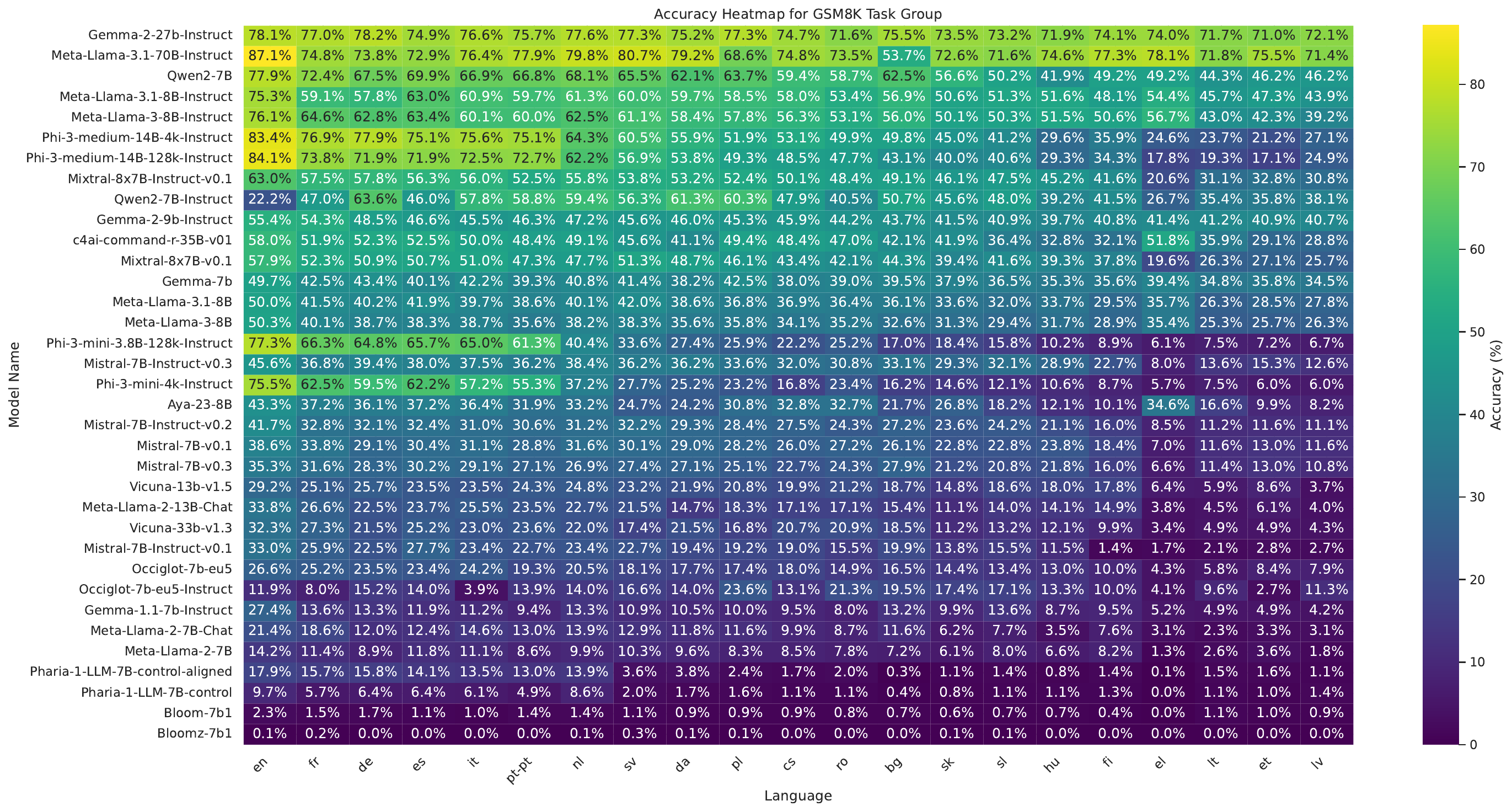}
    \caption{Task Accuracies per Language for EU21-GSM8k.}
    \label{fig:ds-taskacc-EU21-GSM8k}
\end{figure*}
\twocolumn

\clearpage
\section{Correlation with Human Preferences}\label{appendix:lmsys}
\subsection{Battle dataset overview}\label{appendix:lmsys:battles_overview}
The battle dataset\footnote{\url{https://storage.googleapis.com/arena_external_data/public/clean_battle_20240814_public.json} (as of 2024-10-15)} dataset we used for computing the Elo scores was published by LMSYS together with a Jupyter notebook\footnotemark for reproducibility purposes.
\footnotetext{\url{https://colab.research.google.com/drive/1KdwokPjirkTmpO_P1WByFNFiqxWQquwH} (as of 2024-10-15)}
\Cref{tab:battle_overview} shows the proportions of the different languages in the dataset, according to the classification contained in the dataset itself:
\begin{table}[h]
\centering
\begin{tabular}{lr}
\toprule
\textbf{Language} & \textbf{Number of Battles} \\
\midrule
English       & 1,035,743 \\
German        & 50,419 \\
Spanish       & 31,278 \\
French        & 28,819 \\
Portuguese    & 16,959 \\
Italian       & 14,412 \\
Czech         & 7,517 \\
Polish        & 6,868 \\
Dutch         & 2,748 \\
Danish        & 2,392 \\
Swedish       & 2,325 \\
Finnish       & 2,211 \\
Hungarian     & 2,066 \\
Bulgarian     & 1,737 \\
Romanian      & 1,273 \\
Greek         & 968 \\
Slovak        & 831 \\
Croatian      & 411 \\
Lithuanian    & 395 \\
Estonian      & 256 \\
Slovenian     & 232 \\
Latvian       & 199 \\
Irish         & 134 \\
Maltese       & 91 \\
\bottomrule
\end{tabular}
\caption{Battle Overview}
\label{tab:battle_overview}
\end{table}

\subsection{Computation of model rank}
\label{appendix:lmsys:model_rank}
The model rank for a model $m_A$ is computed by the FastChat codebase as 1 plus the number of models $m_B$ such that the 97.5\% quantile of the Elo score of $m_A$ is lower than the 2.5\% quantile of the Elo score of $m_B$.

For each such $m_B$, there is a 97.5\% probability that the ``true'' Elo score of $m_B$ is greater than its 2.5\% quantile, and this quantile is greater than the 97.5\% quantile of the Elo score of $m_A$, which in turn has a 97.5\% probability of being greater than the ``true'' Elo score of $m_A$.
Thus, there is a probability greater than $97.5\% * 97.5\% = 95.0625\%$ that the ``true'' Elo score of $m_B$ is greater than the ``true'' Elo score of $m_A$.

\subsection{Elo Score Tables}
\label{appendix:lmsys:elo_scores}
Tables~\ref{tab:lmsys_pearson_arc_few_shot}, \ref{tab:lmsys_pearson_gsm8k_few_shot}, \ref{tab:lmsys_pearson_hellaswag_few_shot}, \ref{tab:lmsys_pearson_mmlu_few_shot}, and \ref{tab:lmsys_pearson_truthfulqa_few_shot} each contain the per-task language-wise correlations computed from the Elo scores along with p-values, confidence intervals, the number of models for which Elo scores were available, and the number of model ranks distinguishable with about 95\% confidence.
Tables~\ref{tab:lmsys_pearson_alltasks_few_shot} and \ref{tab:lmsys_pearson_alltasks_few_shot} contains the same information for the average of the accuracy across all tasks, and~\ref{tab:lmsys_pearson_reasoning_few_shot} for the average of the accuracy across all tasks except EU21-TQA.

\onecolumn
\begin{longtable}[htbp]{lllrlrr}
\caption{Correlation Coefficients for Task \acrshort{xARC} between Few-Shot Accuracy and LMSYS ELO Score, incl. 95\% confidence intervals for Pearson. Numbers behind @ are p values, $+$ indicates value below 0.05. NUR: num. unique ranks with ~95\% confidence, $+$ indicates value below 50\%.} \label{tab:lmsys_pearson_arc_few_shot} \\
\toprule
Task & Lang. & Pearson & Pearson CI & Spearman & N & NUR \\
\midrule
\endfirsthead
\caption[]{Correlation Coefficients for Task \acrshort{xARC} between Few-Shot Accuracy and LMSYS ELO Score, incl. 95\% confidence intervals for Pearson. Numbers behind @ are p values, $+$ indicates value below 0.05. NUR: num. unique ranks with ~95\% confidence, $+$ indicates value below 50\%.} \\
\toprule
Task & Lang. & Pearson & Pearson CI & Spearman & N & NUR \\
\midrule
\endhead
\midrule
\multicolumn{7}{r}{Continued on next page} \\
\midrule
\endfoot
\bottomrule
\endlastfoot
\acrshort{xARC} & CS & $.600@.00541\,(+)$ & $(.249, 1)$ & $.499@.02070\,(+)$ & 17 & $9\,(+)$ \\
\acrshort{xARC} & EN & $.722@.00054\,(+)$ & $(.439, 1)$ & $.691@.00106\,(+)$ & 17 & $16\,(+)$ \\
\acrshort{xARC} & FR & $.772@.00014\,(+)$ & $(.526, 1)$ & $.770@.00015\,(+)$ & 17 & $14\,(+)$ \\
\acrshort{xARC} & DE & $.815@.00003\,(+)$ & $(.606, 1)$ & $.779@.00011\,(+)$ & 17 & $14\,(+)$ \\
\acrshort{xARC} & IT & $.700@.00088\,(+)$ & $(.403, 1)$ & $.696@.00096\,(+)$ & 17 & $15\,(+)$ \\
\acrshort{xARC} & PL & $.868@.00000\,(+)$ & $(.709, 1)$ & $.836@.00001\,(+)$ & 17 & $13\,(+)$ \\
\acrshort{xARC} & PT & $.779@.00012\,(+)$ & $(.539, 1)$ & $.784@.00010\,(+)$ & 17 & $14\,(+)$ \\
\acrshort{xARC} & ES & $.743@.00032\,(+)$ & $(.475, 1)$ & $.760@.00020\,(+)$ & 17 & $14\,(+)$ \\
\end{longtable}

\begin{longtable}[htbp]{lllrlrr}
\caption{Correlation Coefficients for Task \acrshort{xGSM8K} between Few-Shot Accuracy and LMSYS ELO Score, incl. 95\% confidence intervals for Pearson. Numbers behind @ are p values, $+$ indicates value below 0.05. NUR: num. unique ranks with ~95\% confidence, $+$ indicates value below 50\%.} \label{tab:lmsys_pearson_gsm8k_few_shot} \\
\toprule
Task & Lang. & Pearson & Pearson CI & Spearman & N & NUR \\
\midrule
\endfirsthead
\caption[]{Correlation Coefficients for Task \acrshort{xGSM8K} between Few-Shot Accuracy and LMSYS ELO Score, incl. 95\% confidence intervals for Pearson. Numbers behind @ are p values, $+$ indicates value below 0.05. NUR: num. unique ranks with ~95\% confidence, $+$ indicates value below 50\%.} \\
\toprule
Task & Lang. & Pearson & Pearson CI & Spearman & N & NUR \\
\midrule
\endhead
\midrule
\multicolumn{7}{r}{Continued on next page} \\
\midrule
\endfoot
\bottomrule
\endlastfoot
\acrshort{xGSM8K} & CS & $.465@.03006\,(+)$ & $(.064, 1)$ & $.302@.11966\,(-)$ & 17 & $9\,(+)$ \\
\acrshort{xGSM8K} & EN & $.671@.00159\,(+)$ & $(.357, 1)$ & $.603@.00520\,(+)$ & 17 & $16\,(+)$ \\
\acrshort{xGSM8K} & FR & $.723@.00052\,(+)$ & $(.441, 1)$ & $.694@.00101\,(+)$ & 17 & $14\,(+)$ \\
\acrshort{xGSM8K} & DE & $.717@.00060\,(+)$ & $(.431, 1)$ & $.613@.00441\,(+)$ & 17 & $14\,(+)$ \\
\acrshort{xGSM8K} & IT & $.616@.00422\,(+)$ & $(.272, 1)$ & $.586@.00674\,(+)$ & 17 & $15\,(+)$ \\
\acrshort{xGSM8K} & PL & $.785@.00010\,(+)$ & $(.549, 1)$ & $.713@.00065\,(+)$ & 17 & $13\,(+)$ \\
\acrshort{xGSM8K} & PT & $.786@.00009\,(+)$ & $(.553, 1)$ & $.743@.00032\,(+)$ & 17 & $14\,(+)$ \\
\acrshort{xGSM8K} & ES & $.685@.00119\,(+)$ & $(.380, 1)$ & $.662@.00190\,(+)$ & 17 & $14\,(+)$ \\
\end{longtable}

\begin{longtable}[htbp]{lllrlrr}
\caption{Correlation Coefficients for Task \acrshort{xHella} between Few-Shot Accuracy and LMSYS ELO Score, incl. 95\% confidence intervals for Pearson. Numbers behind @ are p values, $+$ indicates value below 0.05. NUR: num. unique ranks with ~95\% confidence, $+$ indicates value below 50\%.} \label{tab:lmsys_pearson_hellaswag_few_shot} \\
\toprule
Task & Lang. & Pearson & Pearson CI & Spearman & N & NUR \\
\midrule
\endfirsthead
\caption[]{Correlation Coefficients for Task \acrshort{xHella} between Few-Shot Accuracy and LMSYS ELO Score, incl. 95\% confidence intervals for Pearson. Numbers behind @ are p values, $+$ indicates value below 0.05. NUR: num. unique ranks with ~95\% confidence, $+$ indicates value below 50\%.} \\
\toprule
Task & Lang. & Pearson & Pearson CI & Spearman & N & NUR \\
\midrule
\endhead
\midrule
\multicolumn{7}{r}{Continued on next page} \\
\midrule
\endfoot
\bottomrule
\endlastfoot
\acrshort{xHella} & CS & $.581@.00720\,(+)$ & $(.221, 1)$ & $.479@.02574\,(+)$ & 17 & $9\,(+)$ \\
\acrshort{xHella} & EN & $.430@.04248\,(+)$ & $(.020, 1)$ & $.436@.03999\,(+)$ & 17 & $16\,(+)$ \\
\acrshort{xHella} & FR & $.722@.00053\,(+)$ & $(.440, 1)$ & $.679@.00136\,(+)$ & 17 & $14\,(+)$ \\
\acrshort{xHella} & DE & $.719@.00057\,(+)$ & $(.435, 1)$ & $.762@.00019\,(+)$ & 17 & $14\,(+)$ \\
\acrshort{xHella} & IT & $.714@.00064\,(+)$ & $(.426, 1)$ & $.669@.00165\,(+)$ & 17 & $15\,(+)$ \\
\acrshort{xHella} & PL & $.851@.00001\,(+)$ & $(.676, 1)$ & $.855@.00001\,(+)$ & 17 & $13\,(+)$ \\
\acrshort{xHella} & PT & $.729@.00045\,(+)$ & $(.452, 1)$ & $.784@.00010\,(+)$ & 17 & $14\,(+)$ \\
\acrshort{xHella} & ES & $.687@.00117\,(+)$ & $(.381, 1)$ & $.752@.00025\,(+)$ & 17 & $14\,(+)$ \\
\end{longtable}

\begin{longtable}[htbp]{lllrlrr}
\caption{Correlation Coefficients for Task \acrshort{xMMLU} between Few-Shot Accuracy and LMSYS ELO Score, incl. 95\% confidence intervals for Pearson. Numbers behind @ are p values, $+$ indicates value below 0.05. NUR: num. unique ranks with ~95\% confidence, $+$ indicates value below 50\%.} \label{tab:lmsys_pearson_mmlu_few_shot} \\
\toprule
Task & Lang. & Pearson & Pearson CI & Spearman & N & NUR \\
\midrule
\endfirsthead
\caption[]{Correlation Coefficients for Task \acrshort{xMMLU} between Few-Shot Accuracy and LMSYS ELO Score, incl. 95\% confidence intervals for Pearson. Numbers behind @ are p values, $+$ indicates value below 0.05. NUR: num. unique ranks with ~95\% confidence, $+$ indicates value below 50\%.} \\
\toprule
Task & Lang. & Pearson & Pearson CI & Spearman & N & NUR \\
\midrule
\endhead
\midrule
\multicolumn{7}{r}{Continued on next page} \\
\midrule
\endfoot
\bottomrule
\endlastfoot
\acrshort{xMMLU} & CS & $.564@.00917\,(+)$ & $(.197, 1)$ & $.416@.04851\,(-)$ & 17 & $9\,(+)$ \\
\acrshort{xMMLU} & EN & $.737@.00037\,(+)$ & $(.465, 1)$ & $.610@.00464\,(+)$ & 17 & $16\,(+)$ \\
\acrshort{xMMLU} & FR & $.814@.00004\,(+)$ & $(.603, 1)$ & $.804@.00005\,(+)$ & 17 & $14\,(+)$ \\
\acrshort{xMMLU} & DE & $.792@.00008\,(+)$ & $(.563, 1)$ & $.782@.00010\,(+)$ & 17 & $14\,(+)$ \\
\acrshort{xMMLU} & IT & $.734@.00040\,(+)$ & $(.460, 1)$ & $.610@.00464\,(+)$ & 17 & $15\,(+)$ \\
\acrshort{xMMLU} & PL & $.885@.00000\,(+)$ & $(.744, 1)$ & $.863@.00000\,(+)$ & 17 & $13\,(+)$ \\
\acrshort{xMMLU} & PT & $.804@.00005\,(+)$ & $(.585, 1)$ & $.760@.00020\,(+)$ & 17 & $14\,(+)$ \\
\acrshort{xMMLU} & ES & $.768@.00016\,(+)$ & $(.520, 1)$ & $.770@.00015\,(+)$ & 17 & $14\,(+)$ \\
\end{longtable}

\begin{longtable}[htbp]{lllrlrr}
\caption{Correlation Coefficients for Task \acrshort{xTQA} between Few-Shot Accuracy and LMSYS ELO Score, incl. 95\% confidence intervals for Pearson. Numbers behind @ are p values, $+$ indicates value below 0.05. NUR: num. unique ranks with ~95\% confidence, $+$ indicates value below 50\%.} \label{tab:lmsys_pearson_truthfulqa_few_shot} \\
\toprule
Task & Lang. & Pearson & Pearson CI & Spearman & N & NUR \\
\midrule
\endfirsthead
\caption[]{Correlation Coefficients for Task \acrshort{xTQA} between Few-Shot Accuracy and LMSYS ELO Score, incl. 95\% confidence intervals for Pearson. Numbers behind @ are p values, $+$ indicates value below 0.05. NUR: num. unique ranks with ~95\% confidence, $+$ indicates value below 50\%.} \\
\toprule
Task & Lang. & Pearson & Pearson CI & Spearman & N & NUR \\
\midrule
\endhead
\midrule
\multicolumn{7}{r}{Continued on next page} \\
\midrule
\endfoot
\bottomrule
\endlastfoot
\acrshort{xTQA} & CS & $.433@.04129\,(+)$ & $(.024, 1)$ & $.401@.05534\,(-)$ & 17 & $9\,(+)$ \\
\acrshort{xTQA} & EN & $.426@.04397\,(+)$ & $(.016, 1)$ & $.510@.01828\,(+)$ & 17 & $16\,(+)$ \\
\acrshort{xTQA} & FR & $.360@.07793\,(-)$ & $(-0.063, 1)$ & $.309@.11389\,(-)$ & 17 & $14\,(+)$ \\
\acrshort{xTQA} & DE & $.343@.08872\,(-)$ & $(-0.082, 1)$ & $.417@.04808\,(-)$ & 17 & $14\,(+)$ \\
\acrshort{xTQA} & IT & $.380@.06619\,(-)$ & $(-0.039, 1)$ & $.277@.14092\,(-)$ & 17 & $15\,(+)$ \\
\acrshort{xTQA} & PL & $.603@.00524\,(+)$ & $(.252, 1)$ & $.706@.00077\,(+)$ & 17 & $13\,(+)$ \\
\acrshort{xTQA} & PT & $.548@.01138\,(+)$ & $(.174, 1)$ & $.569@.00861\,(+)$ & 17 & $14\,(+)$ \\
\acrshort{xTQA} & ES & $.447@.03615\,(+)$ & $(.041, 1)$ & $.473@.02757\,(+)$ & 17 & $14\,(+)$ \\
\end{longtable}

\begin{longtable}[htbp]{lllrlrr}
\caption{Correlation Coefficients for Task Avg. between Few-Shot Accuracy and LMSYS ELO Score, incl. 95\% confidence intervals for Pearson. Numbers behind @ are p values, $+$ indicates value below 0.05. NUR: num. unique ranks with ~95\% confidence, $+$ indicates value below 50\%.} \label{tab:lmsys_pearson_alltasks_few_shot} \\
\toprule
Task & Lang. & Pearson & Pearson CI & Spearman & N & NUR \\
\midrule
\endfirsthead
\caption[]{Correlation Coefficients for Task Avg. between Few-Shot Accuracy and LMSYS ELO Score, incl. 95\% confidence intervals for Pearson. Numbers behind @ are p values, $+$ indicates value below 0.05. NUR: num. unique ranks with ~95\% confidence, $+$ indicates value below 50\%.} \\
\toprule
Task & Lang. & Pearson & Pearson CI & Spearman & N & NUR \\
\midrule
\endhead
\midrule
\multicolumn{7}{r}{Continued on next page} \\
\midrule
\endfoot
\bottomrule
\endlastfoot
Avg. & CS & $.570@.00847\,(+)$ & $(.205, 1)$ & $.430@.04230\,(+)$ & 17 & $9\,(+)$ \\
Avg. & EN & $.716@.00062\,(+)$ & $(.429, 1)$ & $.618@.00412\,(+)$ & 17 & $16\,(+)$ \\
Avg. & FR & $.770@.00015\,(+)$ & $(.523, 1)$ & $.745@.00030\,(+)$ & 17 & $14\,(+)$ \\
Avg. & DE & $.775@.00013\,(+)$ & $(.531, 1)$ & $.740@.00034\,(+)$ & 17 & $14\,(+)$ \\
Avg. & IT & $.694@.00101\,(+)$ & $(.393, 1)$ & $.674@.00150\,(+)$ & 17 & $15\,(+)$ \\
Avg. & PL & $.895@.00000\,(+)$ & $(.765, 1)$ & $.858@.00001\,(+)$ & 17 & $13\,(+)$ \\
Avg. & PT & $.811@.00004\,(+)$ & $(.598, 1)$ & $.806@.00005\,(+)$ & 17 & $14\,(+)$ \\
Avg. & ES & $.751@.00025\,(+)$ & $(.490, 1)$ & $.784@.00010\,(+)$ & 17 & $14\,(+)$ \\
\end{longtable}

\begin{longtable}{lllrlrr}
\caption{Correlation Coefficients for Task Reasoning Avg. between Few-Shot Accuracy and LMSYS ELO Score, incl. 95\% confidence intervals for Pearson. Numbers behind @ are p values, $+$ indicates value below 0.05. NUR: num. unique ranks with ~95\% confidence, $+$ indicates value below 50\%.} \label{tab:lmsys_pearson_reasoning_few_shot} \\
\toprule
Task & Lang. & Pearson & Pearson CI & Spearman & N & NUR \\
\midrule
\endfirsthead
\caption[]{Correlation Coefficients for Task Reasoning Avg. between Few-Shot Accuracy and LMSYS ELO Score, incl. 95\% confidence intervals for Pearson. Numbers behind @ are p values, $+$ indicates value below 0.05. NUR: num. unique ranks with ~95\% confidence, $+$ indicates value below 50\%.} \\
\toprule
Task & Lang. & Pearson & Pearson CI & Spearman & N & NUR \\
\midrule
\endhead
\midrule
\multicolumn{7}{r}{Continued on next page} \\
\midrule
\endfoot
\bottomrule
\endlastfoot
Avg - TQA. & CS & $.572@.00818\,(+)$ & $(.208, 1)$ & $.397@.05715\,(-)$ & 17 & $9\,(+)$ \\
Avg - TQA. & EN & $.722@.00053\,(+)$ & $(.440, 1)$ & $.623@.00380\,(+)$ & 17 & $16\,(+)$ \\
Avg - TQA. & FR & $.793@.00007\,(+)$ & $(.564, 1)$ & $.789@.00008\,(+)$ & 17 & $14\,(+)$ \\
Avg - TQA. & DE & $.792@.00008\,(+)$ & $(.562, 1)$ & $.775@.00013\,(+)$ & 17 & $14\,(+)$ \\
Avg - TQA. & IT & $.707@.00076\,(+)$ & $(.415, 1)$ & $.664@.00182\,(+)$ & 17 & $15\,(+)$ \\
Avg - TQA. & PL & $.901@.00000\,(+)$ & $(.777, 1)$ & $.855@.00001\,(+)$ & 17 & $13\,(+)$ \\
Avg - TQA. & PT & $.819@.00003\,(+)$ & $(.613, 1)$ & $.777@.00012\,(+)$ & 17 & $14\,(+)$ \\
Avg - TQA. & ES & $.756@.00022\,(+)$ & $(.498, 1)$ & $.767@.00016\,(+)$ & 17 & $14\,(+)$ \\
\end{longtable}
\twocolumn

\clearpage
\twocolumn
\section{Influence of translation service}\label{appendix:okapi}

In this Section, we compare the OKAPI dataset with our dataset. We use the OKAPI datasets from the LM Eval Harness, which has been extended by seven languages compared to the original publication \cite{lai-etal-2023-okapi}. 

\subsection{Dataset overview}\label{appendix:okapi:dataset_overview}

Table \ref{tab:sample_nr} compares the number of translated samples in the 11 languages that overlap with both datasets. Our dataset aims to cover all samples in different languages. For the validation splits, we cover all samples across all languages except for Hellaswag (99.93\% translated). For the MMLU and ARC benchmark, the validation split is used for in-context examples in the LM Eval Harness. For the test split, we also covered all samples. The OKAPI translations cover, on average, 92.23\% of the samples for all languages regardless of the split, excluding the MMLU validation split. For the MMLU validation split, 276 samples were translated on average.

\begin{table*}[htbp]
    \centering
    \caption{Average sample number for each task of eleven translations.}
    \begin{tabular}{llrr lr l}
        \hline
         & \textbf{Split} & \textbf{Original} & \multicolumn{2}{c}{\textbf{EU21}} & \multicolumn{2}{c}{\textbf{OKAPI}}  \\ 
        \hline
        MMLU        & Val. & 1531   &  1531  &  &  276.18 &$\pm 8$\\
                    & Test & 14042  &  14042  & & 13198.09 &$\pm 101$\\
        \hline
        Hellaswag   & Val  & 10042  & 10035.18 &$\pm 19$ & 9276.63 &$\pm 112$\\
        \hline
        ARC         & Val. & 299    & 299  & & 272.66 &$\pm 0.5$ \\
                    & Test & 1172   & 1172 & & 1071 &$\pm 2$\\
        \hline
        TruthfulQA  & Val. & 817    & 817   &   & 782.09 &$\pm 6$ \\
        \hline
    \end{tabular}
    \label{tab:sample_nr}
\end{table*}

\subsection{Accuracy Plots}\label{appendix:okapi:accuracy_plots}

\begin{figure*}[htbp]
    \centering
    \includegraphics[width=\textwidth]{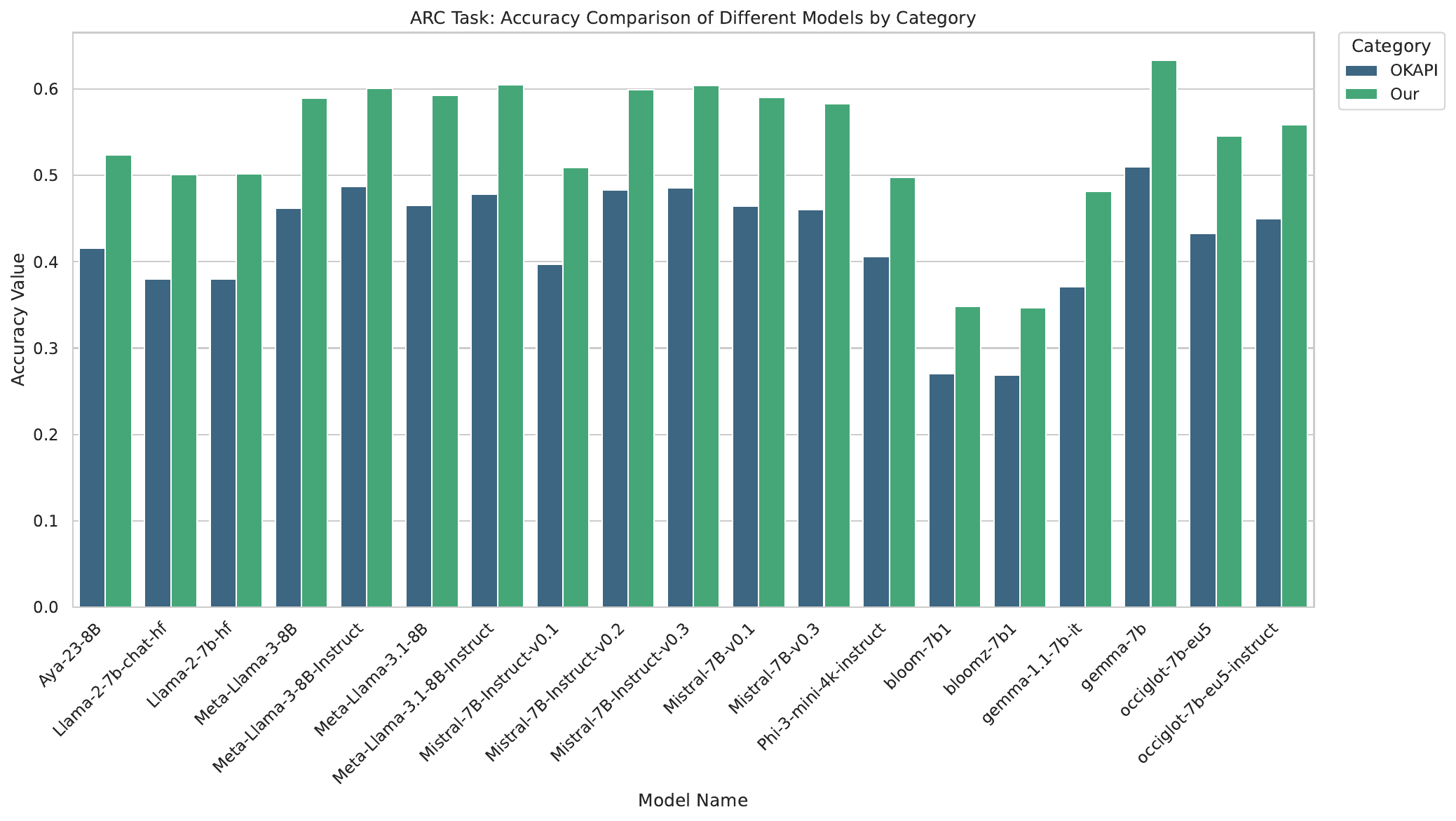}
    \caption{Accuracy comparison of different models on our and OKAPIs translated ARC dataset.}
    \label{fig:okapi-arc}
\end{figure*}

\begin{figure*}[htbp]
    \centering
    \includegraphics[width=\textwidth]{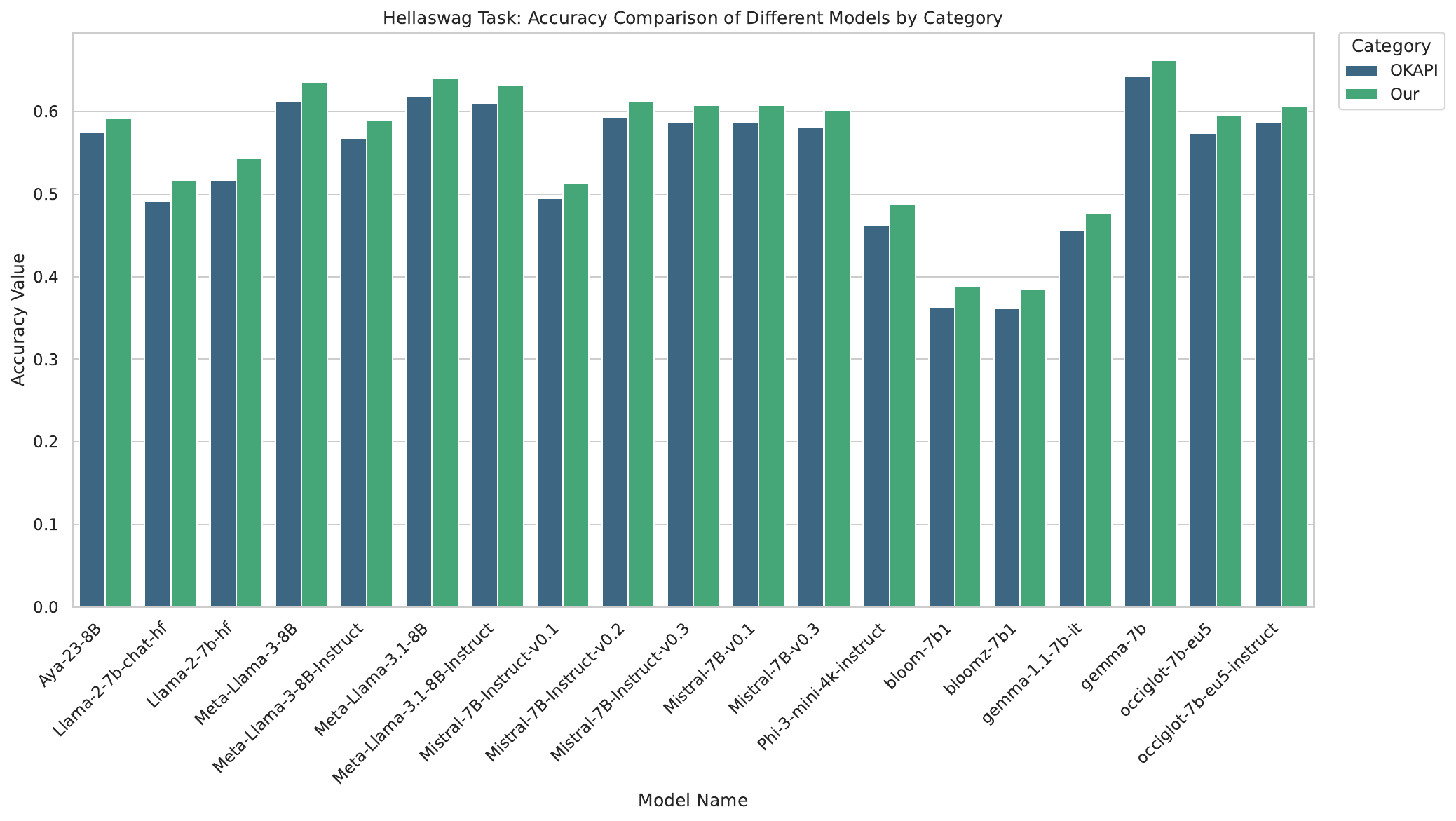}
    \caption{Accuracy comparison of different models on our and OKAPIs translated HELLASWAG dataset.}
    \label{fig:okapi-hellaswag}
\end{figure*}

\begin{figure*}[htbp]
    \centering
    \includegraphics[width=\textwidth]{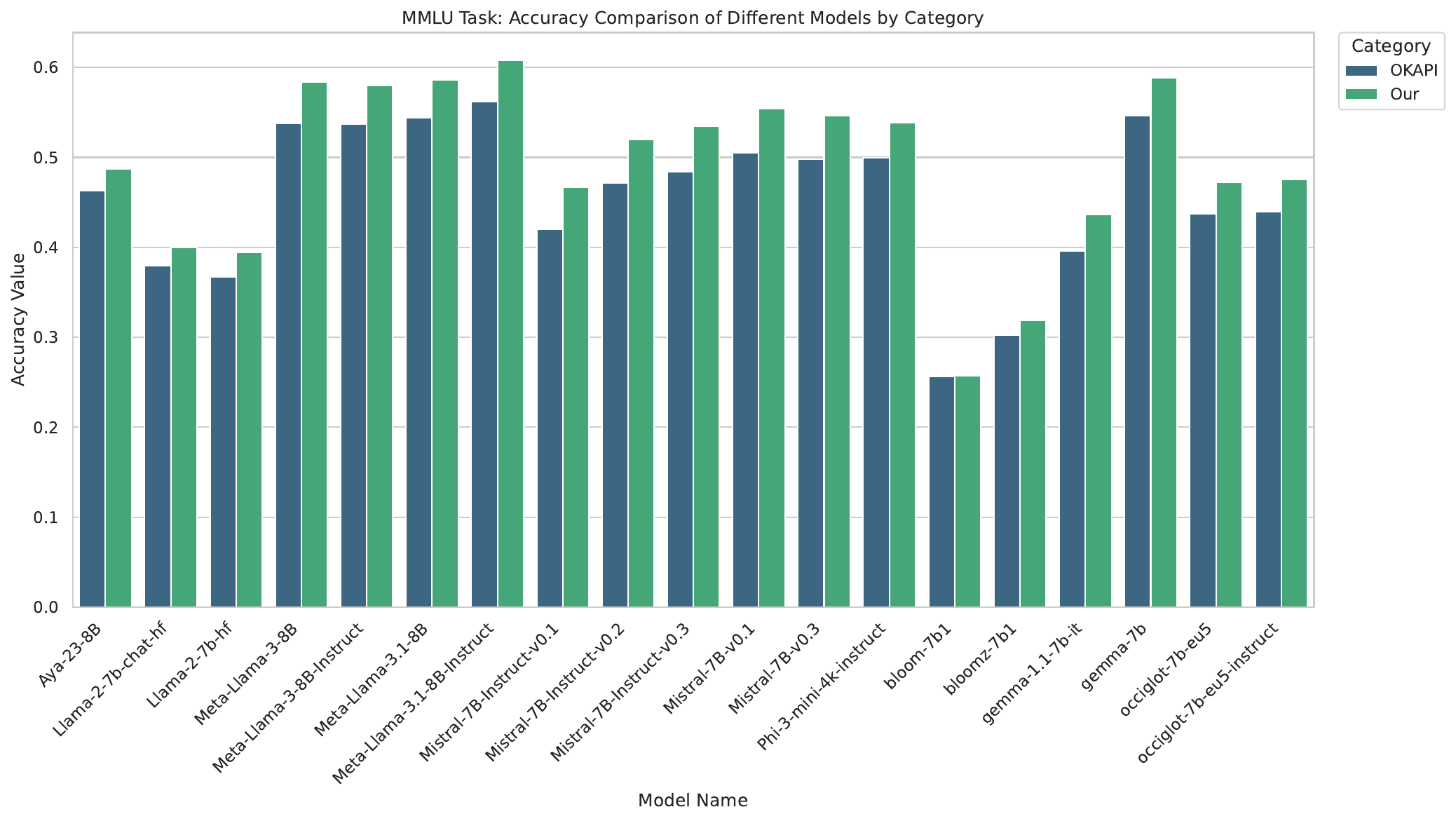}
    \caption{Accuracy comparison of different models on our and OKAPIs translated MMLU dataset.}
    \label{fig:okapi-mmlu}
\end{figure*}

In Figures \ref{fig:okapi-arc} , \ref{fig:okapi-hellaswag}, and \ref{fig:okapi-mmlu}, we compare the accuracy of 19 LLMs on the EU21-ARC, EU21-Hellaswag, and EU21-MMLU with the OKAPI benchmarks, respectively. All models show a higher accuracy on our benchmarks regardless of the benchmark, with the only exception being the results of bloom-7b1 for the MMLU task (cf. Figure \ref{fig:okapi-mmlu}).

\subsection{Correlations}\label{appendix:okapi:correlations}

We have calculated the correlation using the Pearson correlation coefficient for a more detailed analysis of the results from Figures \ref{fig:okapi-arc} , \ref{fig:okapi-hellaswag}, and \ref{fig:okapi-mmlu}. In Table \ref{tab:model_performance_sorted}, \ref{tab:language_correlation}, and \ref{tab:task_correlation} we show the model performance correlation matrix clustered by model name, language, and task, respectively.

\begin{table}[htbp]
\caption{Model Performance Correlation Matrix \\
(PC: Pearson Correlation Coefficient)
}
\centering
\begin{tabular}{lc}
\hline
\textbf{Model Name} & \textbf{PC} \\
\hline
Aya-23-8B & 0.571 \\
Meta-Llama-3-8B-Instruct & 0.562 \\
Mistral-7B-Instruct-v0.3 & 0.562 \\
Meta-Llama-3-8B & 0.556 \\
gemma-7b & 0.555 \\
Llama-2-7b-chat-hf & 0.553 \\
Meta-Llama-3.1-8B & 0.552 \\
gemma-1.1-7b-it & 0.552 \\
Meta-Llama-3.1-8B-Instruct & 0.549 \\
occiglot-7b-eu5 & 0.548 \\
occiglot-7b-eu5-instruct & 0.546 \\
Mistral-7B-v0.1 & 0.542 \\
Mistral-7B-Instruct-v0.2 & 0.539 \\
Mistral-7B-v0.3 & 0.536 \\
Phi-3-mini-4k-instruct & 0.534 \\
Mistral-7B-Instruct-v0.1 & 0.528 \\
Llama-2-7b-hf & 0.520 \\
bloomz-7b1 & 0.439 \\
bloom-7b1 & 0.401 \\
\hline
\end{tabular}
\label{tab:model_performance_sorted}
\end{table}

\begin{table}[htbp]
\caption{Language Correlation Coefficients \\
(PC: Pearson Correlation Coefficient)}
\centering
\begin{tabular}{lc}
\hline
\textbf{Language} & \textbf{PC} \\
\hline
es & 0.589991 \\
fr & 0.586389 \\
it & 0.567793 \\
sv & 0.564892 \\
da & 0.557272 \\
nl & 0.555682 \\
ro & 0.546222 \\
de & 0.545930 \\
sk & 0.473632 \\
hu & 0.444939 \\
\hline
\end{tabular}
\label{tab:language_correlation}
\end{table}

\begin{table}[htbp]
\caption{Task Correlation Coefficients \\
(PC: Pearson Correlation Coefficient)}
\centering
\begin{tabular}{lc}
\hline
\textbf{Task} & \textbf{PC} \\
\hline
ARC & 0.610 \\
HellaSwag & 0.543 \\
MMLU & 0.538 \\
\hline
\end{tabular}
\label{tab:task_correlation}
\end{table}

\subsection{Quality assurance}\label{appendix:okapi:quality_assurance}

In addition to the performance and correlation of the performance of the models, we also analyzed the translation quality using the COMET metric. Here, we performed the translations from our dataset and the OKAPI dataset for all available comparable languages. The metric we used was the COMET-KIWI metric, which is calculated using the “Unbabel/wmt22-cometkiwi-da”  model\footnote{https://huggingface.co/Unbabel/wmt22-cometkiwi-da}. This model is reference-free and built on top of InfoXLM \cite{rei-etal-2023-scaling}. 
For Hellaswag and MMLU, the results of our translations are slightly higher than those of OKAPI with no evidence of significance.

\begin{table*}[htbp]
    \centering
    \label{tab:comet-kiwi_score}
     \caption{Comparison of EU21 and Okapi scores for ARC and HellaSwag benchmarks (Higher values in bold)}
    \begin{tabular}{lcccccc}
        \hline
         & \multicolumn{2}{c}{\textbf{ARC}} & \multicolumn{2}{c}{\textbf{Hellaswag}} & \multicolumn{2}{c}{\textbf{MMLU}} \\
{\textbf{Language}}  & \textbf{EU21} & \textbf{OKAPI} & \textbf{EU21} & \textbf{OKAPI} & \textbf{EU21} & \textbf{OKAPI} \\ 
        \hline
        German    & 0.8289 & 0.8289 & 0.7857 & \textbf{0.7971} & \textbf{0.8110} & 0.8098\\
        Spanish   & \textbf{0.8509} & 0.8503 & 0.7919 & \textbf{0.8050} & \textbf{0.8295} & 0.8287 \\
        French    & \textbf{0.8479} & 0.8471 & 0.8032 & \textbf{0.8122} & \textbf{0.8269} & 0.8236 \\
        Italian   & \textbf{0.8532} & 0.8516 & 0.8014 & \textbf{0.8133} & \textbf{0.8297} & 0.8261 \\
        Hungarian & \textbf{0.8489} & 0.8406 & \textbf{0.8041} & 0.7896 & \textbf{0.8249} & 0.8135 \\
        Danish    & \textbf{0.8369} & 0.8321 & 0.8024 & \textbf{0.8058} & \textbf{0.8150} & 0.8106 \\
        Romanian  & \textbf{0.8515} & 0.8487 & \textbf{0.8068} & 0.7440 & \textbf{0.8336} & 0.8273 \\
        Swedish   & \textbf{0.8465} & 0.8426 & \textbf{0.8157} & 0.8155 & \textbf{0.8249} & 0.8212 \\
        Slovak    & \textbf{0.8493} & 0.8384 & \textbf{0.7982} & 0.7788 & \textbf{0.8238} & 0.8102 \\
        Dutch     & 0.8364 & - & 0.8002 & - & 0.8177 & -\\
        Portuguese& 0.8386 & - & 0.7755 & - & 0.8233 & -\\
        Slovenian & 0.8424 & - & 0.7969 & - & 0.8258 & -\\
        Polish    & 0.8367 & - & 0.7807 & - & 0.8090 & -\\  
        Latvian   & 0.8268 & - & 0.7846 & - & 0.8083 & -\\
        Lithuanian& 0.8324 & - & 0.7896 & - & 0.8123 & -\\
        Finnish   & 0.8510 & - & 0.8197 & - & 0.8333 & -\\
        Estonian  & 0.8275 & - & 0.8044 & - & 0.8062 & -\\
        Bulgarian & 0.8602 & - & 0.8114 & - & 0.8350 & -\\
        Czech     & 0.8456 & - & 0.7920 & - & 0.8178 & -\\
        \hline
    \end{tabular}
\end{table*}

\onecolumn
\subsection{Translation examples}\label{appendix:okapi:translation_examples}
The following are selected examples from the EU21 and OKAPI translations of the datasets.

\begin{longtable}[c]{p{3cm}p{12cm}}
\caption{Hellaswag translation examples compared between EU21 and OKAPI} \label{tab:EU21_okapi_hellaswag} \\
\toprule
Lang./Dataset & \textbf{Prompt} \\  
Score & \\
\midrule
\endfirsthead
\caption[]{Hellaswag translation examples compared between EU21 and OKAPI}\\
\toprule
Lang./Dataset & \textbf{Prompt} \\  
Score & \\
\midrule
\endhead
\midrule
\multicolumn{2}{r}{{Continued on next page}} \\
\midrule
\endfoot
\bottomrule
\endlastfoot

\textbf{English/Original} &[header] How to choose a good budgie [title] Locate budgies for sale from a reputable seller. [step] You can find budgies for sale online, in want-ads, or in pet stores. You can also get one from your local animal shelter.

- Most animal shelters sell budgies for a flat fee, so this will limit your bank expenses. [substeps] You might also be able to compare prices to get a good deal.

- You'll need to get a price point that is reasonable. Here's a guide to locate an experienced seller : [substeps] Generally , 20-50\% off the appraised price is reasonable.

- Wherever you get your budgie from, make sure the seller has healthy and humanely treated birds. [substeps] Look at online reviews for the sellers you find.

- Basically, you need to find a good seller who has had previous owners. [substeps] Check to see if the parent owned the budgie. \\

\hline

German/EU21 COMET: 0.6329 &  Wie man einen guten Wellensittich auswählt. Suchen Sie nach Wellensittichen, die von einem seriösen Verkäufer verkauft werden. Sie können Wellensittiche im Internet, in Kleinanzeigen oder in Zoohandlungen finden, die zum Verkauf stehen. Sie können auch einen Wellensittich aus Ihrem örtlichen Tierheim bekommen.

-  Die meisten Tierheime verkaufen Wellensittiche gegen eine Pauschalgebühr, so dass sich Ihre Bankkosten in Grenzen halten. Vielleicht können Sie auch die Preise vergleichen, um ein gutes Angebot zu bekommen.

- Sie müssen einen angemessenen Preis finden. Im Allgemeinen sind 20-50 \% des Schätzpreises angemessen.

- Wo auch immer Sie Ihren Wellensittich kaufen, stellen Sie sicher, dass der Verkäufer gesunde und artgerecht gehaltene Vögel hat. Schauen Sie sich die Online-Bewertungen der Verkäufer an, die Sie finden.

- Grundsätzlich müssen Sie einen guten Verkäufer finden, der bereits Vorbesitzer hatte. Erkundigen Sie sich, ob der Wellensittich den Eltern gehört hat. \\

\hline

German/OKAPI COMET: 0.6983 & [header] Wie man einen guten Wellensittich auswählt [title] Lokalisieren Sie Wellensittiche zum Verkauf von einem seriösen Verkäufer. [step] Sie können Wellensittiche zum Verkauf online, in Kleinanzeigen oder in Zoohandlungen finden. Sie können auch einen aus Ihrem örtlichen Tierheim bekommen.

- Die meisten Tierheime verkaufen Wellensittiche zu einem Festpreis, was Ihre Bankkosten begrenzen wird. [substeps] Sie können auch Preise vergleichen, um ein gutes Angebot zu erhalten.

- Sie müssen einen vernünftigen Preis finden. Hier ist ein Leitfaden, um einen erfahrenen Verkäufer zu finden: [substeps] In der Regel sind 20-50\% Rabatt auf den geschätzten Preis angemessen.

- Wo immer Sie Ihren Wellensittich her bekommen, stellen Sie sicher, dass der Verkäufer gesunde und menschlich behandelte Vögel hat. [substeps] Schauen Sie sich Online-Bewertungen für die Verkäufer an, die Sie finden.

- Im Grunde genommen müssen Sie einen guten Verkäufer finden, der bereits Vorbesitzer hatte. [substeps] Überprüfen Sie, ob der Vorbesitzer den Wellensittich besaß. \\

\hline
\hline

\textbf{English/Original} &A green title page appears, with words showing that the video will be of a tenor saxophone player as well as the website to receive to lessons. A black male then appears with a green sweater and black hat playing the sax. the man

- bluffs on the drums and the video continues, at various points, to show individuals playing the instrument.

- continues to play and cheers, shakes hands with the man, throws a napkin, writes on the napkin and gives him a card.

- then lays the instrument across his lap and begins talking about what he is doing.

- jumps into a black and white video and temporarily stops playing, saying, " big calibers ". \\

\hline

French/EU21 COMET: 0.5770 & Une page de titre verte apparaît, avec des mots indiquant que la vidéo sera celle d'un joueur de saxophone ténor, ainsi que le site web permettant de recevoir des leçons. Un homme noir apparaît ensuite, avec un pull vert et un chapeau noir, jouant du saxophone. l'homme

- bluffe à la batterie et la vidéo continue, à différents moments, à montrer des personnes jouant de l'instrument.

-

- continue à jouer et applaudit, serre la main de l'homme, lui lance une serviette, écrit sur la serviette et lui donne une carte. pose ensuite l'instrument sur ses genoux et commence à parler de ce qu'il fait.

- saute dans une vidéo en noir et blanc et s'arrête temporairement de jouer en disant " big calibers " (gros calibres). \\

French/OKAPI COMET: 0.7639 & Une page de titre verte apparaît, avec des mots indiquant que la vidéo sera d'un joueur de saxophone ténor ainsi que le site Web pour recevoir des leçons. Un homme noir apparaît alors avec un pull vert et un chapeau noir en jouant du saxophone. l'homme

- bluffe à la batterie et la vidéo continue, à différents moments, de montrer des individus jouant de l'instrument.

- continue à jouer et crier, serre la main de l'homme, jette une serviette, écrit sur la serviette et lui donne une carte.

- puis pose l'instrument sur ses genoux et commence à parler de ce qu'il fait.

- saut dans une vidéo en noir et blanc et arrête temporairement de jouer, en disant, "gros calibres". \\
 
\hline 
\hline

\textbf{English/Original} &A man drinks from a can next to a hole in the snow. he

- swims slowly down a shallow body of water.

- speaks and gestures to the camera.

- gets his teeth bitten by a young man.

- is drawn in with pick and shovel. \\
\hline

Spanish/EU21 COMET: 0.7803 & Un hombre bebe de una lata junto a un agujero en la nieve. él

- nada lentamente por una masa de agua poco profunda.

- habla y gesticula a la cámara.

- un joven le muerde los dientes.

- se acerca con pico y pala. \\

\hline

Spanish/OKAPI COMET: 0.7082 & Un hombre bebe de una lata junto a un agujero en la nieve. él

- nada lentamente por un cuerpo de agua poco profundo.

- habla y gestiona hacia la cámara.

- recibe una mordida de un joven en los dientes.

- es sacado con pico y pala. \\

\hline
\hline

\textbf{{English/Original}} &[header] How to ease a stomach virus [title] Avoid solid foods for a few hours to let your stomach settle. [step] When you first notice symptoms of the stomach virus, you should allow your stomach to settle. Eating food early on could irritate the stomach lining more, worsening your symptoms.

- For a few hours, avoid solid foods. [title] Take small sips of water or chew ice chips.

- [title] Avoid eating something for at least 10 minutes. [step] Studies have not found that eating for a few hours can relieve stomach symptoms.

- Trying to eat a solid meal a few hours before your nausea attack will only make symptoms worse. If you feel ill after the onset of the symptoms, stop eating.

- [substeps] Try to eat a few hours before work or school, particularly if you're feeling sick. Moving around a bit can also minimize your nausea and help your stomach to settle. \\
\hline

Italian/EU21 COMET: 0.8592 &  Come alleviare un virus intestinale. Evitate i cibi solidi per qualche ora per permettere allo stomaco di calmarsi. Quando si avvertono i primi sintomi di un virus intestinale, è necessario lasciare che lo stomaco si stabilizzi. Mangiare cibo subito potrebbe irritare maggiormente la mucosa gastrica, peggiorando i sintomi.

-  Per qualche ora, evitate i cibi solidi. Bevete piccoli sorsi d'acqua o masticate del ghiaccio.

-  Evitare di mangiare qualcosa per almeno 10 minuti. Gli studi non hanno riscontrato che mangiare per qualche ora possa alleviare i sintomi dello stomaco.

- Cercare di mangiare un pasto solido qualche ora prima dell'attacco di nausea non farà altro che peggiorare i sintomi. Se vi sentite male dopo l'inizio dei sintomi, smettete di mangiare.

-  Cercate di mangiare qualche ora prima del lavoro o della scuola, soprattutto se vi sentite male. Anche muoversi un po' può ridurre la nausea e aiutare lo stomaco a stabilizzarsi. \\

\hline

Italian/OKAPI COMET: 0.6923 & [header] Come alleviare un virus dello stomaco [title] Evita i cibi solidi per alcune ore per far riposare lo stomaco. [step] Quando noti i primi sintomi del virus dello stomaco, dovresti far riposare lo stomaco. Mangiare subito potrebbe irritare ancora di più la mucosa gastrica, peggiorando i sintomi. 

- Per alcune ore, evita i cibi solidi. [title] Fai piccoli sorsi d'acqua o mastica cubetti di ghiaccio.

- [title] Evita di mangiare qualcosa per almeno 10 minuti. [step] Gli studi non hanno trovato alcuna relazione tra il mangiare e il sollievo dei sintomi gastrici.

- Prova a mangiare qualche ora prima del lavoro o della scuola, in particolare se ti senti male. Muoverti un po' può anche ridurre la nausea e aiutare lo stomaco a riposare.

- Cercare di mangiare un pasto solido qualche ora prima dell'attacco di nausea renderà solo i sintomi peggiori. Se ti senti male dopo l'inizio dei sintomi, smetti di mangiare. \\

\hline
\hline

\textbf{English/Original} & A man wearing headphones is in a studio, speaking into a microphone. Other men are shown, one of which begins playing a harmonica. he

- stops playing and says something that makes the dj laugh.

- is seen playing intensely with the camera.

- plays with a jack o lantern and leaves the room.

- seems very pleased with the information he has gathered as he allows the host to speak. \\

\hline
 
Swedish/EU21 COMET: 0.8174 & En man med hörlurar befinner sig i en studio och talar i en mikrofon. Andra män visas, varav en börjar spela munspel. han

- slutar spela och säger något som får dj:n att skratta.

- ses leka intensivt med kameran.

- leker med en jack o lantern och lämnar rummet.

- verkar mycket nöjd med den information han har samlat in när han låter programledaren tala. \\

\hline
 
Swedish/OKAPI COMET: 0.7227 & En man som bär hörlurar är i en studio och talar i en mikrofon. Andra män visas, varav en börjar spela munspel. han

- slutar spela och säger något som får DJ:n att skratta.

- ses spela intensivt med kameran.

- spelar med en Halloween-lampa och lämnar rummet.

- verkar mycket nöjd med informationen han har samlat in och låter värdet tala. \\
 
\hline
\hline

\textbf{English/Original} & [header] How to officiate at a nondenominational funeral service [title] Talk to the family. [step] Call the family and arrange to meet them, preferably in their home. Meeting them in public is ok in a bind, but if that's the arrangement, try not to meet more than two to three people.

- [substeps] If the ceremony is inside a church, don't bother sitting outside on a park bench. If you're a minister or midwife, it's best to arrive by the usual time of worship.

- [substeps] Of course, only with a family member and one of their kids is okay. If no one at your funeral is willing to attend, that's fine.

- People aren't always happy with the small acts of an officiant. In some cases, the family can become drunk, agitated, or violent during or around a service.

- If a funeral home is taking care of other arrangements, the funeral directors may have space at the funeral home where you could meet. Invite the family to have other family members and close friends present if they wish, but be sensitive to their needs and to the space considerations.
\\

\hline

Hungarian/EU21 COMET: 0.8367 &  Hogyan kell egy nem felekezeti temetési szertartást levezetni. Beszélj a családdal. Hívja fel a családot, és egyeztessen velük egy találkozót, lehetőleg az otthonukban. A nyilvános találkozás kötöttségek esetén rendben van, de ha ez a megállapodás, próbáljon meg nem két-három embernél többel találkozni.

-  Ha a szertartás templomban lesz, ne fáradjon azzal, hogy kint üljön egy parkban lévő padon. Ha lelkész vagy bába, a legjobb, ha a szokásos istentiszteleti időre érkezik.

-  Természetesen csak egy családtaggal és az egyik gyermekükkel együtt nem baj. Ha a temetéseden senki sem hajlandó részt venni, az sem baj.

- Az emberek nem mindig örülnek a szertartásvezető apró cselekedeteinek. Egyes esetekben a család részeggé, izgatottá vagy erőszakossá válhat a szertartás alatt vagy körül.

- Ha egy temetkezési vállalat gondoskodik egyéb előkészületekről, a temetkezési vállalatnak lehet, hogy van olyan hely a ravatalozóban, ahol találkozhatnátok. Hívja meg a családot, hogy más családtagok és közeli barátok is jelen legyenek, ha kívánják, de legyen tekintettel az ő igényeikre és a térbeli megfontolásokra. \\

\hline

Hungarian/OKAPI COMET: 0.7061 & [header] Hogyan végezzen szertartást nem vallásos temetésen [title] Beszéljen a családdal. [step] Hívja fel a családot és beszéljen meg egy találkozót, lehetőleg az otthonukban. Nyilvános helyen találkozni is lehet, ha nincs más lehetőség, de ha így van, akkor ne találkozzon több mint két-három emberrel.

- [substeps] Ha a szertartást templomban tartják, ne üljön le kinn egy parki padra. Ha lelkész vagy bábák, a szokásos istentiszteleti időben érkezzen.

- [substeps] Természetesen csak egy családtaggal és az egyik gyerekkel találkozni is lehet. Ha senki sem akar az Ön temetésén részt venni, az sem baj.

- Az emberek nem mindig elégedettek a szertartás megtartásával, még akkor sem, ha az pontosan el lett végezve. Néhány esetben a család részére az esemény körüli korlátozások miatt ajtóstul jön a stressz, ezt jogosak.

- Ha a temetkezési vállalkozás többi rendezvényét szervezi, a temetkezés szervezőinek lehetősége nyílhat arra, hogy találkozzanak a temetőnél. Azonban ha valaki akarja, hogy ott legyenek családtagok vagy közel álló barátok is, tartsuk észben a helyzetet és az igényeket.\\

\hline
\hline

\textbf{English/Original} &A man is running while holding a pole on a track. he

- uses the pole to vault through the air.

- runs down the track, wrapping his legs around the pole.

- runs into a sand pit and starts to jump.

- lands on a pile of sand.\\

\hline

Slovak/EU21 COMET: 0.8336 & Muž beží a pritom drží tyč na dráhe. on

- používa tyč na skok vo vzduchu.

- beží po dráhe a obopína nohy okolo tyče.

- vbehne do pieskoviska a začne skákať.

- dopadne na hromadu piesku. \\
\hline
 
Slovak/OKAPI COMET: 0.7494 & Muž beží po dráhe a drží tyč. on

- používa tyč na skákanie cez vzduch.

- beží po dráhe a obopína nohy okolo tyče.

- beží do pieskového výskoku.

- pristane na kupke piesku. \\

\hline
\hline

\textbf{English/Original} &A small girl runs away from the pinata while holding the stick. The child returns and hits the pinata one time. the people

- continue to remove key ingredients from the bags.

- cheer for the child getting hit.

- watch while the girl throws the stick twice.

- all clap and the girl smiles and turns toward the camera.\\

\hline

Danish/EU21 COMET: 0.8450 & En lille pige løber væk fra pinataen, mens hun holder pinden. Barnet vender tilbage og rammer pinataen én gang. personerne

- fortsætter med at fjerne nøgleingredienser fra poserne.

- jubler over barnet, der bliver ramt.

- ser til, mens pigen kaster pinden to gange.

- alle klapper, og pigen smiler og vender sig mod kameraet. \\

\hline

Danish/OKAPI COMET: 0.7696 & En lille pige løber væk fra piñataen, mens hun holder stokken. Barnet vender tilbage og rammer piñataen en gang. folkene

- fortsætter med at fjerne vigtige ingredienser fra poserne.

- hepper på, at barnet rammer piñataen.

- ser mens pigen kaster stokken to gange.

- klapper alle og pigen smiler og vender sig mod kameraet. \\
 
\hline
\hline

\textbf{English/Original} &A man is sitting on a roof. he

- is using wrap to wrap a pair of skis.

- is ripping level tiles off.

- is holding a rubik's cube.

- starts pulling up roofing on a roof.\\

\hline

Romanian/EU21 COMET: 0.6928 &  Un bărbat stă pe un acoperiș. el

- folosește folie pentru a înfășura o pereche de schiuri.

- smulge țiglele de nivel.

- ține în mână un cub Rubik.

- începe să ridice acoperișul de pe un acoperiș.\\
 
\hline

Romanian/OKAPI COMET: 0.6716 & Un bărbat stă pe acoperiș. el

- folosește învelișul pentru a înveli o pereche de schiuri.

- smulge țigle de nivelare.

- ține un cub Rubik.

- începe să ridice acoperișul de pe acoperiș.\\
 
\hline
\hline

\textbf{English/Original} &[header] How to officiate at a nondenominational funeral service [title] Talk to the family. [step] Call the family and arrange to meet them, preferably in their home. Meeting them in public is ok in a bind, but if that's the arrangement, try not to meet more than two to three people.

- [substeps] If the ceremony is inside a church, don't bother sitting outside on a park bench. If you're a minister or midwife, it's best to arrive by the usual time of worship.

- [substeps] Of course, only with a family member and one of their kids is okay. If no one at your funeral is willing to attend, that's fine.

- People aren't always happy with the small acts of an officiant. In some cases, the family can become drunk, agitated, or violent during or around a service.

- If a funeral home is taking care of other arrangements, the funeral directors may have space at the funeral home where you could meet. Invite the family to have other family members and close friends present if they wish, but be sensitive to their needs and to the space considerations. \\

\hline

Hungarian/EU21 COMET: 0.8367 &  Hogyan kell egy nem felekezeti temetési szertartást levezetni. Beszélj a családdal. Hívja fel a családot, és egyeztessen velük egy találkozót, lehetőleg az otthonukban. A nyilvános találkozás kötöttségek esetén rendben van, de ha ez a megállapodás, próbáljon meg nem két-három embernél többel találkozni.

-  Ha a szertartás templomban lesz, ne fáradjon azzal, hogy kint üljön egy parkban lévő padon. Ha lelkész vagy bába, a legjobb, ha a szokásos istentiszteleti időre érkezik.

-  Természetesen csak egy családtaggal és az egyik gyermekükkel együtt nem baj. Ha a temetéseden senki sem hajlandó részt venni, az sem baj.

- Az emberek nem mindig örülnek a szertartásvezető apró cselekedeteinek. Egyes esetekben a család részeggé, izgatottá vagy erőszakossá válhat a szertartás alatt vagy körül.

- Ha egy temetkezési vállalat gondoskodik egyéb előkészületekről, a temetkezési vállalatnak lehet, hogy van olyan hely a ravatalozóban, ahol találkozhatnátok. Hívja meg a családot, hogy más családtagok és közeli barátok is jelen legyenek, ha kívánják, de legyen tekintettel az ő igényeikre és a térbeli megfontolásokra. \\

\hline

Hungarian/OKAPI COMET: 0.7060 & [header] Hogyan végezzen szertartást nem vallásos temetésen [title] Beszéljen a családdal. [step] Hívja fel a családot és beszéljen meg egy találkozót, lehetőleg az otthonukban. Nyilvános helyen találkozni is lehet, ha nincs más lehetőség, de ha így van, akkor ne találkozzon több mint két-három emberrel.

- [substeps] Ha a szertartást templomban tartják, ne üljön le kinn egy parki padra. Ha lelkész vagy bábák, a szokásos istentiszteleti időben érkezzen.

- [substeps] Természetesen csak egy családtaggal és az egyik gyerekkel találkozni is lehet. Ha senki sem akar az Ön temetésén részt venni, az sem baj.

- Az emberek nem mindig elégedettek a szertartás megtartásával, még akkor sem, ha az pontosan el lett végezve. Néhány esetben a család részére az esemény körüli korlátozások miatt ajtóstul jön a stressz, ezt jogosak.

- Ha a temetkezési vállalkozás többi rendezvényét szervezi, a temetkezés szervezőinek lehetősége nyílhat arra, hogy találkozzanak a temetőnél. Azonban ha valaki akarja, hogy ott legyenek családtagok vagy közel álló barátok is, tartsuk észben a helyzetet és az igényeket. \\

\hline
\end{longtable}

\begin{longtable}[c]{p{2.5cm}p{12cm}}
\caption{ARC and MMLU translation examples each compared between EU21 and OKAPI}
\label{tab:EU21_okapi_arc_mmlu} \\
\toprule
Lang./Dataset & \textbf{Prompt} \\  
Score & \\
\midrule
\endfirsthead
\caption[]{one random ARC and MMLU translation examples each compared between EU21 and OKAPI}\\
\toprule
Lang./Dataset & \textbf{Prompt} \\
Score & \\
\midrule
\endhead
\midrule
\multicolumn{2}{r}{{Continued on next page}} \\
\midrule
\endfoot
\bottomrule
\endlastfoot

\textbf{English/MMLU} &When anatomically modern humans first arrived in the Middle East, who did they encounter?

- Neandertals, the evolutionary descendants of the premodern human inhabitants of Europe and Asia

- Homo erectus, the evolutionary descendants of the premodern human inhabitants of Europe and Asia

- Homo sapiens, who had evolved independently in Asia and arrived there by 200,000 years ago

- All of the above \\

\hline

German/EU21 COMET: 0.8345 &  Auf wen trafen die anatomisch modernen Menschen bei ihrer ersten Ankunft im Nahen Osten?

- Neandertaler, die evolutionären Nachfahren der vormodernen menschlichen Bewohner Europas und Asiens

- Homo erectus, die evolutionären Nachfahren der vormodernen menschlichen Bewohner Europas und Asiens

- Homo sapiens, der sich unabhängig von den Menschen in Asien entwickelt hatte und dort vor 200.000 Jahren ankam

- Alle oben genannten \\

\hline

German/OKAPI COMET: 0.8402 & Als anatomisch moderne Menschen erstmals im Nahen Osten ankamen, wen trafen sie?

- Neandertaler, die evolutionären Nachkommen der frühzeitlichen menschlichen Bewohner Europas und Asiens

- Homo erectus, die evolutionären Nachkommen der frühzeitlichen menschlichen Bewohner Europas und Asiens

- Homo sapiens, die sich unabhängig in Asien weiterentwickelt hatten und bis vor 200.000 Jahren dort ankamen

- Alle oben Genannten  \\

\hline
\hline

\textbf{English/ARC} & A class plans an investigation to see which brand of light bulb lasts the longest. Which of these steps should come first?

- Repeat the investigation.

- Write a report of the results.

- Make a table for recording data.

- Make daily observations of the light bulbs. \\

\hline

German/EU21 COMET: 0.8546 & Eine Klasse plant eine Untersuchung, um herauszufinden, welche Marke von Glühbirnen am längsten hält. Welcher dieser Schritte sollte zuerst erfolgen?

- Wiederhole die Untersuchung.

- Schreibe einen Bericht über die Ergebnisse.

- Fertige eine Tabelle an, um die Daten festzuhalten.

- Beobachte täglich die Glühbirnen. \\

\hline

German/OKAPI COMET: 0.8548 & Eine Klasse plant eine Untersuchung, um herauszufinden, welche Marke von Glühbirnen am längsten hält. Welcher dieser Schritte sollte als Erstes durchgeführt werden?

- Wiederholen Sie die Untersuchung.

- Schreiben Sie einen Bericht über die Ergebnisse.

- Erstellen Sie eine Tabelle zur Aufzeichnung von Daten.

- Machen Sie tägliche Beobachtungen der Glühbirnen. \\
\end{longtable}

\twocolumn

\section{Other}\label{appendix:other}
\subsection{License}
\label{appendix:exprtimental:license}
MMLU, Hellaswag and GSM8K were released under MIT license; ARC under CC-BY-SA-4.0 license; TruthfulQA under Apache 2.0 license. 

EU20-MMLU, EU20-HellaSwag, EU20-GSM8K and EU20-TruthfulQA will be released with an permissive license, we ensure that all original licenses and notices are included. EU20-ARC will be released under the CC-BY-SA-4.0 license under the exactly the same terms as the original dataset ARC.

\subsection{Use of AI Assistants}
\label{appendix:ai_assistants}
We acknowledge the use of ChatGPT-4 to generate code snippets for the Figures 1-3. The code was then extended and carefully reviewed.

\end{document}